\documentclass{article}

\usepackage{microtype}
\usepackage{graphicx}
\usepackage{booktabs} 
\usepackage{multirow}
\usepackage{amsmath}
\usepackage{array}
\usepackage{lipsum}
\usepackage{makecell}
\usepackage{ragged2e}
\usepackage[section]{placeins}
\usepackage{subcaption}
\usepackage{caption}
\usepackage{tcolorbox}
\tcbuselibrary{skins, breakable}
\usepackage{xcolor}
\usepackage{colortbl}
\usepackage{tabularx}
\usepackage{booktabs}
\definecolor{transgray}{gray}{0.92}
\usepackage{hyperref}
\usepackage{wrapfig}
\usepackage{enumitem}
\usepackage[ruled,vlined,algo2e]{algorithm2e}
\usepackage{amssymb}
\usepackage{float}

\usepackage[accepted]{icml2025}

\usepackage{amsmath}
\usepackage{amssymb}
\usepackage{mathtools}
\usepackage{amsthm}

\usepackage[capitalize,noabbrev]{cleveref}

\theoremstyle{plain}
\newtheorem{theorem}{Theorem}[section]
\newtheorem{proposition}[theorem]{Proposition}

\theoremstyle{definition}
\newtheorem{definition}[theorem]{Definition}
\newtheorem{assumption}[theorem]{Assumption}
\theoremstyle{remark}

\definecolor{lightgray}{RGB}{230,230,230}
\definecolor{lightred}{RGB}{250,220,220}

\usepackage[textsize=tiny]{todonotes}

\icmltitlerunning{Diversified Scaling Inference in Time Series Foundation Models}

\makeatletter
\newcommand{\icmlpresetaffiliation}[1]{%
  \ifcsname the@affil#1\endcsname\else
    \stepcounter{@affiliationcounter}%
    \newcounter{@affil#1}%
    \setcounter{@affil#1}{\value{@affiliationcounter}}%
  \fi
}
\makeatother

\begin{document}

\twocolumn[
\icmltitle{Diversified Scaling Inference in Time Series Foundation Models}



\icmlsetsymbol{equal}{*}
\icmlsetsymbol{lead}{\textdagger}
\icmlsetsymbol{correspondence}{$\ddagger$}

\icmlpresetaffiliation{cmu}
\icmlpresetaffiliation{hust}
\icmlpresetaffiliation{mbzuai}
\icmlpresetaffiliation{ox}

\begin{icmlauthorlist}
\icmlauthor{Ruijin Hua}{equal,hust}
\icmlauthor{Zichuan Liu}{equal,lead,cmu}
\icmlauthor{Kun Zhang}{cmu,mbzuai}
\icmlauthor{Yiyuan Yang}{correspondence,ox}
\end{icmlauthorlist}

\icmlaffiliation{hust}{HUST}
\icmlaffiliation{ox}{University of Oxford}
\icmlaffiliation{cmu}{Carnegie Mellon University}
\icmlaffiliation{mbzuai}{MBZUAI}

\icmlcorrespondingauthor{Yiyuan Yang}{yiyuan.yang@cs.ox.ac.uk}

\icmlkeywords{Machine Learning, ICML}

\vskip 0.3in
]



\printAffiliationsAndNotice{\icmlEqualContribution\icmlLead} 

\begin{abstract}
The advancement of Time Series Foundation Models (TSFMs) has been driven primarily by large-scale pre-training, but inference-time compute potential remains largely untapped.
This work systematically investigates two questions: how do TSFMs behave under standard sampling-based inference scaling, and can controlled sampling diversity enhance performance? 
We first examine the properties of TSFMs under standard sampling often fail to adhere to scaling laws due to insufficient exploration of the solution space.
Building on this, we then delve into diversified inference scaling via tailored time series perturbations to expand the generative distribution’s support. 
We theoretically analyze the diversity-fidelity trade-off and derive a critical sample threshold for diversified sampling to outperform standard sampling. 
Extensive experiments across various TSFMs and datasets show proper diversified inference scaling yields substantial performance gains without parameter updates, establishing inference design as a critical, compute-efficient dimension of TSFM optimization.
As an application, we propose RobustMSE, a rigorous metric to quantify the headroom performance of TSFM under a fixed budget.
Overall, our findings clarify these factor interactions, enabling reliable performance via diverse large-scale inference time series in parallel environments without re-training TSFMs.
\end{abstract}

\section{Introduction}
The emergence of Time Series Foundation Models (TSFMs) has fundamentally altered the landscape of forecasting~\cite{das2024decoder,li2025tsfm}. 
By leveraging large-scale pre-training on heterogeneous corpora, models such as Chronos, TimesFM, and Moirai have demonstrated remarkable zero-shot generalization and the ability to learn transferable temporal representations~\cite{goswami2024moment,ansari2024chronoslearninglanguagetime,das2024decoder,woo2024unified}. 
To date, the primary driver of this progress has been the scaling of training compute, increasing model parameters and dataset size to refine representation quality and  generalization~\cite{kaplan2020scaling,hoffmann2022trainingcomputeoptimallargelanguage}. 
However, the potential of inference compute, i.e., the computational resources expended at test time to refine predictions, remains largely untapped in the time series domain. 
This leaves a critical open question: \textit{To what extent can the performance of a fixed TSFM be improved solely by scaling inference strategies, without additional training or parameter updates?}

\begin{figure}[!t]
    \centering
    \includegraphics[width=1\linewidth]{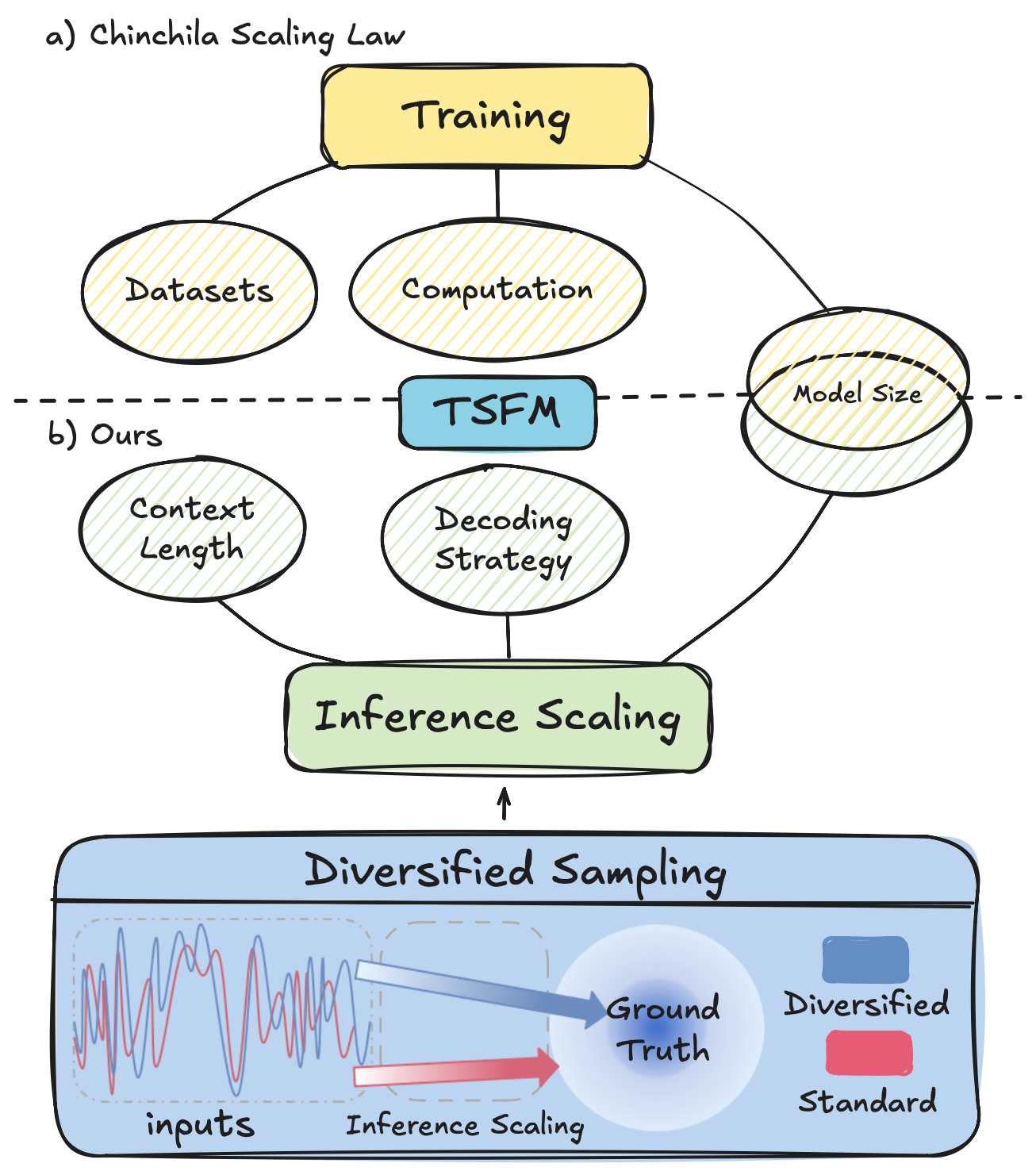}\vspace{-3mm}
    \caption{Illustration the relationship between scaling training~\cite{hoffmann2022trainingcomputeoptimallargelanguage, yao2024towards} and diversified scaling inference in time series foundation models.}
    \label{fig:intro} \vspace{-5mm}
\end{figure}

In adjacent generative domains, particularly Natural Language Processing (NLP) and reasoning, extending computation at inference time has proven to be a potent lever for performance gains~\cite{chen2024alphamath,wang2025diversified,kadavath2022language}.
Strategies such as chain-of-thought, best-of-$N$ sampling, and verifiers allow models to explore a broader solution space, effectively trading test-time compute for accuracy~\cite{wei2022chain,wang2022self,sun2024fast}, especially when combined with multi-sample aggregation~\cite{wang2022self}.
Critically, recent findings suggest that diversity is a prerequisite for effective scaling. Simply sampling more from a biased distribution yields diminishing returns, whereas introducing controlled stochasticity or input diversity expands the effective support of the generation distribution, enabling the model to uncover higher-quality predictions~\cite{wang2025diversified,vijayakumar2016diverse}.
These findings suggest that inference exploration when properly structured can serve as a powerful complement to large foundation models.

Motivated by these developments, we present a systematic study of diversified scaling inference in TSFMs, as shown in Figure~\ref{fig:intro}. 
We posit that the stochastic nature of time series forecasting makes it well-suited for inference-time search, whereas standard greedy or low-temperature sampling methods may often converge to sub-optimal modes due to limited exploration.
To address this, we investigate two complementary axes of scaling: (i) \textbf{Inference Scaling}, where we analyze the effects of expanding the candidate pool size under varied model sizes, context lengths, and temperatures, and (ii) \textbf{Diversified Sampling}, where we introduce controlled diversity at the input level via task-agnostic and task-specific perturbations.
Unlike standard sampling, which is constrained by the fixed conditioning of the time series context, diversified sampling approximates a marginalization over the input neighborhood, theoretically expanding the search space to cover regions closer to the ground truth.
We unify these strategies under a rigorous evaluation framework, utilizing aggregation mechanisms such as majority voting and exact match to synthesize candidate predictions. 
Furthermore, to fairly quantify the attainable utility of a model given a compute budget, we propose RobustMSE, a metric designed to capture the performance envelope of TSFMs under optimal sampling strategies.

Our empirical analysis across multiple architectures and datasets reveals consistent scaling laws for TSFM inference. 
We observe that while scaling the number of samples alone improves stability, there is still an upper limit for TSFM that does not follow the laws similar to NLP~\cite{kaplan2020scaling,hoffmann2022trainingcomputeoptimallargelanguage}. 
For example, smaller models occasionally surpass larger models, and the performance of length and inference times cannot achieve linear growth under log scales.
In addition, we demonstrated that when appropriate diversified sampling takes effect, demonstrating that significant gains in accuracy can be unlocked with different aggregations.
These findings establish inference design not merely as a post-processing step, but as a critical dimension of TSFM scaling.
By clarifying how these factors interact, our study helps users easily achieve reliable performance by designing diverse large-scale time series in parallel environments, rather than pre-training a TSFM.

Our contributions are summarized as follows:\vspace{-3mm}
\begin{itemize}
    \item We provide the first comprehensive study of inference-time computation in TSFMs, characterizing the interaction between model capacity, context length, and temperature, establishing that TSFMs exhibit predictable performance gains when scaling test-time compute.
    \item We theoretically motivate and empirically validate the necessity of exploration diversity. We show that diversifying the input via controlled perturbations strictly expands the support of the sampling distribution.
    \item We systematically perform an empirical evaluation of diversified sampling in various TSFMs and datasets, showing that appropriate perturbations can consistently improve performance under scaling inference.
    \item As an application, we introduce RobustMSE, a new metric allowing for a fair comparison of the headroom of TSFMs under a fixed inference budget.
\end{itemize}

\section{Preliminaries}
\paragraph{Time Series Foundation Model} A forecasting TSFM is regarded as a generative model $f: \mathbb{R}^{L \times D} \to \mathbb{R}^{H \times D_{\text{out}}}$ with parameters $\theta$ pre-trained on a massive, heterogeneous collection of time series, where $D$ and $D_{\text{out}}$ denote the input and output dimensions respectively (typically $1 \leq D_{\text{out}} \leq D$). 
Given a historical sequence $\mathbf{X} = \{\mathbf{x}_1, \dots, \mathbf{x}_L\} \in  \mathbb{R}^{L \times D}$ with a context length $L$, the model produces a predicted sequence $\hat{\mathbf{Y}} = \{\hat{\mathbf{y}}_{L+1}, \dots, \hat{\mathbf{y}}_{L+H}\} \in \mathbb{R}^{H \times D_{\text{out}}}$ for a forecast horizon $H$. The model is optimized by minimizing a global loss $\mathcal{L}$ over the time series as
\begin{equation*}
\theta^* = \arg\min_\theta \sum_{\mathbf{X} } \mathcal{L} ( \mathbf{Y}, \hat{\mathbf{Y}} ),
\end{equation*}
where $\mathbf{Y} = \{\mathbf{y}_{L+1}, \dots, \mathbf{y}_{L+H}\} \in \mathbb{R}^{H \times D_{\text{out}}}$ represents the ground-truth values of target variables. 
This formulation allows the model to leverage multi-dimensional context (e.g., exogenous variables) to predict specific target dimensions, while maintaining robust zero-shot generalization across diverse temporal patterns and scales.

\begin{figure*}[!t]
    \centering
    \begin{subfigure}{0.3\textwidth}
        \includegraphics[width=\linewidth]{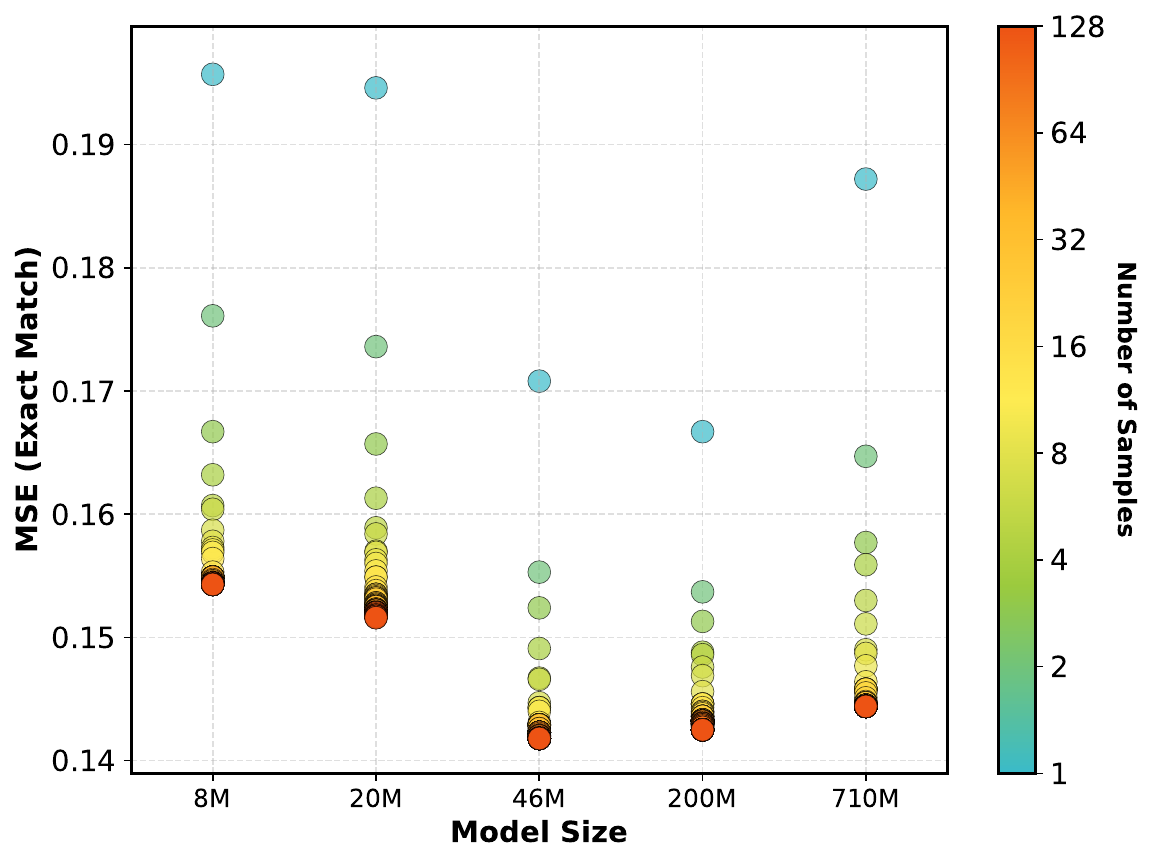}\vspace{-1mm}
        \caption{EM  vs. Model Size}
    \end{subfigure}
    \begin{subfigure}{0.34\textwidth}
        \includegraphics[width=\linewidth]{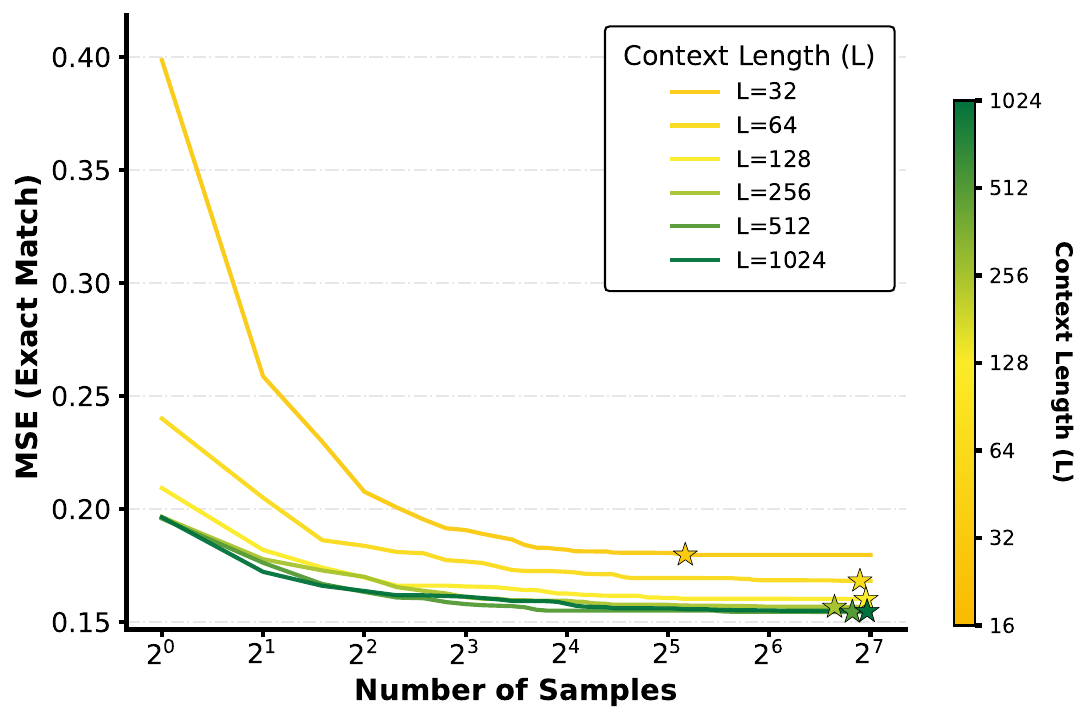}\vspace{-1mm}
        \caption{EM  vs. Context Length}
    \end{subfigure}
    \begin{subfigure}{0.34\textwidth}
        \includegraphics[width=\linewidth]{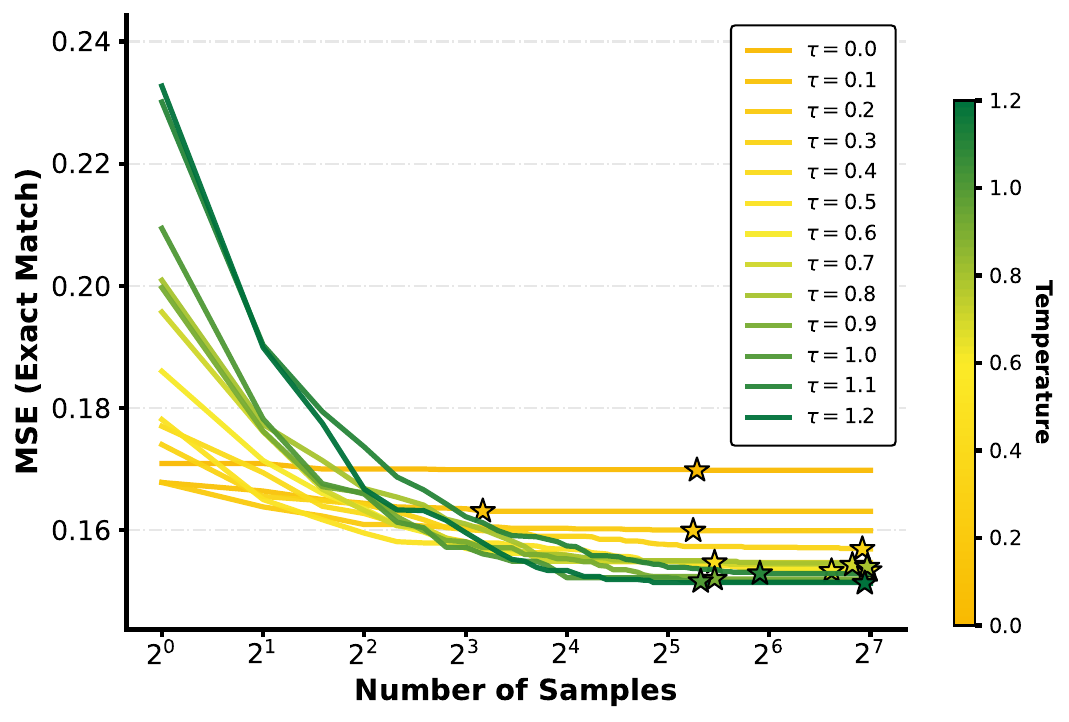}\vspace{-1mm}
        \caption{EM  vs. Temperature}
    \end{subfigure}
    \\
    \begin{subfigure}{0.3\textwidth}
        \includegraphics[width=\linewidth]{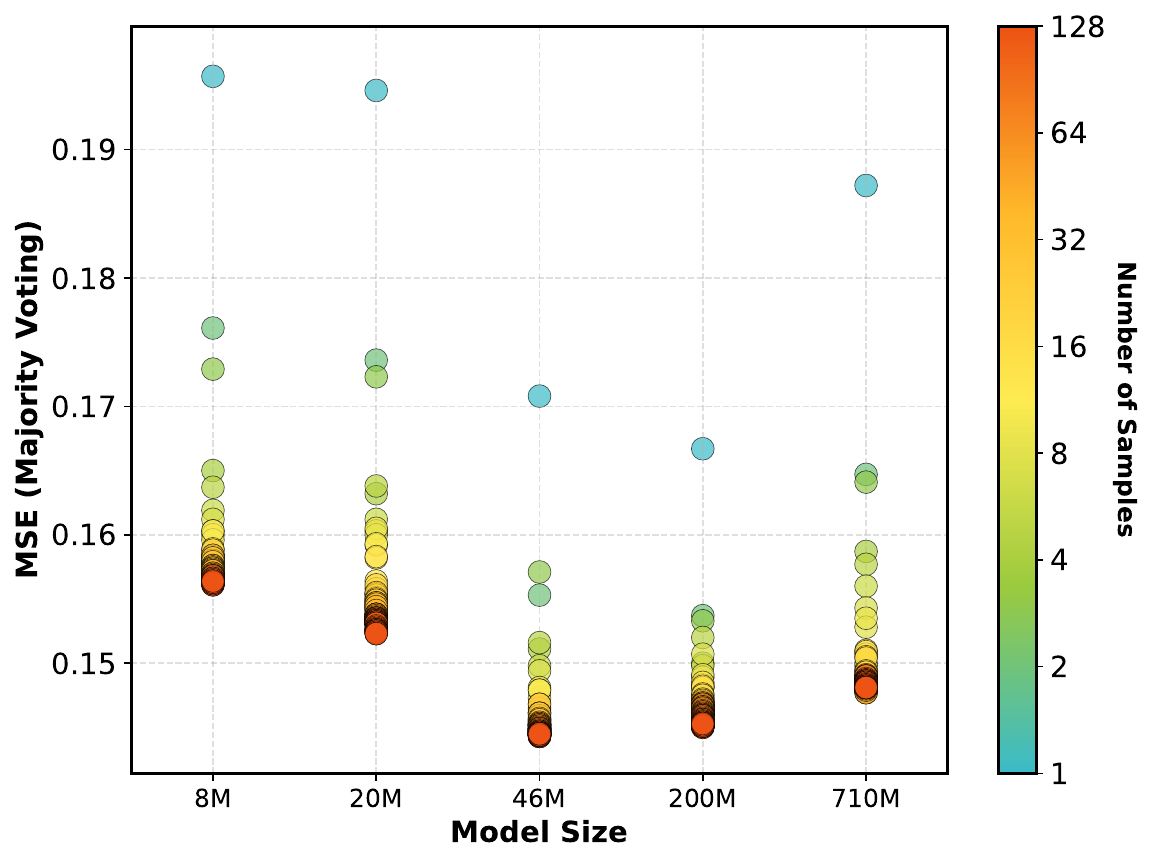}\vspace{-1mm}
        \caption{MV  vs. Model Size}
    \end{subfigure}
    \begin{subfigure}{0.34\textwidth}
        \includegraphics[width=\linewidth]{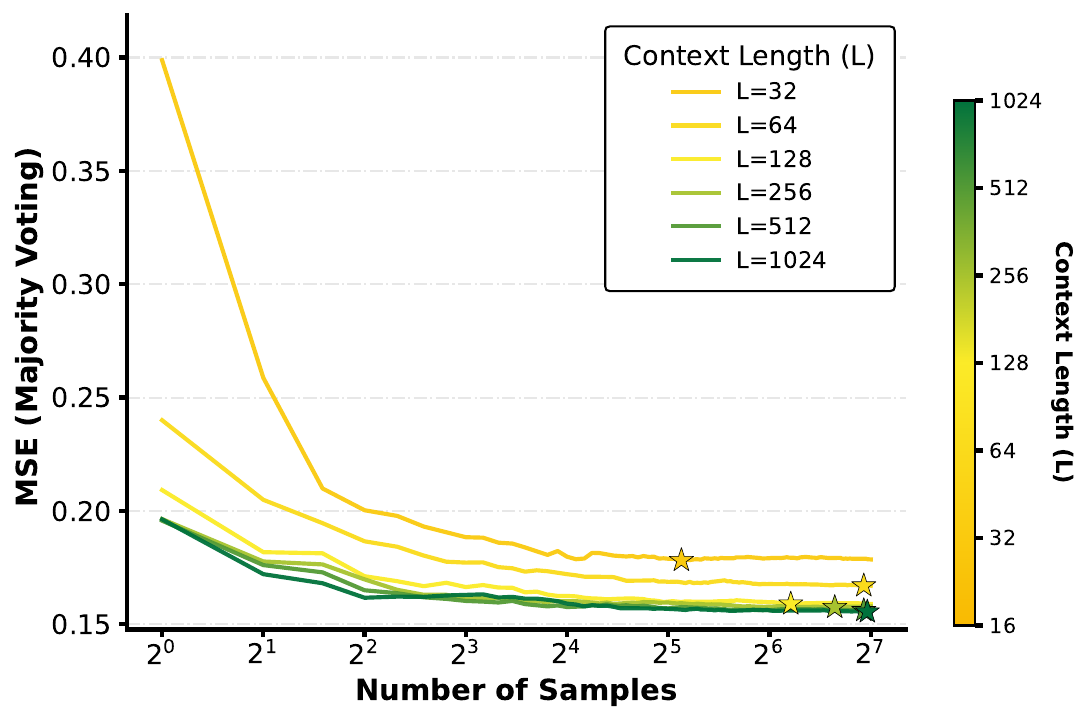}\vspace{-1mm}
        \caption{MV  vs. Context Length}
    \end{subfigure}
    \begin{subfigure}{0.34\textwidth}
        \includegraphics[width=\linewidth]{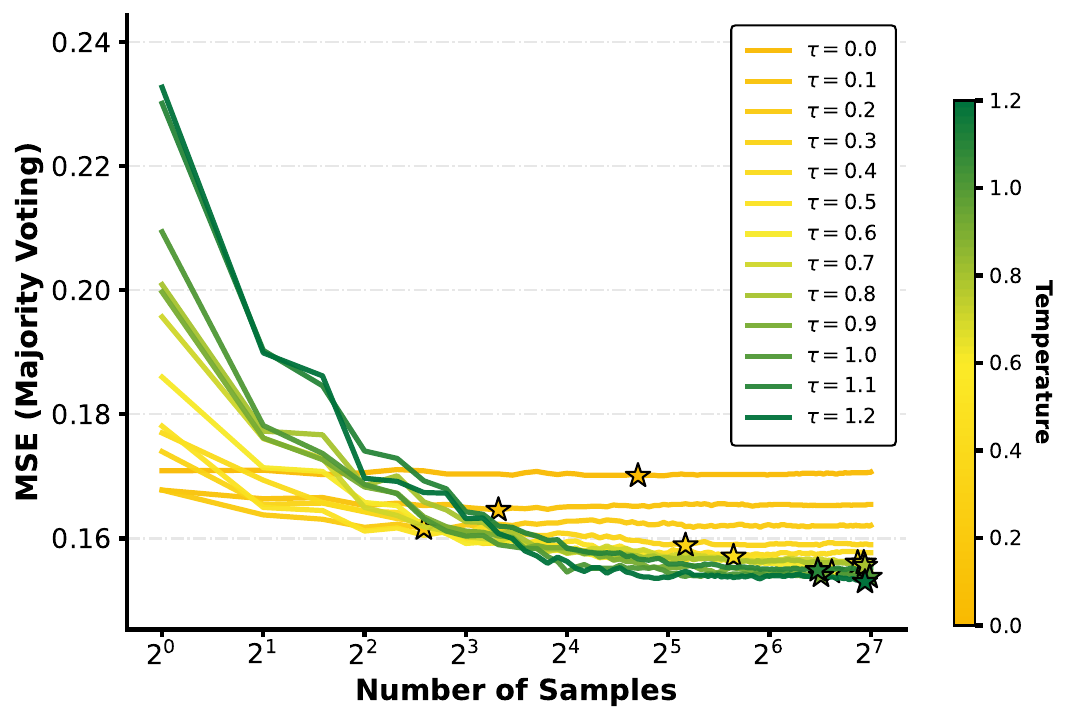}\vspace{-1mm}
        \caption{MV  vs. Temperature}
    \end{subfigure}
\vspace{-3mm}
    \caption{\textbf{Performance of \textit{Chronos} on the data \textit{ETTh1} under different scaling factors and aggregation functions}. We systematically evaluate the impact of three critical factors: model size, context length, and temperature. 
    The number of sampling candidates is set to 128, plotted on a logarithmic axis.
    The stars indicate the initial convergence number to the minimum MSE for each configuration.
    }    \label{fig:main}\vspace{-4mm}
\end{figure*}

\paragraph{Sampling Strategy}
Given an input $\mathbf{X}$ and a generative model $f(\cdot \mid \mathbf{X})$, a sampling strategy $S$ defines a probabilistic procedure for generating a finite set of candidate outputs
\begin{equation*}
\mathcal{Y} = \{\hat{\mathbf{Y}}_1, \hat{\mathbf{Y}}_2, \ldots, \hat{\mathbf{Y}}_N\}, \quad \hat{\mathbf{Y}}_i \sim f(\cdot \mid \mathbf{X}, S),
\end{equation*}
where $N$ is the number of sampled candidates.
The sampling strategy may correspond to a fixed decoding distribution such as temperature $\tau$ and top-$p$ nucleus sampling, or a diversified strategy $\pi_S$ from perturbed samples $\mathbf{X}'$, formally as $\mathbf{X}' \sim \pi_S(\cdot \mid \mathbf{X})$.
The resulting candidate set $\mathcal{Y}$ is subsequently combined by an aggregation function $A$, e.g., exact match or majority voting~\cite{wu2024inference}, to produce the final prediction.
These sampling strategies have prepared us adequately for the scaling inference of TSFMs.

\paragraph{Scenarios and Settings}
To better investigate the impact of scaling inference on model performance in time series forecasting, we select 4 datasets under long-term settings, including \textit{ETTh1}, \textit{ETTm1}, \textit{Electricity}, and \textit{Traffic} from TSLib\footnote{\href{https://github.com/thuml/Time-Series-Library}{Time-Series-Library}}, which cover variations in real-world application domains, data frequencies, and noise levels. To verify the universal applicability of the diversified sampling approach we propose, we choose 4 representative TSFMs based on these datasets: \textit{TimesFM}~\cite{das2024decoder} and \textit{Chronos}~\cite{ansari2024chronoslearninglanguagetime} for the encoder-decoder, \textit{Time-MoE}~\cite{shi2024time} for the decoder-only, and \textit{Moirai}~\cite{woo2024unified} for the encoder-only.  All evaluations are assessed based on the Mean Squared Error~(MSE) loss. We then systematically evaluate the general performance gains of scaling-based inference strategies across these TSFMs. Detailed experimental settings are provided in Appendix~\ref{app:exp_params}.

\section{Scaling Inference in TSFM}
\label{Sec:Scaling Inference in TSFM}

While prior work~\cite{yao2024towards} has investigated neural scaling laws in the training of TSFMs, the behavior during the inference phase remains largely underexplored. 
Evidence from the LLMs~\cite{wang2025diversified} suggests that allocating additional compute at inference time can yield greater efficiency gains than simply increasing model parameters. 
Motivated by this paradigm shift, we explore a fundamental question in this section: \textit{Do TSFMs exhibit inference scaling behaviors analogous to those observed in LLMs?}
To answer this question, we proceed from the aggregators and examine three dimensions of TSFMs: model size, context length, and decoding strategy.

First, we consider two widely adopted aggregators for candidates, Exact Match (EM) and Majority Voting (MV). Notably, EM adopts oracle style selection by exploiting the optimal individual candidate, typically used for validation; whereas MV leverages distributed aggregation to enhance robustness, typically deployed for inference.

\begin{itemize}[left=0pt]
\item \textbf{Exact Match}  also termed best-of-$N$ in reasoning~\cite{wang2022self} and pass@$N$ in code~\cite{chen2024alphamath} tasks, we consider it as EM@$N$ in our time series forecasting task. This aggregator prepares a set of time series candidates $\mathcal{Y}$ and picks the one with the lowest MSE loss compared with the ground-truth:
\begin{equation}
\hat{\mathbf{Y}}_{\texttt{EM}} = \arg\min_{\hat{\mathbf{Y}}_i \in \mathcal{Y}} \frac{1}{H}|| \hat{\mathbf{Y}}_i - \mathbf{Y} ||^2_2, ~ \texttt{EM} = \mathcal{L}(\hat{\mathbf{Y}}_{\texttt{EM}}, \mathbf{Y}).
\end{equation}

\item \textbf{Majority Voting} is referred to as self-consistency in reasoning tasks~\cite{brown2024large}, generating multiple candidates and determining the final answer as the most frequently occurring one among all outputs.
In our scenarios, MV computes the median across candidates:
\begin{equation}
\hat{\mathbf{Y}}_{\texttt{MV}}  = \operatorname{quantile}_{0.5}(\mathcal{Y}), ~ \texttt{MV} = \mathcal{L} (\hat{\mathbf{Y}}_{\texttt{MV}}, \mathbf{Y}).
\end{equation}
\end{itemize}

Now, we empirically evaluate the scaling inference of TSFMs in model size, context length, and decoding strategy.

\paragraph{Effect of Model Size} 
Based on the research on the scaling laws relating the performance and parameter size of LLMs~\cite{kaplan2020scaling}, our study systematically evaluates the effect of TSFM size on the scaling inference. 
We conduct our experiments on different sizes of the same model to test the joint effect of model size and number of samples under both aggregation functions. 
In Figure~\ref{fig:main} \textit{(a)} and \textit{(d)}, and consistently across additional TSFMs in Appendix~\ref{app:scaling inference experiments}, MSE loss generally decreases as model size increases. 
However, this improvement is not strictly monotonic on log scales. 
In several settings (e.g., Chronos on the data ETTh1), performance reaches its optimum at an intermediate model size and slightly degrades as the model is further scaled. 
Overall, across all datasets and TSFMs, increasing model sizes yields substantial performance gains up to approximately $80\%$ in the best cases, yet the absence of consistent indicates that performance in TSFMs does not strictly follow scaling inference. 
This observation suggests that simply enlarging model capacity is insufficient to guarantee stable inference improvements, highlighting the need for complementary mechanisms beyond parameter scaling.

\paragraph{Effect of Context Length}
We next analyze the impact of historical context length ($L$) on inference scaling behavior, as context size is known to play a critical role in temporal modeling~\cite{wu2023timesnettemporal2dvariationmodeling}. 
In all experiments (Figure~\ref{fig:main} \textit{(b)} and \textit{(e)} and Appendix~\ref{app:scaling inference experiments}), a consistent pattern emerges when jointly considering context length and the number of samples. 
With a small sampling budget, increasing the context length generally reduces MSE, whereas as the number of samples increases, prediction errors decrease rapidly and often converge across different context lengths, yielding performance gains of up to approximately $90\%$. However, such convergence is not universal. 
Across different models and datasets, increasing the number of samples often reduces MSE but does not consistently eliminate performance differences across context lengths, suggesting that the effectiveness of longer context depends on both model architecture and data characteristics. Moreover, under limited sampling, MV consistently achieves faster error reduction and earlier convergence than EM. Overall, these results show that extending context length alone does not produce a strictly monotonic or universal inference scaling effect.

\paragraph{Varying Sampling Temperatures} 

As sampling temperature ($\tau$) can directly control the randomness of generation and higher temperatures can increase output diversity~\cite{holtzman2019curious}. We explore the performance of scaling inference by adjusting varying sampling temperatures. 
Provided in Figure~\ref{fig:main} \textit{(c)} and \textit{(f)}, and consistently observed across Appendix~\ref{app:scaling inference experiments}, we evaluate temperature swept over in 0.1 increments from 0.0 to 1.2. Across all datasets, temperature exhibits heterogeneous effects. Under small sampling budgets, higher temperatures often increase MSE due to excessive stochasticity, while low temperatures yield nearly flat MSE curves, reflecting limited diversity and marginal gains from additional samples. As $N$ increases, higher temperature settings increasingly benefit from aggregation, with MSE rapidly decreasing and often surpassing low temperature performance. Overall, temperature does not independently induce inference scaling; rather, its effect is tightly coupled with sampling budget and decoding capability.

\begin{tcolorbox}[colback=orange!10, colframe=orange, boxrule=0.5pt, sharp corners, width=\linewidth, top=2pt, bottom=2pt]
\small
\includegraphics[width=0.03\linewidth]{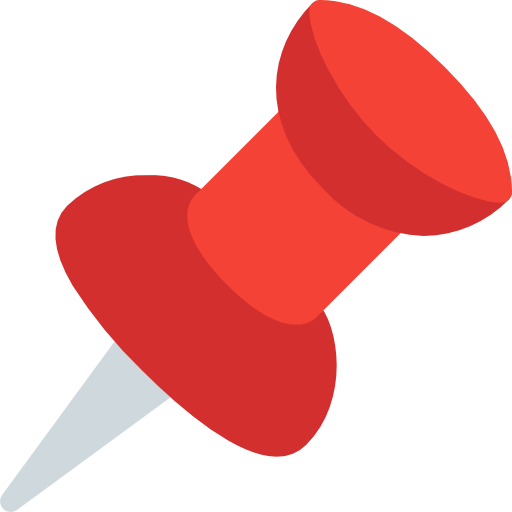}  \textit{\textbf{Takeaways:} The results do not strictly adhere to logarithmic scaling laws, and sampling alone is insufficient to foster meaningful diversity. Specifically, when predicting under standard sampling, aggregation primarily mitigates noise rather than yielding complementary information.}
\normalsize
\end{tcolorbox}

\textbf{Remark.} The bottleneck of standard scaling lies in its distributional limitations. While scaling inference improves performance, standard sampling saturates rapidly due to distributional bias in the fixed predictive distribution $P_\theta(\hat{\mathbf{Y}}|\mathbf{X})$. Simply increasing the sample size merely refines the approximation of this potentially biased distribution, leading to over-exploration of sub-optimal regions while failing to reach the ground truth in low-probability tails. To break this limitation, we introduce input diversity, which constructs a marginal mixture distribution to expand the support set and calibrate the inference toward the true solution.

\section{Encouraging Diversified Sampling}
\label{sec:diversified sampling}
In this section, we first present a theoretical insight into sample diversification strategies for enhancing TSFM performance.
Guided by this, we instantiate a set of perturbations that are likely to be effective for diversified scaling inference, including task-agnostic and task-specific perturbations.

\subsection{Theoretical Analysis of Diversified Sampling}
\label{sec:mathematical proof}

To understand why diversified sampling improves the TSFM performance in best-of-$N$, we provide a theoretical perspective. We examine the behavior of the minimum loss as the sample size $N$ approaches infinity, as well as the expected performance under finite $N$.
 The proofs are presented in Appendix~\ref{proof}.
First, we formalize the settings. 
\begin{definition}[\textbf{Sampling Strategies}]Let $P(\hat{\mathbf{Y}} | \mathbf{X})$ be the probability density function of the TSFM output given $\mathbf{X}$. Let $\mathcal{L}(\mathbf{Y}, \hat{\mathbf{Y}} ) $  be the forecasting loss.  We first consider two sampling strategies to generate a candidate set $\mathcal{Y}$: 

\textit{Standard Sampling}: The candidate set is generated by sampling $N$ times, i.i.d. with a fixed configuration $\zeta$ (e.g., model size, context length, decoding seed, temperature, etc.), from a fixed distribution:
\begin{equation}
    \hat{\mathbf{Y}}_i \sim P(\cdot | \mathbf{X}, \zeta), \quad \forall i \in \{1, \dots, N\}.
\end{equation}
We denote the minimum EM for the standard sampling as $\texttt{EM}_N^{\text{std}} = \min_{i=1}^N \mathcal{L}(\hat{\mathbf{Y}}_i, \mathbf{Y})$.

\textit{Diversified Sampling}: Let $Q(\mathbf{X}' | \mathbf{X}, \pi_S)$ be a distribution of $\pi_S$ over inputs, i.e., $Q$ is a test-time augmentation. The candidate set is generated by the marginal mixture distribution:
\begin{equation}
    \hat{\mathbf{Y}}_i \sim P_{\text{mix}}= \int P(\cdot | \mathbf{X}', \zeta) Q(\mathbf{X}' | \mathbf{X}, \pi_S) d\mathbf{X}'.
\end{equation}
Equivalently, for each $i$, we first sample $\mathbf{X}' \sim Q(\cdot|\mathbf{X}, \pi_S)$ and then $ \hat{\mathbf{Y}}_i  \sim P(\cdot|\mathbf{X}', \zeta)$. We denote the  EM for diversified sampling as $\texttt{EM}_N^{\text{div}} = \min_{i=1}^N \mathcal{L}(\hat{\mathbf{Y}}_i, \mathbf{Y})$.
\end{definition}

\paragraph{Asymptotic Analysis ($N \to \infty$)} Our first result concerns the potential of diversified sampling when the sample budget is unlimited. 
We demonstrate that introducing input perturbations strictly expands the search space without discarding the original solution capabilities. 

\begin{proposition}[\textbf{Asymptotic Lower Bound}]
\label{prop:asymptotic}
Let $\mathcal{S}_{\text{std}}$ and $\mathcal{S}_{\text{div}}$ be the support sets of the standard decoding and diversified sampling distributions, respectively. 
Under the mild assumption that the perturbation distribution covers the original input, we have $\mathcal{S}_{\text{std}} \subseteq \mathcal{S}_{\text{div}}$. Consequently, as $N \to \infty$, the minimum loss of diversified sampling is bounded by:
\begin{equation}
    \lim_{N \to \infty} \texttt{EM}_N^{\text{div}} \le \lim_{N \to \infty} \texttt{EM}_N^{\text{std}}.
\end{equation}
The inequality becomes strict (e.g., $<$) if the perturbation allows the model to explore regions of the manifold closer to the ground truth $\mathbf{Y}$ that are inaccessible under $\zeta$.
\end{proposition}

\textbf{Remark.} This proposition suggests that diversified sampling effectively mitigates the inductive bias of the fixed model configuration. 
By expanding the support set, the method can access blind spots of the TSFM. As $N$ increases, the probability of hitting these superior regions approaches certainty, guaranteeing a lower bound for EM.

\paragraph{Finite Sample Analysis ($N < \infty$)}
While asymptotic analysis guarantees superiority in the limit, practical TSFM applications operate with a finite budget $N$. 
We model this as a trade-off between the \textit{quality of the best subspace} and the \textit{probability of sampling from it}. We model the expected minimum EM to derive a critical sample size $N^*$.

\begin{proposition}[\textbf{Critical Sample Threshold}]
\label{prop:finite}
Let $\mathcal{L}_0$ be the baseline loss. For a diversified strategy that introduces a variance in performance (yielding lower loss $\mathcal{L}_{\text{good}}$ with probability $\rho$ and higher loss $\mathcal{L}_{\text{bad}}$ otherwise), there exists a critical threshold $N^*$ given by:
\begin{equation}
    N^* = \frac{\ln \left( \frac{\mathcal{L}_{\text{bad}} - \mathcal{L}_{\text{good}}}{\mathcal{L}_0 - \mathcal{L}_{\text{good}}} \right)}{\ln \left( \frac{1}{1-\rho} \right)}.
\end{equation}
Diversified sampling outperforms standard sampling in expectation (i.e., $\mathbb{E}[\texttt{EM}_N^{\text{div}}] < \mathcal{L}_0$) if and only if the sample budget satisfies $N > N^*$.
\end{proposition}

\textbf{Remark.} This result highlights a ``risk-reward" dynamic. When $N$ is small ($N \le N^*$), the stability of standard sampling is preferred to avoid the risk of harmful perturbations ($\mathcal{L}_{\text{bad}}$). However, as the budget $N$ exceeds the critical threshold $N^*$, the likelihood of sampling a high-quality output ($\mathcal{L}_{\text{good}}$) dominates, making diversified sampling the statistically superior strategy. 
However, in contrast to exact match at $N$, majority voting fails to yield comparable performance gains from repeated sampling and may even induce degradation in bad scenarios~\cite{wang2025diversified}. 
These theoretical findings align with our empirical observations that EM gains are more pronounced at larger $N$.

\subsection{Perturbation Design in Sampling}
\label{sec:perturbation design}

According to the theoretical insight, we delineate two distinct categories of the perturbation function $\pi_S$ in diversified sampling: (1) Task-agnostic perturbations, including Prefix Padding, Suffix Padding, Middle Insertion, Gaussian Noise, Random Offset, and Missing Data; and (2) Task-specific perturbations, including Task Sensitivity, Task Dependency, and Task Reconstruction.
Task-agnostic perturbations are independent of the target task or dataset, relying on commonly used methods to achieve sample diversification. 
By contrast, task-specific ones are tightly linked to samples in accordance with their inherent temporal structure or task-defined patterns.
Detailed definitions and visualizations of all test-time augmentation variants can be found in Appendix~\ref{app:perturbations}.

\section{The Efforts of Diversified Sampling in Scaling Inference}

Building on the theory in Section~\ref{sec:diversified sampling}, we investigate conditions and mechanisms under which strategies improve performance. We first identify which techniques yield significant gains and then quantify their effect on forecasting accuracy across TSFMs and datasets. In addition, we conduct ablation studies on inference scaling under different perturbations, with detailed results reported in the Appendix~\ref{app:ablation}.

\begin{table*}[t!]
    \centering 
    \caption{Effect of diversified sampling strategies across TSFMs and datasets with 128 trials. The cells highlighted in red represent failed perturbations that result in an MSE greater than 20\% compared to standard sampling. }
    \label{tab:model_dataset_perturbation}
    \label{tab:model_dataset_gray_none}
    \resizebox{0.95\textwidth}{!}{%
        \begin{tabular}{l|cccccccccccc|c}
            \toprule 
            \multirow{2}{*}{\makecell{ \,\textbf{Diversified}\\ \,\textbf{Strategy}}} & \multicolumn{3}{c}{\textbf{\texttt{Chronos-T5-Tiny}}} & \multicolumn{3}{c}{\textbf{\texttt{Moirai-1.1-R-Small}}} & \multicolumn{3}{c}{\textbf{\texttt{TimesFM-2.5-200M}}} & \multicolumn{3}{c|}{\textbf{\texttt{TimeMoE-50M}}} &  \multirow{2}{*}{\makecell{ \,\textbf{Fail}\\ \,\textbf{Cout.}}}\\
            \cmidrule(lr){2-4} \cmidrule(lr){5-7} \cmidrule(lr){8-10} \cmidrule(lr){11-13} 
            & ETTh1 & Traffic & Electricity & ETTh1 & Traffic & Electricity & ETTh1 & Traffic & Electricity & ETTh1 & Traffic & Electricity & \\
             \hline
            \rowcolor{gray!20}
None         &   0.1969$\pm$\tiny0.0056  &  0.2970$\pm$\tiny0.0067  &  0.6318$\pm$\tiny0.0532   &  0.7412$\pm$\tiny0.7458  &  0.1856$\pm$\tiny0.1151  &  0.6457$\pm$\tiny0.1441  &  0.1411$\pm$\tiny0.0000  &   0.0440$\pm$\tiny0.0000   &   0.1887$\pm$\tiny0.0000   &   0.2412$\pm$\tiny0.0000  &  0.3205$\pm$\tiny0.0000  &  0.9099$\pm$\tiny0.0000  &  -  \\
\hline
            Prefix      &  0.1967$\pm$\tiny0.0057 & 0.2970$\pm$\tiny0.0094 & 0.6483$\pm$\tiny0.0833 &  0.6283$\pm$\tiny0.2240 & 0.1940$\pm$\tiny0.1380 & 0.6808$\pm$\tiny0.0705  &  0.1508$\pm$\tiny0.0047 & 0.0437$\pm$\tiny0.0001 & 0.1857$\pm$\tiny0.0012 &  0.2410$\pm$\tiny0.0001 & 0.3210$\pm$\tiny0.0003 & 0.9104$\pm$\tiny0.0005 & 0 \\
Suffix        &  \cellcolor{red!10}    0.8747$\pm$\tiny0.3240   &  \cellcolor{red!10}    1.2307$\pm$\tiny0.7274   &  \cellcolor{red!10}   1.0046$\pm$\tiny0.2605    &  \cellcolor{red!10}     1.2239$\pm$\tiny0.5135  &  \cellcolor{red!10} 1.3229$\pm$\tiny0.8577  &  \cellcolor{red!10} 1.2391$\pm$\tiny0.4795   &  \cellcolor{red!10}    0.3370$\pm$\tiny0.1154   &  \cellcolor{red!10}   1.0543$\pm$\tiny0.8516   &  \cellcolor{red!10}    0.7706$\pm$\tiny0.3820   &     0.2829$\pm$\tiny0.0296   &  \cellcolor{red!10}   1.6084$\pm$\tiny0.8924   &  \cellcolor{red!10}   1.3404$\pm$\tiny0.2703   &  \textbf{11}  \\
Insertion     &    0.1940$\pm$\tiny0.0059   &   0.3174$\pm$\tiny0.0373   &   0.6727$\pm$\tiny0.0953   &  0.6422$\pm$\tiny0.2749  &  \cellcolor{red!10} 0.2581$\pm$\tiny0.1142  &  0.7368$\pm$\tiny0.0591   &    0.1416$\pm$\tiny0.0009   &  \cellcolor{red!10}   0.0577$\pm$\tiny0.0127   &   0.2055$\pm$\tiny0.0156    &    0.2410$\pm$\tiny0.0002   &   0.3227$\pm$\tiny0.0035   &   0.9039$\pm$\tiny0.0060   &  2  \\
Gaussian      &   0.1701$\pm$\tiny0.0068   &   0.2218$\pm$\tiny0.0290   &   0.6315$\pm$\tiny0.0721    &  \cellcolor{red!10}  3.3840$\pm$\tiny3.8760  &  \cellcolor{red!10} 0.2417$\pm$\tiny0.0768  &  \cellcolor{red!10} 0.7909$\pm$\tiny0.1558   &    0.1418$\pm$\tiny0.0013   &   0.0458$\pm$\tiny0.0019   &   0.1931$\pm$\tiny0.0041   &    0.2406$\pm$\tiny0.0018   &   0.3168$\pm$\tiny0.0022   &   0.9071$\pm$\tiny0.0017   &  3  \\
Random        &      0.1937$\pm$\tiny0.0104   &   0.2614$\pm$\tiny0.0177   &   0.6901$\pm$\tiny0.0934   &  \cellcolor{red!10}    3.8157$\pm$\tiny3.4359  &  \cellcolor{red!10} 0.3185$\pm$\tiny0.0844  &  \cellcolor{red!10} 0.9280$\pm$\tiny0.4061    &    0.1398$\pm$\tiny0.0017   &   0.0474$\pm$\tiny0.0020   &   0.1966$\pm$\tiny0.0047   &     0.2389$\pm$\tiny0.0018   &   0.3190$\pm$\tiny0.0017   &   0.9040$\pm$\tiny0.0030    &  3  \\
Missing       &  \cellcolor{red!10}   0.4079$\pm$\tiny0.1268   &  \cellcolor{red!10}   0.3758$\pm$\tiny0.1251   &  \cellcolor{red!10}    0.7877$\pm$\tiny0.1283   &  \cellcolor{red!10}   6.3197$\pm$\tiny24.2453  &  \cellcolor{red!10} 0.4693$\pm$\tiny0.1497  &  \cellcolor{red!10} 1.1945$\pm$\tiny0.4356   &    0.1402$\pm$\tiny0.0026   &  \cellcolor{red!10}   0.0641$\pm$\tiny0.0157   &  \cellcolor{red!10}   0.2897$\pm$\tiny0.0236   &   0.2328$\pm$\tiny0.0022   &   0.3193$\pm$\tiny0.0050   &   0.8525$\pm$\tiny0.0318   &  \textbf{8}  \\
\hline
Sensitivity   &    0.1958$\pm$\tiny0.0051   &   0.3109$\pm$\tiny0.0097   &   0.6612$\pm$\tiny0.0736   &  0.6760$\pm$\tiny0.3648  &  0.1810$\pm$\tiny0.0168  &  0.6582$\pm$\tiny0.0233    &    0.1413$\pm$\tiny0.0002   &   0.0443$\pm$\tiny0.0002   &   0.1887$\pm$\tiny0.0001    &   0.2454$\pm$\tiny0.0016   &   0.3243$\pm$\tiny0.0016   &   0.9251$\pm$\tiny0.0064   &  0  \\
Dependency    &   0.1944$\pm$\tiny0.0058   &   0.3048$\pm$\tiny0.0069   &   0.6827$\pm$\tiny0.1045   &    0.7344$\pm$\tiny0.5802  &  0.1863$\pm$\tiny0.0446  &  0.6505$\pm$\tiny0.0159   &    0.1420$\pm$\tiny0.0006   &   0.0451$\pm$\tiny0.0007   &   0.1893$\pm$\tiny0.0004   &     0.2463$\pm$\tiny0.0022   &   0.3261$\pm$\tiny0.0025   &   0.9280$\pm$\tiny0.0076   &  0  \\
Reconstruction  &    0.1991$\pm$\tiny0.0062  &  0.3409$\pm$\tiny0.1014  &  0.6926$\pm$\tiny0.0719   &   0.6505$\pm$\tiny0.3703  &  \cellcolor{red!10} 0.2346$\pm$\tiny0.1427  &  0.7380$\pm$\tiny0.0439   &  0.1471$\pm$\tiny0.0016  &  0.0433$\pm$\tiny0.0016  &  0.1920$\pm$\tiny0.0022  &  0.2424$\pm$\tiny0.0010  &  0.3064$\pm$\tiny0.0069  &  0.9118$\pm$\tiny0.0050 &  1  \\
            \bottomrule
        \end{tabular}
    }\vspace{-2mm}
\end{table*}

\begin{figure*}[t!]
    \centering
    \begin{subfigure}{0.32\textwidth}
        \includegraphics[width=\linewidth]{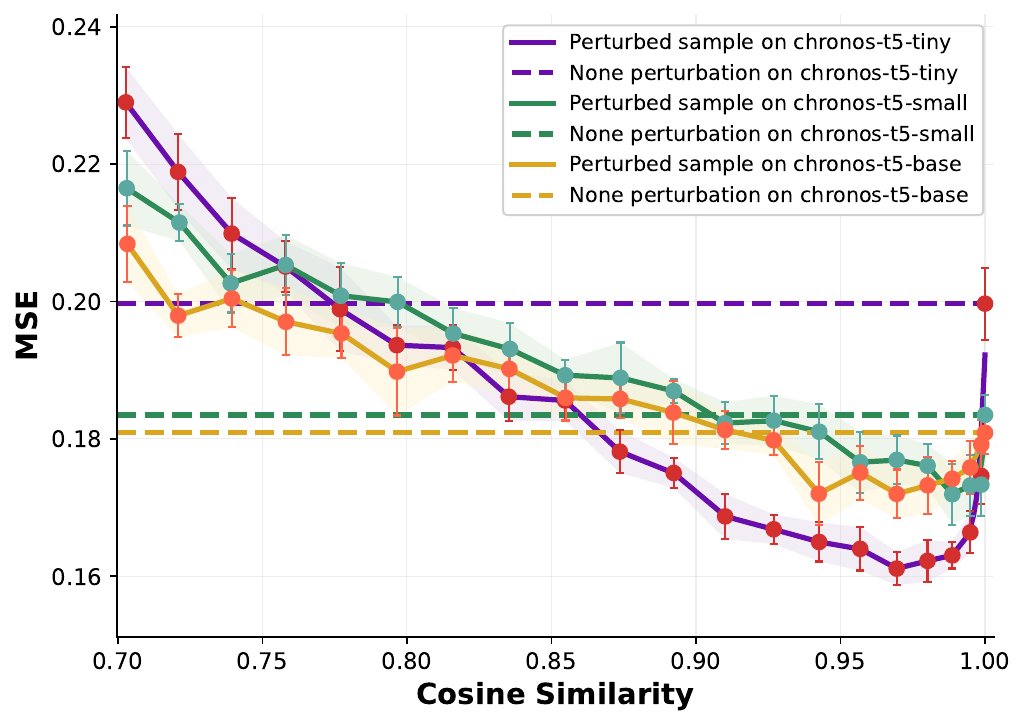}
        \caption{Performance on \textit{ETTh1}}
    \end{subfigure}
    \begin{subfigure}{0.32\textwidth}
        \includegraphics[width=\linewidth]{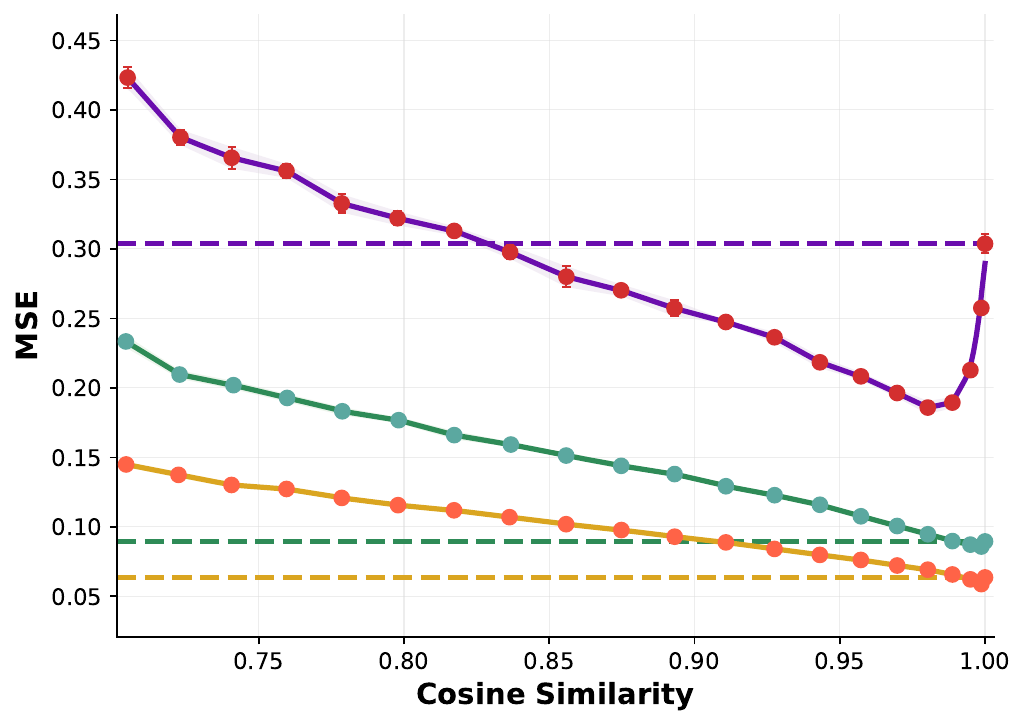}
        \caption{Performance on \textit{Traffic}}
    \end{subfigure}
    \begin{subfigure}{0.32\textwidth}
        \includegraphics[width=\linewidth]{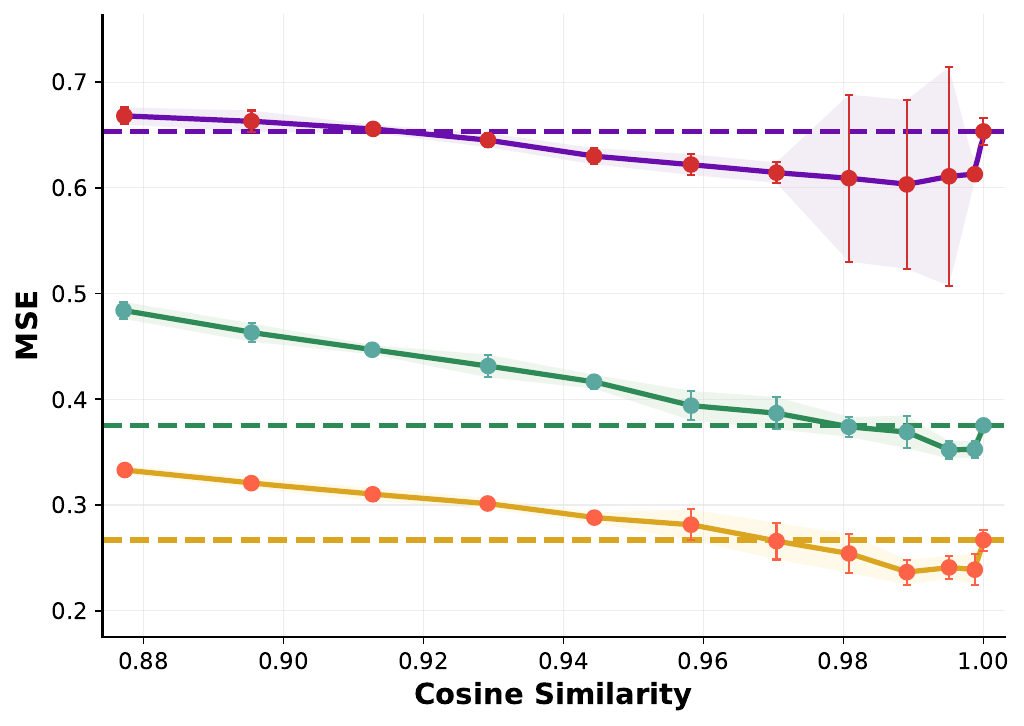}
        \caption{Performance on \textit{Electrictiy}}
    \end{subfigure}
\vskip -0.1in
    \caption{\textbf{Relationship between cosine similarity of time series samples and forecasting performance.} Results are obtained on three datasets using \textit{Chronos-T5} models of different sizes. Solid lines represent performance under diversified sampling, with shaded regions indicating standard deviation across multiple runs. Horizontal lines correspond to the standard sampling for each model.}\vspace{-4mm}
    \label{fig:diversified sampling similarity}
\end{figure*}

\subsection{When Does It Become Effective?}
\label{sec:when}
Diversified sampling fundamentally changes the effective support of the forecasting distribution, its benefit depends on whether the induced diversity expands fidelity or noise. Hence, we explicitly study when it is effective.

\begin{figure*}[t!]
    \centering
    \includegraphics[width=0.97\linewidth]{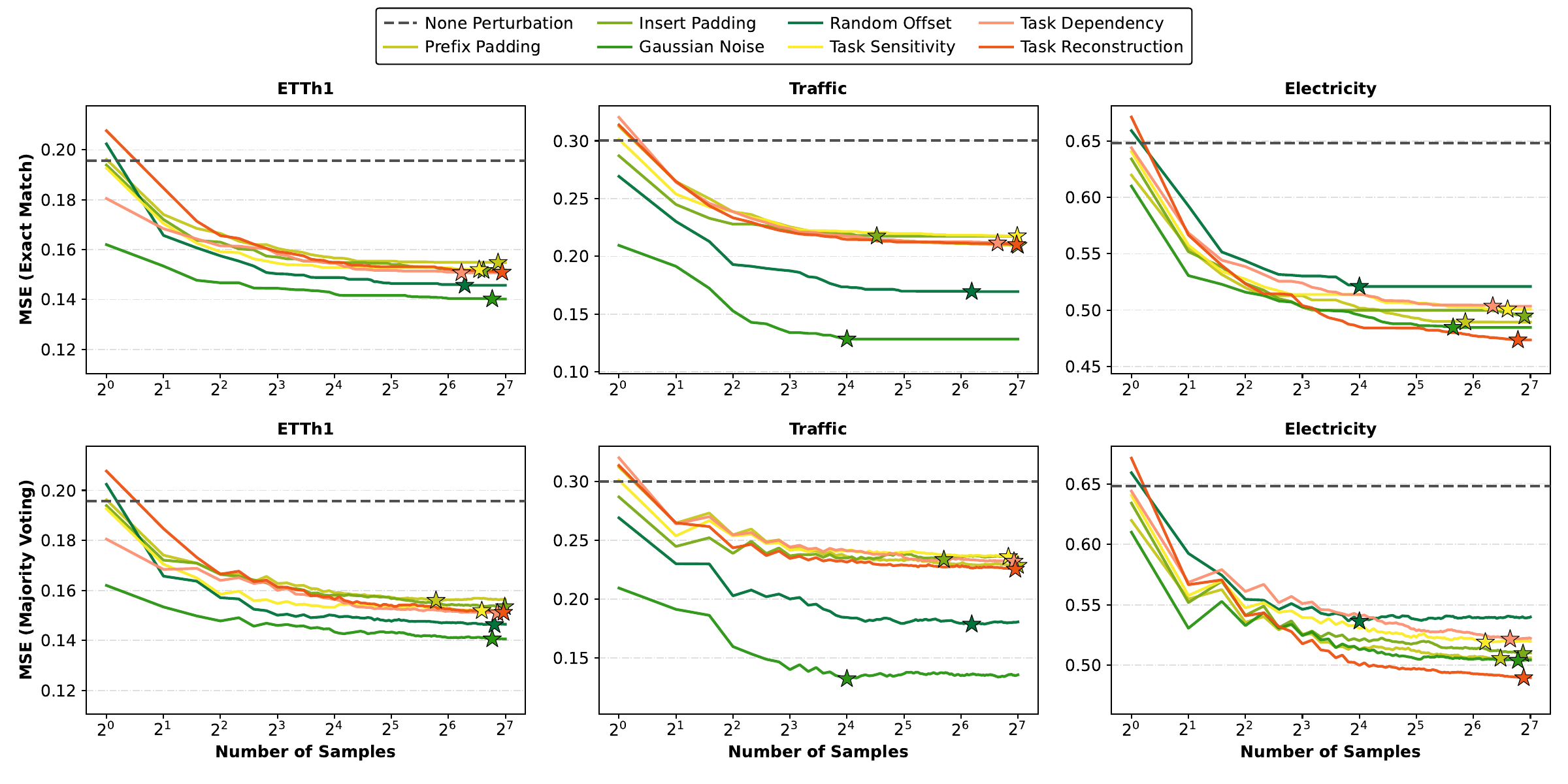}
        \vspace{-6.5mm}
    \caption{\textbf{Performance of \textit{Chronos} on different datasets with effective perturbations.} The dashed line represents the error of a single standard sampling, and results are reported under both EM and MV aggregators.}\vspace{-6mm}
    \label{fig:task}
\end{figure*}

\paragraph{Some Diversified Strategies Often Fail} 
Diversified sampling does not always improve inference performance, its effectiveness depends on how randomness interacts with model structure and task characteristics~\cite{holtzman2019curious,ahmed2025intentfactoredgenerationunleashing,wang2025diversified}. We therefore evaluate the perturbations in Section~\ref{sec:perturbation design} across multiple TSFMs by measuring their performance changes relative to non-diversified baselines. We define failures as cases where performance drops by more than $20\%$. The results in Table~\ref{tab:model_dataset_perturbation} show that unlike language models in NLP tasks, \textit{Suffix} and \textit{Missing} perturbations incur the highest failure rates and often provide little benefit or even degrade performance, suggesting that they introduce noise rather than useful diversity. Accordingly, we discard these strategies and restrict subsequent analysis to the validated set $\pi_S \in \Pi_{\mathrm{valid}}$.

\paragraph{The Trade-off of Fidelity-Diversity}
We further examine when diversified sampling can effectively improve inference performance. 
Introducing diversification is beneficial only if the induced predictive distribution increases probability mass near $\mathbf{Y}$, while preserving sufficient fidelity to the original input. 
This reveals a fundamental trade-off: overly conservative perturbations fail to expand the predictive distribution, whereas overly aggressive ones increase diversity at the cost of degraded sample quality, leading to a worse $N^*$ limit. Consequently, diversified sampling is effective only within a limited regime where perturbations provide sufficient exploration without compromising fidelity.
Empirically, this trade-off is illustrated in Figure~\ref{fig:diversified sampling similarity}. Across all three datasets, diversified sampling performance exhibits a clear non-monotonic relationship with perturbation relevance: MSE decreases as relevance increases, reaches an optimal minimum, and then slightly rises. Notably, the best performance consistently occurs within a narrow relevance band, approximately between cosine similarity 0.95 and 1.00, confirming the fidelity–diversity trade-off. Beyond this global trend, diversified sampling also induces qualitative changes in relative model performance. On ETTh1, a representative pattern emerges in which smaller models exhibit higher sensitivity but greater upside under diversified sampling, illustrating that properly balanced diversification can reverse relative performance ordering across models.

\begin{tcolorbox}[colback=orange!10, colframe=orange, boxrule=0.5pt, sharp corners, width=\linewidth, top=2pt, bottom=2pt]
\small
\includegraphics[width=0.03\linewidth]{pic/pin.png}  \textit{\textbf{Takeaways:} Diversified sampling works only if perturbations expand the predictive support while preserving fidelity. Excessive perturbations will elevate harmful sample probability, raising the critical budget. So it is necessary to design a good perturbation.}
\normalsize
\end{tcolorbox}

\subsection{How Effective Is It?}
We then investigate how diversified sampling affects forecasting performance in practice, focusing on its interaction with inference scaling across datasets and model sizes. Experiments are conducted on multiple TSFMs with different perturbations. The results of \textit{Chronos} are in Figure~\ref{fig:task}, and other models' results are reported in Appendix~\ref{app:how}.

\vspace{-3mm}
\paragraph{Performance Scaling under Baseline and Diversified Inference}
We first examine how forecasting performance evolves with increasing inference budget. As shown in Figure~\ref{fig:task}, the sampling temperature is fixed to $0.7$ for models, and the number of generated samples is varied from $2^0$ to $2^7$ to control the inference budget.
The standard sampling serves as a fixed reference, since direct inference produces only a single prediction and therefore does not benefit from inference scaling.
In contrast, introducing input perturbations enables diversified sampling and yields clear inference scaling effects. For perturbed inputs, the forecasting performance consistently decreases as the sampling budget grows and converges to levels superior to the standard sampling, with peak improvements of up to approximately $50\%$. However, the relative effectiveness of different perturbations varies across models, indicating that inference scaling gains depend critically on the alignment between perturbation design and model characteristics. In conclusion, diversified sampling is effective but not universally beneficial, and must be tailored to the underlying model and inference setup.

\vspace{-4mm}
\paragraph{Effect of Aggregation and Model-Datasets Interaction}
We further analyze how aggregation functions and TSFM characteristics influence the effectiveness of diversified inference. All perturbation strategies are evaluated under identical inference budgets using both EM and MV. While both aggregators preserve the overall scaling trend induced by diversified sampling, EM consistently yields smoother scaling behavior and lower optimal errors.
Beyond aggregation, the gains from diversified inference vary systematically across models and datasets. Models with stronger standard sampling performance generally achieve lower absolute errors. However, compared to models equipped with decoding strategies, models without decoding strategies (e.g., TimesFM and TimeMoE) exhibit relatively weaker inference scaling effects, although diversified sampling under most perturbations still improves upon standard sampling.

\vspace{-1.5mm}
\begin{tcolorbox}[colback=orange!10, colframe=orange, boxrule=0.5pt, sharp corners, width=\linewidth, top=2pt, bottom=2pt]
\small
\includegraphics[width=0.03\linewidth]{pic/pin.png}  \textit{\textbf{Takeaways:} Inputting perturbations alone will only expand the support, while incorporating scaling inference increases the hit probability, enabling diversified sampling to surpass the finite sample and achieve improvements.
}
\normalsize
\end{tcolorbox}
\vspace{-0.5mm}

\begin{wrapfigure}{br}{0.31\textwidth}
\begin{center}
\vspace{-7mm}
\setlength{\columnsep}{1pt} 
\centerline{\includegraphics[width=\linewidth]{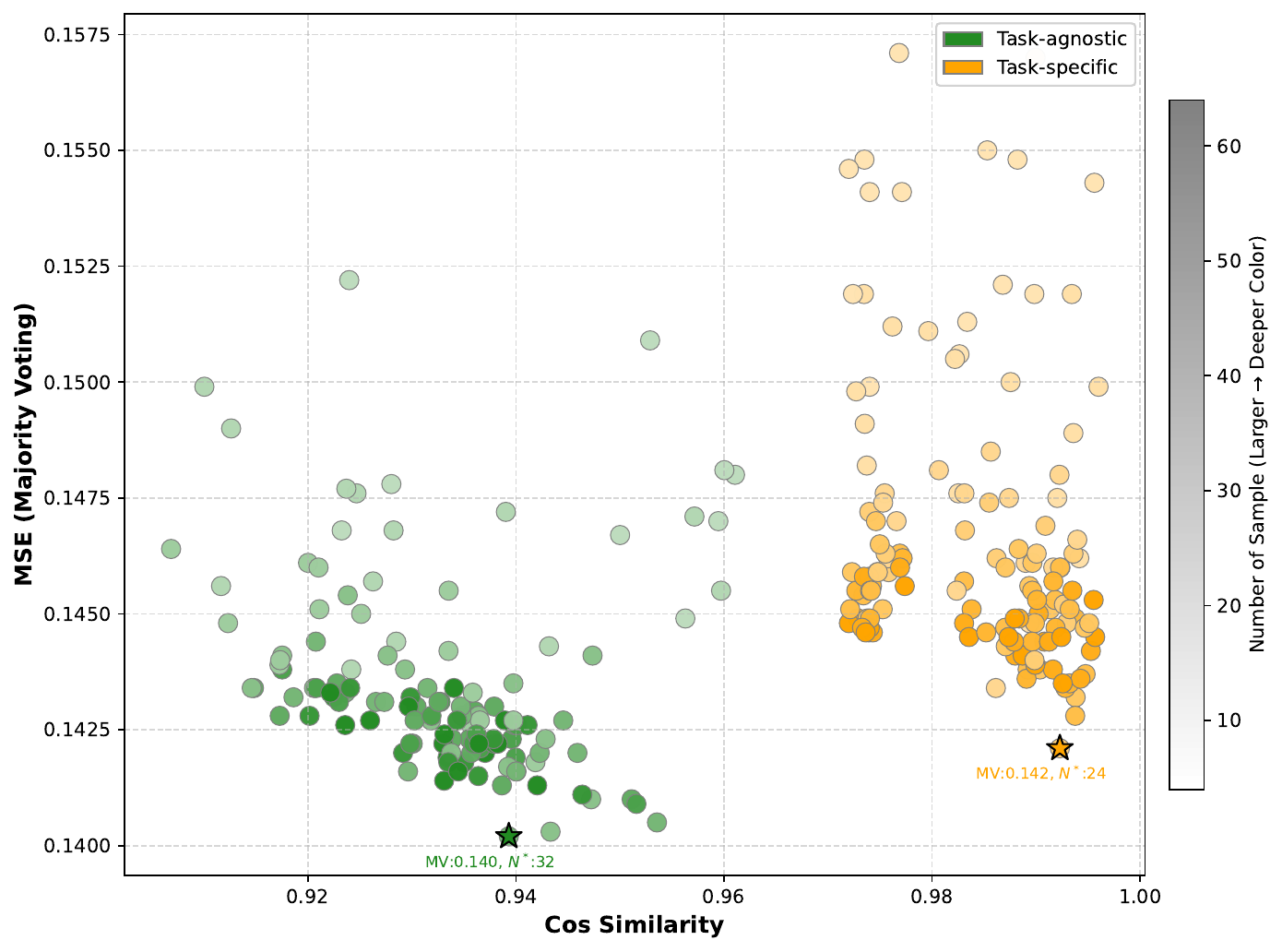}}
\vspace{-5.3mm}
 \caption{Computation-Diversity.}
    \label{fig:tsfm_sampling}
\vskip -1.02cm
\end{center}
\end{wrapfigure}

\textbf{Actionable Insights}\,
Combining ``When"~and~``How", we derive diversified sampling should be configured through an explicit coupling between inference computation and diversity, as an example in Figure~\ref{fig:tsfm_sampling}. 
Operationally, given an inference budget $N$: Introduce perturbations to enable inference scaling, while restricting them to high-relevance variants that preserve strong similarity to the original input (e.g., cosine similarity $\geq$ 0.92), ensuring that diversification expands predictive support without overwhelming sample fidelity. We suggest using noise-based perturbations, as MV typically decreases with increasing $N$ and converges to a stable level (often around $N=2^5$ for MV), beyond which additional sampling yields diminishing returns and can be safely terminated. Aggregation provides an effective validity check: if aggregated performance fails to improve as $N$ increases, the current perturbation is misaligned with the inference budget and should be adjusted.

\begin{table*}[!t]
\centering 
\caption{RobustMSE evaluations of TSFMs under diversified scaling inference ($N=64$) with EM and MV. The MSE is a reference.}
\label{tab:robustmse}
\resizebox{0.96\textwidth}{!}{%
\begin{tabular}{llcccccccccccc}
\toprule
\multirow{2}{*}{\textbf{Strategy}} & \multirow{2}{*}{\textbf{Dataset}} & \multicolumn{3}{c}{\textbf{\texttt{Chronos-T5-Base}}} & \multicolumn{3}{c}{\textbf{\texttt{Moirai-1.1-R-Base}}} & \multicolumn{3}{c}{\textbf{\texttt{TimesFM-2.5-200M}}} & \multicolumn{3}{c}{\textbf{\texttt{TimeMoE-200M}}} \\
\cmidrule(lr){3-5} \cmidrule(lr){6-8} \cmidrule(lr){9-11} \cmidrule(lr){12-14}
 & & {EM} & {MV} & {MSE} & {EM} & {MV} & {MSE} & {EM} & {MV} & {MSE}  & {EM} & {MV}  & {MSE}  \\
\midrule

\multirow{4}{*}{Task-Agnostic} 
 & ETTh1       & 0.1394$\pm$\tiny0.0003& 0.1424$\pm$\tiny0.0005& 0.1805$\pm$\tiny0.0068 & \underline{0.1383}$\pm$\tiny0.0004& \underline{0.1407}$\pm$\tiny0.0020& \underline{0.1458}$\pm$\tiny0.0075 & \textbf{0.1334}$\pm$\tiny0.0003& \textbf{0.1354}$\pm$\tiny0.0006& \textbf{0.1435}$\pm$\tiny0.0053 & 0.2353$\pm$\tiny0.0003& 0.2396$\pm$\tiny0.0002  &  0.2401$\pm$\tiny0.0015 \\
 & Traffic     & 0.0468$\pm$\tiny0.0004& 0.0467$\pm$\tiny0.0004 & \underline{0.0669}$\pm$\tiny0.0449 & \underline{0.0438}$\pm$\tiny0.0015& \textbf{0.0431}$\pm$\tiny0.0004 &  0.0670$\pm$\tiny0.0699 & \textbf{0.0437}$\pm$\tiny0.0000& \underline{0.0439}$\pm$\tiny0.0002 & \textbf{0.0481}$\pm$\tiny0.0066 & 0.3172$\pm$\tiny0.0017& 0.3227$\pm$\tiny0.0008  &  0.3232$\pm$\tiny0.0131 \\
 & Electricity &   0.2072$\pm$\tiny0.0013& 0.2088$\pm$\tiny0.0026 & 0.2645$\pm$\tiny0.0375 & \underline{0.1945}$\pm$\tiny0.0002& \underline{0.1953}$\pm$\tiny0.0002 & \underline{0.2141}$\pm$\tiny0.0325 & \textbf{0.1845}$\pm$\tiny0.0007& \textbf{0.1852}$\pm$\tiny0.0003 &  \textbf{0.1947}$\pm$\tiny0.0101 & 0.8968$\pm$\tiny0.0031& 0.9057$\pm$\tiny0.0012  & 0.9068$\pm$\tiny0.0145 \\
 & \cellcolor{gray!20} \, Average 
 & \cellcolor{gray!20} 0.1311$\pm$\tiny0.0007& \cellcolor{gray!20} 0.1326$\pm$\tiny0.0012 &\cellcolor{gray!20} 0.1706$\pm$\tiny0.0297 & \cellcolor{gray!20} \underline{0.1255}$\pm$\tiny0.0007& \cellcolor{gray!20} \underline{0.1264}$\pm$\tiny0.0009 &\cellcolor{gray!20} \underline{0.1423}$\pm$\tiny0.0366 & \cellcolor{gray!20} \textbf{0.1205}$\pm$\tiny0.0003& \cellcolor{gray!20} \textbf{0.1215}$\pm$\tiny0.0004 &\cellcolor{gray!20} \textbf{0.1288}$\pm$\tiny0.0073 & \cellcolor{gray!20} 0.4831$\pm$\tiny0.0017& \cellcolor{gray!20} 0.4893$\pm$\tiny0.0007  & \cellcolor{gray!20} 0.4900$\pm$\tiny0.0097  \\
 \hline

\multirow{4}{*}{Task-Specific} 
 &  ETTh1       & 0.1416$\pm$\tiny0.0004& 0.1445$\pm$\tiny0.0004 & 0.1843$\pm$\tiny0.0055 & \underline{0.1413}$\pm$\tiny0.0003& \underline{0.1417}$\pm$\tiny0.0002 & \underline{0.1478}$\pm$\tiny0.0021 & \textbf{0.1401}$\pm$\tiny0.0003& \textbf{0.1411}$\pm$\tiny0.0000 &  \textbf{0.1435}$\pm$\tiny0.0028 & 0.2390$\pm$\tiny0.0008& 0.2435$\pm$\tiny0.0002 &  0.2448$\pm$\tiny0.0023 \\
 & Traffic   & 0.0497$\pm$\tiny0.0002& 0.0501$\pm$\tiny0.0002 & 0.0725$\pm$\tiny0.0160 & \textbf{0.0420}$\pm$\tiny0.0001& \textbf{0.0422}$\pm$\tiny0.0001 &  \textbf{0.0490}$\pm$\tiny0.0153 & \underline{0.0441}$\pm$\tiny0.0002& \underline{0.0444}$\pm$\tiny0.0001 & \underline{0.0500}$\pm$\tiny0.0087 & 0.3112$\pm$\tiny0.0019& 0.3271$\pm$\tiny0.0001  & 0.3244$\pm$\tiny0.0100 \\
  & Electricity & 0.2084$\pm$\tiny0.0009& 0.2139$\pm$\tiny0.0015 &  0.3157$\pm$\tiny0.0750 & \underline{0.1919}$\pm$\tiny0.0002& \underline{0.1920}$\pm$\tiny0.0000 &  \textbf{0.2027}$\pm$\tiny0.0160 & \textbf{0.1881}$\pm$\tiny0.0003& \textbf{0.1887}$\pm$\tiny0.0000 &  \underline{0.2054}$\pm$\tiny0.0216 & 0.9028$\pm$\tiny0.0053& 0.9248$\pm$\tiny0.0009  & 0.9275$\pm$\tiny0.0107 \\
 & \cellcolor{gray!20} \, Average 
 & \cellcolor{gray!20} 0.1332$\pm$\tiny0.0005& \cellcolor{gray!20} 0.1362$\pm$\tiny0.0007 &\cellcolor{gray!20} 0.1908$\pm$\tiny0.0322 & \cellcolor{gray!20} \underline{0.1251}$\pm$\tiny0.0002& \cellcolor{gray!20} \underline{0.1253}$\pm$\tiny0.0001 &\cellcolor{gray!20} \underline{0.1332}$\pm$\tiny0.0111 & \cellcolor{gray!20} \textbf{0.1241}$\pm$\tiny0.0002& \cellcolor{gray!20} \textbf{0.1247}$\pm$\tiny0.0000 &\cellcolor{gray!20} \textbf{0.1330}$\pm$\tiny0.0110 & \cellcolor{gray!20} 0.4843$\pm$\tiny0.0027& \cellcolor{gray!20} 0.4985$\pm$\tiny0.0004 & \cellcolor{gray!20} 0.4989$\pm$\tiny0.0077 \\
 \hline

\rowcolor{red!10}
\multicolumn{2}{c}{\textbf{1\textsuperscript{st} Count}} & 0 & 0 & 0  & 1 & 2  & 2 & 5 & 4  & 4 & 0 & 0  & 0 \\

\bottomrule
\end{tabular}
}\vspace{-3mm}
\end{table*}

\section{RobustMSE: A New Metric}

\paragraph{Why RobustMSE?} Standard time series forecasting typically relies on deterministic metrics (e.g., MSE, MAE), assessed via a single inference pass~\cite{hyndman2006another}. While sufficient for discriminative models, this paradigm is inadequate for generative TSFMs operating under stochastic settings. 
In the inference, prediction quality is not defined by a single realization, but by the distribution of candidate sequences and decoding strategies~\cite{holtzman2019curious,chen2021evaluating}. 
Consequently, single-run metrics exhibit high variance due to stochasticity and fail to capture the model's potential when afforded a computational budget for exploration.
Based on our actionable insights, we introduce \textit{RobustMSE}, a sampling-based metric that reflects a TSFM’s repeatable best achievable performance under a fixed inference budget. 
Unlike MSE, which often reflects a lower-bound performance tied to unoptimized decoding, RobustMSE aggregates the headroom achievable under validated sampling strategies. 
This provides a fair comparison by measuring how effectively a TSFM can leverage diversified scaling inference to minimize error.

\vspace{-1mm}
\paragraph{Computation of RobustMSE} Following the empirical findings in Section~\ref{sec:when}, we exclude ineffective perturbations and restrict evaluation to a critical sample size. 
For each model on varying datasets, we fix the number of samples $N=64$ and perform diversified inference under all remaining effective perturbations.
In each trial, we record the EM and MV across all valid diversified configurations, and RobustMSE is computed as the average of $T=5$ trials:\vspace{-2mm}
\begin{equation}
    \mathbf{RobustMSE}_\texttt{EM|MV} = \frac{1}{T}\sum_{i=1}^T\operatorname{MSE}(\hat{\mathbf{Y}}^i_{\texttt{EM|MV}}, \mathbf{Y}).
\end{equation}

\vspace{-3mm}
\paragraph{Horizontal Comparison} 
Table~\ref{tab:robustmse} reports the RobustMSE performance of mainstream TSFMs on approximately 200M active parameters across various datasets. 
These results confirm that TSFMs effectively utilize the headroom provided by diversified sampling to minimize prediction error.
Overall, TimesFM demonstrates superior performance, achieving the lowest RobustMSE in 5 out of 6 comparison scenarios, as it was pre-trained on some datasets.
Moirai serves as a highly competitive runner-up, consistently ranking second and exhibiting exceptional stability.
In contrast, TimeMoE exhibits significantly higher variance and error margins compared to other transformer-based baselines in this evaluation, indicating potential limitations in leveraging diversified inference budgets.
Furthermore, we note that the standard deviation of RobustMSE has always been in a small state compared to MSE, indicating the stability of the metric.

\vspace{-3pt}
\section{Related Work}
The recent advancement of TSFMs have demonstrated the efficacy of large-scale pre-training for cross-domain temporal representation learning~\cite{li2025tsfm,ansari2024chronoslearninglanguagetime,woo2024unified}. However, while the NLP community has successfully leveraged scaling inference, increasing test-time computation via deeper search to enhance performance without parameter updates~\cite{wu2024inference,wang2025diversified,vijayakumar2016diverse}, such strategies remain underexplored in time series. Unlike existing works that focus on training scaling laws~\cite{liang2024foundation,kottapalli2025foundation}, we systematically investigate inference-time scaling to improve forecasting accuracy and robustness in TSFMs. Due to space limitations, we put the comprehensive review in Appendix~\ref{app:related_work}.

\vspace{-3pt}
\section{Conclusion}
In this work, we systematically explore the underexplored inference scaling and diversified sampling for TSFMs.
By dissecting the effect of standard sampling on performance, we uncover counterintuitive findings: traditional scaling levers such as increasing TSFM parameters or extending length yield inconsistent, dataset-dependent gains, with their effects often overshadowed by the benefits of aggregating multiple inferences. 
To address the inherent limitations, we further formalize a theoretical framework for diversified sampling, proving that input perturbations expand the model’s solution space. 
Guided by this theory, we design perturbation schemes and conduct extensive evaluations across TSFM architectures and real-world datasets, quantifying consistent gains in multi-step forecasting accuracy and robustness.
As a practical instantiation, we introduce RobustMSE, a novel evaluation metric quantifying the headroom of TSFMs, enabling fair comparisons of models under fixed compute budgets. 
Collectively, our findings highlight inference design as a pivotal yet underutilized lever for optimizing TSFM performance, offering a cost-effective alternative to resource-heavy pre-training or finetuning pipelines.

\newpage
\section*{Impact Statements}
This paper presents work whose goal is to advance the field of machine learning. There are many potential societal consequences of our work, none of which we feel must be specifically highlighted here.

\bibliography{reference}
\bibliographystyle{icml2025}

\newpage
\appendix
\onecolumn

\section{Related Work}
\label{app:related_work}

\textbf{Time Series Foundation Model}~(TSFM) has established a pre-training paradigm that learns domain-agnostic temporal dynamics from large-scale and heterogeneous time series, and recent progress has demonstrated that such large-scale pre-training enables models to acquire cross-domain temporal representations across diverse data sources~\cite{li2025tsfm,ansari2024chronoslearninglanguagetime,liang2024foundation,woo2024unified,shi2024time}. 
Existing TSFM architectures can be grouped into decoder-based and encoder-based formulations. Decoder-based models, such as Chronos~\cite{ansari2024chronoslearninglanguagetime}, TimesFM~\cite{das2024decoder}, and TimeMoE~\cite{shi2024time}, treat forecasting as conditional language modeling, using normalization, symbolization, and sparse expert mixtures to support multi-step forecasting and cross-domain generalization.
Encoder-based models, such as Moirai~\cite{woo2024unified}, learn unified representations across variables, frequencies, and tasks through flexible attention mechanisms. Additional developments~\cite{garza2023timegpt,goswami2024moment,wang2024timexer} further expand the capabilities of TSFMs. Across these architectures, the common strategy is to leverage large-scale pre-training to acquire transferable temporal patterns and improve robustness and generalization in downstream tasks.

\textbf{Scaling Inference} improves language models' performance by increasing test time computation via deeper decoding, broader search, and lightweight aggregation without changing model parameters~\cite{wu2024inference,wang2025diversified}. 
Prior studies have shown that multi-pass reasoning~\cite{wei2022chain,yao2023tree} and test-time ensembling~\cite{vijayakumar2016diverse,wang2022self} can substantially enhance model performance without additional training.
For example, these approaches implement deeper decoding or stochastic constraint relaxation by extending chain- or tree-structured searches~\cite{wei2022chain,yao2023tree}, promoting sampling to generate different candidate sequences through randomness and diversity~\cite{holtzman2019curious,vijayakumar2016diverse} to expand alternative inference paths.
However, in time series, although existing work has studied training scaling laws~\cite{liang2024foundation,kottapalli2025foundation,yao2024towards}, the inference phase remains unchanged and lacks expanding computational depth or breadth.
This gap motivates our systematic investigation of scaling inference, where we examine temporal expansion strategies and evaluate their effects on prediction accuracy, robustness, and generalization.

\textbf{Diversified Sampling} provides a mechanism to enhance model robustness and uncertainty handling by generating multiple heterogeneous outputs during inference~\cite{vijayakumar2016diverse}. 
Such strategies have been extensively explored in generative modeling~\cite{wang2025diversified,sun2024fast,wang2022self} where multi-path generation and voting-based aggregation can substantially improve reliability in reasoning and math tasks. 
Similarly, existing time series robustness studies primarily focus on assessing model sensitivity through input perturbations~\cite{cheng2024robusttsftheorydesignrobust,tang2024perturbation}. 
These methods capture signal level variability but lack systematic perturbation studies, and leave unexplored whether diversifying samples under fixed model parameters can enhance forecasting accuracy or robustness. 
Consequently, current time series inference frameworks remain unable to fully exploit the potential benefits of diversified sampling~\cite{yao2024towards,kottapalli2025foundation}. 
In this work, we incorporate diversified sampling mechanisms in time series inference to expand the feasible solution space and enhance prediction performance under both task-specific and task-agnostic perturbations.

\section{Omitted Proofs}\label{proof}

\subsection{Asymptotic Analysis ($N \to \infty$)}
We first characterize the theoretical lower bound of the loss when the sample budget approaches infinity. This analysis relies on the support sets of the respective distributions.

\begin{assumption}[\textbf{Identity Inclusion}]
\label{ass:inclusion}
The perturbation distribution $Q(\mathbf{X}' | \mathbf{X}, \pi_S)$ assigns non-zero probability mass (or density) to the original input $\mathbf{X}$, or equivalently, to an $\epsilon$-neighborhood of $\mathbf{X}$.
\end{assumption}

Let $\mathcal{S}_{\text{std}} = \text{supp}(P(\cdot | \mathbf{X}, \zeta))$ and $\mathcal{S}_{\text{div}} = \text{supp}(P_{\text{mix}})$ denote the support sets of the standard and diversified distributions, respectively. Under Assumption~\ref{ass:inclusion}, it logically follows that $\mathcal{S}_{\text{std}} \subseteq \mathcal{S}_{\text{div}}$.

\begin{proposition}[\textbf{Asymptotic Lower Bound}]
As the sample size $N \to \infty$, the expected minimum EM of diversified sampling is strictly bounded by that of standard sampling:
\begin{equation}
    \lim_{N \to \infty} \texttt{EM}_N^{\text{div}} \le \lim_{N \to \infty} \texttt{EM}_N^{\text{std}}.
\end{equation}
The inequality holds strictly ($<$) if there exists $\mathbf{Y}^* \in \mathcal{S}_{\text{div}} \setminus \mathcal{S}_{\text{std}}$ such that $\mathcal{L}(\mathbf{Y}^*, \mathbf{Y}) < \inf_{\hat{\mathbf{Y}} \in \mathcal{S}_{\text{std}}} \mathcal{L}(\hat{\mathbf{Y}}, \mathbf{Y})$.
\end{proposition}

\textit{Proof.} According to extreme value theory, the minimum statistic of i.i.d. samples converges to the essential infimum of the distribution's support. Thus:
\begin{align}
    \lim_{N \to \infty} \texttt{EM}_N^{\text{std}} &= \inf_{\hat{\mathbf{Y}} \in \mathcal{S}_{\text{std}}} \mathcal{L}(\hat{\mathbf{Y}}, \mathbf{Y}), \\
    \lim_{N \to \infty} \texttt{EM}_N^{\text{div}} &= \inf_{\hat{\mathbf{Y}} \in \mathcal{S}_{\text{div}}} \mathcal{L}(\hat{\mathbf{Y}}, \mathbf{Y}).
\end{align}
Since $\mathcal{S}_{\text{std}} \subseteq \mathcal{S}_{\text{div}}$, and the infimum operator is monotonic (i.e., $\inf_{\mathbf{X} \in A} f(\mathbf{X}) \ge \inf_{\mathbf{X}' \in B} f(\mathbf{X}')$ if $A \subseteq B$), the proposition holds. Geometrically, this implies diversified sampling explores an expanded manifold that may contain solutions closer to the ground truth $\mathbf{Y}$, which are inaccessible under the fixed configuration $\zeta$. \hfill $\square$

\subsection{Finite Sample Analysis ($N < \infty$)}

In the finite regime, simply expanding the support is insufficient due to the variance introduced by perturbations. We must balance the \textit{potential for lower loss} against the \textit{probability of sampling it}. We derive a critical sample size $N^*$ using a simplified probabilistic model.

\begin{proposition}[\textbf{Critical Sample Threshold}]
Let $\mathcal{L}_0$ be the baseline EM of standard sampling at one time. Assume the diversified distribution yields a reduced loss $\mathcal{L}_{\text{good}} < \mathcal{L}_0$ with probability $\rho$ (valid perturbation), and an increased loss $\mathcal{L}_{\text{bad}} > \mathcal{L}_0$ with probability $1-\rho$.
Diversified sampling outperforms standard sampling in expectation (i.e., $\mathbb{E}[\texttt{EM}_N^{\text{div}}] < \mathcal{L}_0$) if and only if the sample size $N$ exceeds a critical threshold $N^*$:
\begin{equation}
    N > N^* = \frac{\ln \left( \frac{\mathcal{L}_{\text{bad}} - \mathcal{L}_{\text{good}}}{\mathcal{L}_0 - \mathcal{L}_{\text{good}}} \right)}{\ln \left( \frac{1}{1-\rho} \right)}.
\end{equation}
\end{proposition}

\textit{Proof.}
The expected minimum EM for diversified sampling is the weighted sum of the loss when at least one ``good" sample is found, and the loss when all samples are ``bad":
\begin{equation}
    \mathbb{E}[\texttt{EM}_N^{\text{div}}] = \underbrace{(1-\rho)^N}_{\text{All failure}} \mathcal{L}_{\text{bad}} + \underbrace{\left[1 - (1-\rho)^N\right]}_{\text{At least one success}} \mathcal{L}_{\text{good}}.
\end{equation}
We seek the condition where $\mathbb{E}[\texttt{EM}_N^{\text{div}}] < \mathcal{L}_0$. Substituting the expectation:
\begin{equation}
    (1-\rho)^N \mathcal{L}_{\text{bad}} + \mathcal{L}_{\text{good}} - (1-\rho)^N \mathcal{L}_{\text{good}} < \mathcal{L}_0,
\end{equation}
grouping terms by $(1-\rho)^N$:
\begin{equation}
    (1-\rho)^N (\mathcal{L}_{\text{bad}} - \mathcal{L}_{\text{good}}) < \mathcal{L}_0 - \mathcal{L}_{\text{good}}.
\end{equation}
Then, dividing by $(\mathcal{L}_{\text{bad}} - \mathcal{L}_{\text{good}})$ (which is positive) and taking the natural logarithm:
\begin{equation}
    N \ln(1-\rho) < \ln \left( \frac{\mathcal{L}_0 - \mathcal{L}_{\text{good}}}{\mathcal{L}_{\text{bad}} - \mathcal{L}_{\text{good}}} \right).
\end{equation}
Since $\ln(1-\rho) < 0$, dividing by it reverses the inequality:
\begin{equation}
    N > \frac{\ln \left( \frac{\mathcal{L}_{\text{bad}} - \mathcal{L}_{\text{good}}}{\mathcal{L}_0 - \mathcal{L}_{\text{good}}} \right)}{-\ln(1-\rho)} = \frac{\ln \left( \frac{\mathcal{L}_{\text{bad}} - \mathcal{L}_{\text{good}}}{\mathcal{L}_0 - \mathcal{L}_{\text{good}}} \right)}{\ln \left( \frac{1}{1-\rho} \right)}=N^*.
\end{equation}
This concludes the proof. \hfill $\square$

\section{The Definition of Diversified Strategies}
\label{app:perturbations}
\textbf{Task-Agnostic Perturbations.} 
This category introduces controlled perturbations to raw data sequences without accounting for the model’s designated forecasting objective. 
The primary goals of these methods are to assess model robustness and the impact of input diversity.
Task-agnostic perturbations are further partitioned into two orthogonal subclasses:  (i) \textit{Structural Perturbations}, encompassing Prefix Padding, Suffix Padding, and Middle Insertion, which tamper with the boundary constraints or internal topological structure of the sequence. Notably, the degree of diversity induced by such perturbations exhibits a positive correlation with the perturbation length~\cite{iwana2021empirical}.
(ii) \textit{Noise Perturbations} which cover Gaussian Noise, Random Offset, and Missing Data implantation, where the intensity of perturbation  $\eta$ serves as the direct control variable for tuning the level of sample diversity. This suite is widely recognized as a standard benchmark for evaluating model robustness in time series analysis~\cite{cheng2024robusttsftheorydesignrobust}.

\textbf{Task-Specific Perturbations.} 
To probe the model’s prediction mechanism and simulate realistic contextual distribution shifts, we design three dedicated perturbation injection paradigms tailored to the inherent characteristics of time series tasks: (i) \textit{Task Sensitivity} Perturbation mimics latent time distribution drifts by systematically modulating the global trend of the input time series. This perturbation strategy aligns with real-world scenarios where temporal data often undergoes gradual trend shifts due to external environmental factors~\cite{gama2014survey}.
(ii) \textit{Task Dependency} Perturbation embeds a predefined, task-indicative pattern into the original sequence.  The core objective of this perturbation is to expose potential patterns associated with the exploitation of spurious task-specific features~\cite{yang2025time}.
(iii) \textit{Task Reconstruction} leverages a pre-specified TSFM to reconstruct time series $\mathbf{X}'\sim f(\cdot|\mathbf{X}, \zeta)$ under a given configuration  $\zeta$.
By sampling from the reconstructed time series, this paradigm generates inputs that are semantically congruent with the original sequence yet structurally divergent, thereby augmenting the diversity of the sample pool~\cite{yoon2019time,rasul2021autoregressive}.

While the definitions above provide a structural overview, the core motivation for this taxonomy lies in the fundamental trade-off between universality and fidelity in inference scaling. Task-Agnostic strategies serve as a cost-effective, universal baseline, allowing us to quantify the model's inherent robustness against pure stochasticity without assuming any domain priors. In contrast, Task-Specific strategies are designed to ensure semantic consistency. By leveraging intrinsic temporal properties (such as trend and seasonality), they constrain the exploration within a plausible manifold. This distinction is essential for our analysis, as it enables us to isolate whether performance gains stem merely from tolerance to random noise or from structurally valid exploration that aligns with the time series' physical nature.

\begin{figure}[!t]
    \centering
    \includegraphics[width=0.9\linewidth]{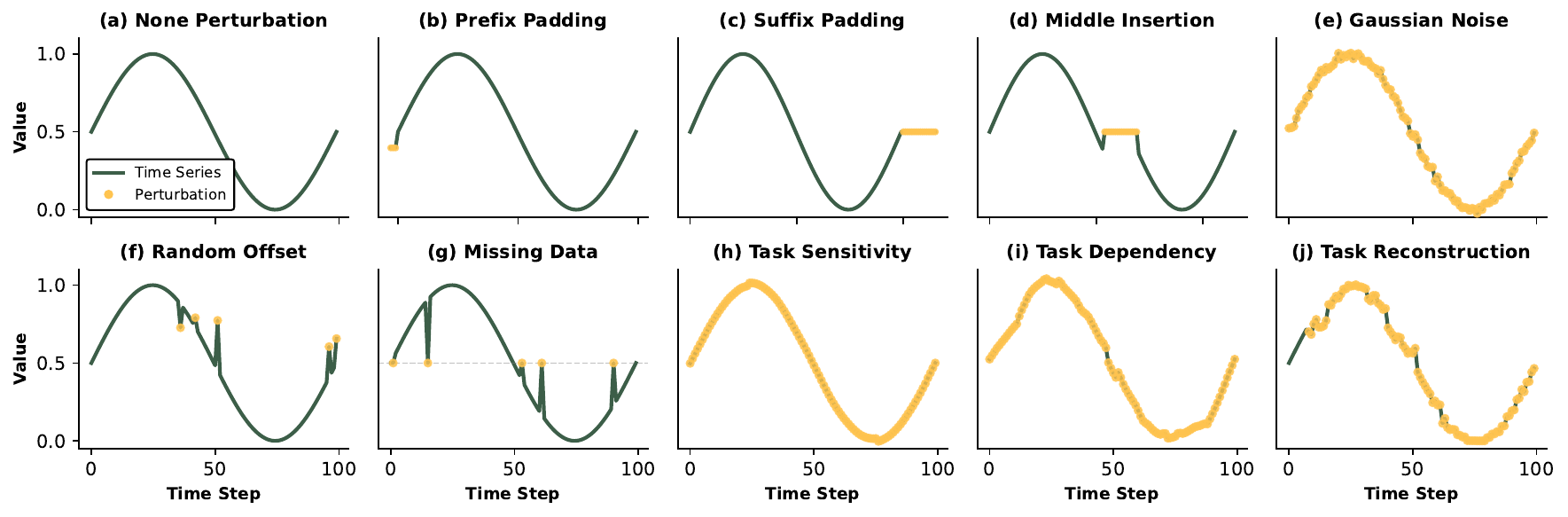}\vspace{-5mm}
    \caption{Visualization of time series with different perturbations.}
    \label{fig:perturbations}\vspace{-3mm}
\end{figure}

Furthermore, we summarize nine diversified strategies $\pi_S$ considered in this work in Table~\ref{tab:timeseries_nlp_perturbations}, providing a unified taxonomy of time series perturbations together with their analogous counterparts in natural language processing. 
We categorize all perturbation methods into \textit{Task-Agnostic} and \textit{Task-Specific} classes according to whether the injected modifications are agnostic to the downstream forecasting objective. 
Throughout the Table~\ref{tab:timeseries_nlp_perturbations}, $\mathbf{X}'$ denotes the perturbed time series derived from the original input $\mathbf{\mathbf{X}}$ by $\pi_S$. 
The index $i$ indicates the perturbation position, where $i=0$ corresponds to prefix insertion, $i=L$ to suffix insertion, and $0<i<L$ to middle insertion. 
$\mathbf{c}^{\ell \times D}$ is padding for structural perturbations, where $\ell$ and $c$ are the perturbation length and padding value, respectively.
For noise perturbations, $\sigma$ denotes the standard deviation of $\mathbf{X}$ and $\eta$ controls the perturbation intensity. 
For task-specific perturbations, $\pi_S^s$,  $\pi_S^d$, $\pi_S^r$ represent parameterized perturbation functions, respectively, where $\pi_S^r$ uses a TSFM (Chronos in here) for time series reconstruction. 
In detail, $\pi_S^s$ denotes structure-sensitive perturbations modulated by task difficulty, where regions that are more difficult to predict or intrinsically unstable receive stronger perturbations, while more predictable and stable structures are perturbed more mildly. 
$\pi_S^d$ means structure-aligned, task-dependent perturbations, where perturbation directions are guided by the task-relevant structural components of the input, so that the perturbations induce maximal changes in the task output while preserving the natural data geometry.
In order to visually demonstrate different strategies, we present a schematic diagram for toy data in Figure~\ref{fig:perturbations}.

\begin{table*}[ht]
\vspace{-5mm}
\centering
\caption{Description of time series perturbation strategies with analogous methods in language models.}
\label{tab:timeseries_nlp_perturbations}
\resizebox{0.99\columnwidth}{!}{
\setlength{\tabcolsep}{6pt}
\renewcommand{\arraystretch}{1.1}
\small
\begin{tabular}{
>{\centering\arraybackslash}m{1.6cm}
>{\centering\arraybackslash}m{1.8cm} 
>{\centering\arraybackslash}m{1.8cm}
>{\centering\arraybackslash}m{4.8cm}
>{\raggedright\arraybackslash}m{3.1cm}
>{\raggedright\arraybackslash}m{3.8cm}
>{\centering\arraybackslash}m{2.5cm}
}
\toprule
\textbf{Taxonomy} &
\textbf{Perturbation Type} &
\textbf{Method} &
\textbf{Definition} &
\textbf{Default Setting} &
\textbf{Analogous Method in LLMs} &
\textbf{Ref.} \\
\midrule

\multirow{20}{*}{\textbf{Task-Agnostic}}
& \multirow{10}{*}{\parbox{1.8cm}{\centering Structural Perturbations}}
& Prefix Padding
& $\mathbf{X}' = [\mathbf{c}, \mathbf{X}],\ i=0$
& $\ell \sim \mathcal{U}\{16,32\}$;\newline $c \sim \mathcal{U}(-1,1)$
& \textbf{Original:} ``The result is significant.'' \newline
  \textbf{Perturbed:} ``According to the professor, the result is significant.''
& \cite{perez2021true,zhao2021calibrate}
\\ \cline{3-7}

& & Suffix Padding
& $\mathbf{X}' = [\mathbf{X}, \mathbf{c}],\ i=L$
& $\ell \sim \mathcal{U}\{16,32\}$;\newline $c \sim \mathcal{U}(-1,1)$
& \textbf{Original:} ``The experiment failed.'' \newline
  \textbf{Perturbed:} ``The experiment failed. [END OF REPORT]''
& \cite{wallace2019universal}
\\ \cline{3-7}

& & Middle Insertion
& $\mathbf{X}' = [\mathbf{X}_{1:i}, \mathbf{c}, \mathbf{X}_{i+1:L}]$,\newline
$\ 0<i<L$
& $\ell \sim \mathcal{U}\{16,32\}$;\newline
  $i \sim \mathcal{U}(0.3L,0.7L)$,\newline
  $c=\mathbf{x}_i$
& \textbf{Original:} ``The policy is harmful.'' \newline
  \textbf{Perturbed:} ``The policy is, this is a neutral statement, harmful.''
& \cite{jia2017adversarial,liu-etal-2024-lost}
\\ \cline{2-7}

& \multirow{6}{*}{\parbox{1.8cm}{\centering Noise Perturbations}}
& Gaussian Noise
& $\mathbf{X}' = \mathbf{X} + \mathcal{N}(0,(\eta\sigma)^2\mathbf{I})$
& $\eta \in \{0.05,0.1,0.2\}$
& $e'_i = e_i + \epsilon,\ \epsilon \sim \mathcal{N}(0,\sigma^2\mathbf{I})$
& \cite{liu2024explaining}
\\ \cline{3-7}

& & Random Offset
& $\mathbf{x}'_i = \mathbf{x}_i + c$,
$\ 0<i<L$
& $c \in \mathcal{U}\{-\sigma,\sigma\}$
& \textbf{Original:} ``The result is important.'' \newline
  \textbf{Perturbed:} ``The result is crucial.''
& \cite{wei2019eda}
\\ \cline{3-7}

& & Missing Data
& $\mathbf{X}' = \mathbf{M}\cdot\mathbf{X}$
& Mask $\mathbf{M}\in \{0,1\}^{L \times D}$
& \textbf{Original:} ``The market reacts strongly.'' \newline
  \textbf{Perturbed:} ``The market [MASK] strongly.''
& \cite{liu2024protecting}
\\
\midrule

\multirow{8}{*}{\textbf{Task-Specific}}
& \multirow{8}{*}{\parbox{1.8cm}{\centering Task-specific Perturbations}}
& Task Sensitivity
& $\mathbf{X}' = \mathbf{X} + \eta \pi_S^s(\mathbf{X},t)$
& $\eta \sim \mathcal{U}(0.01, 0.05)$
& -
& Refer to Algorithm\ref{Task_sensitive}
\\ \cline{3-7}

& & Task Dependency
& $\mathbf{X}' = \mathbf{X} + \eta \pi_S^d(\mathbf{X},t)$
& $\eta \sim \mathcal{U}(0.01, 0.05)$
& -
& Refer to Algorithm\ref{Task_dependent}
\\ \cline{3-7}

& & Task Reconstruction
& $\mathbf{X}' \sim \pi_S^r(\cdot\mid \mathbf{X})$
& $\tau \in \{0.3,0.4,\cdots,0.9\}$
& \textbf{Original:} ``The experiment failed due to noise.'' \newline
  \textbf{Paraphrase:} ``Noise caused the experiment to be unsuccessful.'' 
& \cite{ribeiro2020beyond,wang2025diversified}
\\
\bottomrule
\end{tabular}
}
\end{table*}

\begin{algorithm2e}[!t]
\SetAlgoLined
\KwIn{Original input sequence $\mathbf{X}$, decomposition period $p$, intensity control $\eta$}
\KwOut{Perturbed time series $\mathbf{X}'$}
\BlankLine
\tcp{1. Decompose input into structural components for sensitivity analysis}
$(\mathbf{T}, \mathbf{S}, \mathbf{R}) \leftarrow \text{STL}(\mathbf{X}, p)$\;
\BlankLine
\tcp{2. Modulate perturbation strength by task difficulty and intrinsic instability}
$\mathbf{f}_T \leftarrow |\nabla \mathbf{T}|$ \tcp*{Heightened sensitivity to trend shifts}
$\mathbf{S}_{shift} \leftarrow \text{Roll}(\mathbf{S}, p)$\;
$\mathbf{f}_S \leftarrow |\mathbf{S} - \mathbf{S}_{shift}|$ \tcp*{Capturing seasonal non-stationarity}
$\mathbf{f}_R \leftarrow |\mathbf{R}|$ \tcp*{Identifying unpredictable residual regions}
\BlankLine
\tcp{3. Consolidate perturbations with sign and mean constraints}
$\mathbf{f} \leftarrow \mathbf{f}_T + \mathbf{f}_S + \mathbf{f}_R$\;
$\mathbf{f} \leftarrow \text{sign}(\mathbf{X} + \epsilon) \odot |\mathbf{f}|$\;
$\mathbf{f} \leftarrow \mathbf{f} - \text{mean}(\mathbf{f})$\;
\BlankLine
\tcp{4. Final intensity normalization for output consistency}
$\eta \leftarrow \eta \cdot \frac{\|\mathbf{X}\|_2}{\|\mathbf{f}\|_2 + \epsilon}$\;
$\mathbf{X}' \leftarrow \mathbf{X} + \eta \mathbf{f}$\;
\Return $\mathbf{X}'$
\caption{Structure-sensitive Perturbation $\pi_S^s$ Modulated by Task Difficulty}
\label{Task_sensitive}
\end{algorithm2e}

\section{Experimental Settings Details}

\begin{algorithm2e}[!t]
\SetAlgoLined
\KwIn{Original input sequence $\mathbf{X}$, decomposition period $p$, intensity control $\eta$, local window $w$}
\KwOut{Perturbed time series $\mathbf{X}'$}
\BlankLine
\tcp{1. Extract task-relevant structural components via STL decomposition}
$(\mathbf{T}, \mathbf{S}, \mathbf{R}) \leftarrow \text{STL}(\mathbf{X}, p)$\;
\BlankLine
\tcp{2. Guide perturbation directions to preserve natural data geometry}
$\mathbf{f}_T \leftarrow \nabla \mathbf{T}$ \tcp*{Trend-aligned directional guidance}
$\mathbf{f}_S \leftarrow |\mathbf{S}|$ \tcp*{Seasonal structural alignment}
$\mathbf{f}_R \leftarrow \text{LocalStd}(\mathbf{R}, w)$ \tcp*{Residual local variance capturing geometry}
\BlankLine
\tcp{3. Enforce sign consistency and zero-mean constraints}
$\mathbf{f} \leftarrow \mathbf{f}_T + \mathbf{f}_S + \mathbf{f}_R$\;
$\mathbf{f} \leftarrow \text{sign}(\mathbf{X} + \epsilon) \odot |\mathbf{f}|$\;
$\mathbf{f} \leftarrow \mathbf{f} - \text{mean}(\mathbf{f})$\;
\BlankLine
\tcp{4. Scale perturbation intensity according to input signal-to-noise ratio}
$\eta \leftarrow \eta \cdot \frac{\|\mathbf{X}\|_2}{\|\mathbf{f}\|_2 + \epsilon}$\;
$\mathbf{X}' \leftarrow \mathbf{X} + \eta \mathbf{f}$\;
\Return $\mathbf{X}'$
\caption{Structure-aligned Task-dependent Perturbation $\pi_S^d$}
\label{Task_dependent}
\end{algorithm2e}

\label{app:exp_params}
\subsection{Datasets}
We evaluate the universal forecasting and inference capabilities of TSFMs on three widely used public datasets, including ETTh1, ETTm1, Electricity, and Traffic from TSLib\footnote{\url{https://github.com/thuml/Time-Series-Library}}.
These datasets span diverse sampling frequencies and domain characteristics, and are commonly adopted in prior long-term forecasting literature~\cite{liang2024foundation,yao2024towards,wang2024timemixerdecomposablemultiscalemixing}.
Table~\ref{tab:dataset_stats} summarizes the datasets used in our inference-only evaluation.
For all datasets, we extract the univariate Output Target (OT) as the forecasting variable (i.e., $D=D_{\text{out}}=1$), as some TSFMs only support univariate prediction.
Following prior benchmarks~\cite{wu2023timesnettemporal2dvariationmodeling}, each dataset is originally evaluated under multiple prediction horizons.
In this work, unless otherwise specified, we fix the prediction horizon to $H=96$ and perform fixed-horizon forecasting for all inference and robustness evaluations, in order to ensure fair and consistent comparisons across different models.

\begin{table*}[!h]
    \centering
    \vspace{-5mm}
    \caption{Time series forecasting datasets for diversified scaling inference experiments.}
    \resizebox{0.65\columnwidth}{!}{
    \setlength{\tabcolsep}{1.5pt}
    \renewcommand{\arraystretch}{1.3}
    \small
    \begin{tabular}{
        >{\centering\arraybackslash}m{2.2cm}
        >{\centering\arraybackslash}m{1.6cm}
        >{\centering\arraybackslash}m{3cm}
        >{\centering\arraybackslash}m{2cm}
        >{\centering\arraybackslash}m{3cm}
    }
        \toprule
        \textbf{Dataset} 
        & \textbf{Dimension} 
        & \textbf{Dataset Size}
        & \textbf{Frequency} 
        & \textbf{Prediction Horizon} \\
        \midrule
        ETTh1        & 7 & (8640, 2880, 2880) & Hourly  & 96 \\
        ETTm1        & 7 & (34560, 11520, 11520) & 15 mins  & 96 \\
        Electricity & 321 & (18412, 2630, 5261) & Hourly  & 96 \\
        Traffic     & 862 & (12280, 1754, 3509) & Hourly  & 96 \\
        \bottomrule
    \end{tabular}
    }\label{tab:dataset_stats}
    \vspace{-5mm}
\end{table*}

\subsection{Time Series Foundation Models (TSFMs) Architectures}
We evaluates four representative TSFMs, including TimesFM~\cite{liang2024foundation}, Chronos~\cite{ansari2024chronoslearninglanguagetime}, Time-MoE~\cite{shi2024time} and Moirai~\cite{woo2024unified}, where all used versions are summarized in the Table~\ref{tab:models}. 
Except for scaled model size variants, the default TSFM we use is indicated with \colorbox{transgray}{a gray background}.
Specifically, each model is invoked through a common inference interface, receiving identical input preprocessing, context length, prediction horizon, sliding-window stride, decoding strategy, and perturbation configurations. 
All TSFMs are evaluated under a unified, inference-only framework that standardizes the forecasting protocol across architectures~\cite{bommasani2021opportunities}.
\begin{table*}[ht]
\vspace{-4mm}
\caption{Summary of TSFMs and links used in our diversified scaling inference experiments.}
\label{tab:models}
\centering
\resizebox{0.95\columnwidth}{!}{
\setlength{\tabcolsep}{2.5pt}
\begin{tabular}{
>{\centering\arraybackslash}m{2.2cm}
>{\centering\arraybackslash}m{3.8cm}
>{\centering\arraybackslash}m{2cm}
>{\raggedright\arraybackslash}m{12cm}
}
\toprule
\textbf{Model} & \textbf{Name} & \textbf{Size} & \textbf{HuggingFace Link} \\
\midrule
\multirow{2}{*}{TimesFM} 
&  \cellcolor{transgray} \texttt{TimesFM-2.5-200M }& \cellcolor{transgray} 200M & \cellcolor{transgray}\url{https://huggingface.co/google/timesfm-2.5-200m-pytorch} \\
& \texttt{TimesFM-2.0-500M} & 500M & \url{https://huggingface.co/google/timesfm-2.0-500m-pytorch} \\
\midrule
\multirow{3}{*}{Moirai} 
& \cellcolor{transgray}\texttt{Moirai-1.1-R-Small} &\cellcolor{transgray} 55M  & \cellcolor{transgray}\url{https://huggingface.co/Salesforce/moirai-1.1-R-small} \\
& \texttt{Moirai-1.1-R-Base}  & 365M & \url{https://huggingface.co/Salesforce/moirai-1.1-R-base} \\
& \texttt{Moirai-1.1-R-Large} & 1.24B & \url{https://huggingface.co/Salesforce/moirai-1.1-R-large} \\
\midrule

\multirow{5}{*}{Chronos} & \cellcolor{transgray}\texttt{Chronos-T5-Tiny}  & \cellcolor{transgray} 8M  & \cellcolor{transgray} \url{https://huggingface.co/amazon/chronos-t5-tiny} \\
& \texttt{Chronos-T5-Mini}  & 20M  & \url{https://huggingface.co/amazon/chronos-t5-mini} \\
& \texttt{Chronos-T5-Small} & 46M  & \url{https://huggingface.co/amazon/chronos-t5-small} \\
& \texttt{Chronos-T5-Base}  & 200M & \url{https://huggingface.co/amazon/chronos-t5-base} \\
& \texttt{Chronos-T5-Large} & 710M & \url{https://huggingface.co/amazon/chronos-t5-large} \\
\midrule

\multirow{2}{*}{TimeMoE} 
& \cellcolor{transgray} \texttt{TimeMoE-50M}  & \cellcolor{transgray} 113M & \cellcolor{transgray} \url{https://huggingface.co/Maple728/TimeMoE-50M} \\
& \texttt{TimeMoE-200M} & 453M & \url{https://huggingface.co/Maple728/TimeMoE-200M} \\

\bottomrule
\end{tabular}
}
\vspace{-4mm}
\end{table*}

\subsection{Experimental Setting}

For the scaling inference experiments, all are under a unified inference-only evaluation, \textit{without any model finetuning or architectural modification.} 
Our experiments are conducted on a single NVIDIA H100 GPU, and the inference process is highly amenable to parallelization.
The detailed experimental settings are provided in Table~\ref{tab:configs}. 
Specifically, we adopt a sliding window \textbf{forecasting protocol} over each time series $\{\mathbf{x}_1, \dots, \mathbf{x}_L\}$~\cite{lai2018modeling}. 
Each forecast sample is set to a length $L=512$ of the context window and a prediction horizon of $H=96$ as defaults~\cite{zhou2021informerefficienttransformerlong}. We also vary $L \in \{32, 64, 128, 256, 512, 1024\}$ in the scaling inference experiments.
The sliding window advances along the time axis with a stride of $32$, producing overlapping but distinct time series samples.
The \textbf{decoding strategy} is stochastic decoding with top-$p=1.0$, and we set a maximum of $N=128$ inferences per context window to scale inference.
The temperature $\tau$ in TSFMs is usually set to $0.7$, but in the scaling inference experiment, we also explored values between $0.0$ to $1.2$ for decoding diversity.
For \textbf{diversified sampling}, each single forecast sample is obtained through a $\pi_S$ of six task-agnostic or three task-specific perturbations, and the results are aggregated through repeated decoding in the input of a kind of independent perturbation. The detailed configurations of diversified strategies are in Table~\ref{tab:timeseries_nlp_perturbations}.

\begin{table}[ht]
    \centering
    \vspace{-4mm}
    \caption{Configurations of our scaling inference experiments for all TSFMs.}
    \resizebox{0.8\columnwidth}{!}{
    \label{tab:configs} 
    \setlength{\tabcolsep}{3pt}  
    \renewcommand{\arraystretch}{1.3} 
    \small
    \begin{tabular}{
        >{\Centering\arraybackslash}m{3.5cm} 
        >{\Centering\arraybackslash}m{4.5cm}
        >{\Centering\arraybackslash}m{6cm} 
    }
        \toprule
        \textbf{Category} & \textbf{Variable Name} & \textbf{Variable Value} \\
        \midrule
        \multirow{4}{*}[-0.8ex]{\textbf{Forecasting Protocol}} 
        & Prediction Horizon & 96 \\
        & Context Length & $\{32, 64, 128, 256, 512, 1024\}$ \\
        & Sliding Window Stride & 32 \\
        & Aggregation Function & $\{$Exact Match, Majority Voting$\}$\\
        \midrule
        \multirow{4}{*}[-0.8ex]{\textbf{Decoding Strategy}}
        & Decoding Type & Stochastic Decoding \\
        & Temperature  & $\{0.0, 0.1, 0.2, \dots, 1.2\}$ \\
        & Top-$p$ & 1.0 \\ 
        & Numbers of Sampling & 128 \\
        \midrule
        \multirow{2}{*}[-0.2ex]{\textbf{Diversified Sampling}}
        & Task-agnostic Perturbations & $\{$Prefix, Suffix, Insert, Missing,\newline Gaussian, Random$\}$\\
        & Task-specific Perturbations & $\{$Sensitivity, Dependency, Reconstruction$\}$ \\  
        \bottomrule
    \end{tabular}
    }
\end{table}

\section{Additional Scaling Inference Experiments}
\label{app:scaling inference experiments}
In this section, we provide a detailed relationship for each type of TSFMs between the forecasting task performance and the amount of compute expended during various model families, model sizes, and decoding strategies.

\subsection{Chronos}
\begin{figure*}[h!]
    \centering
    \begin{subfigure}{0.3\textwidth}
        \includegraphics[width=\linewidth]{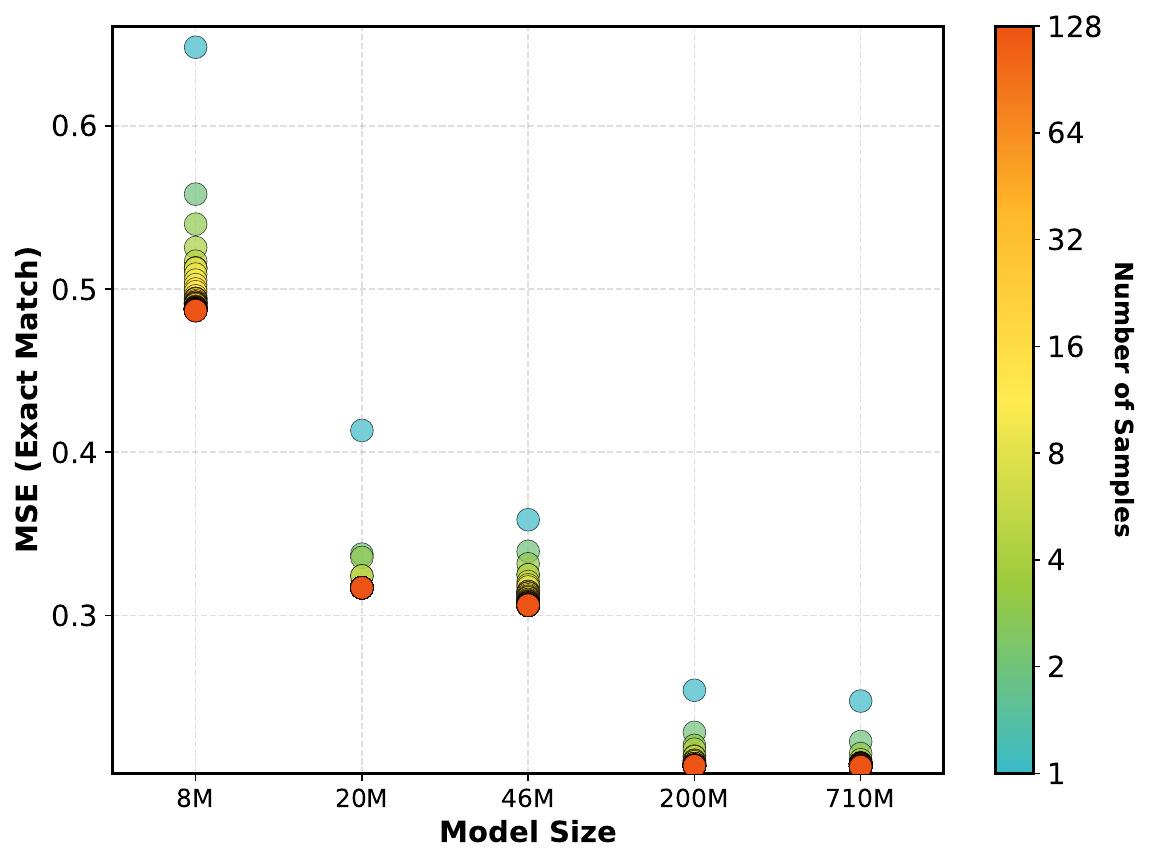}
        \caption{EM  vs. Model Size}
    \end{subfigure}
    \begin{subfigure}{0.34\textwidth}
        \includegraphics[width=\linewidth]{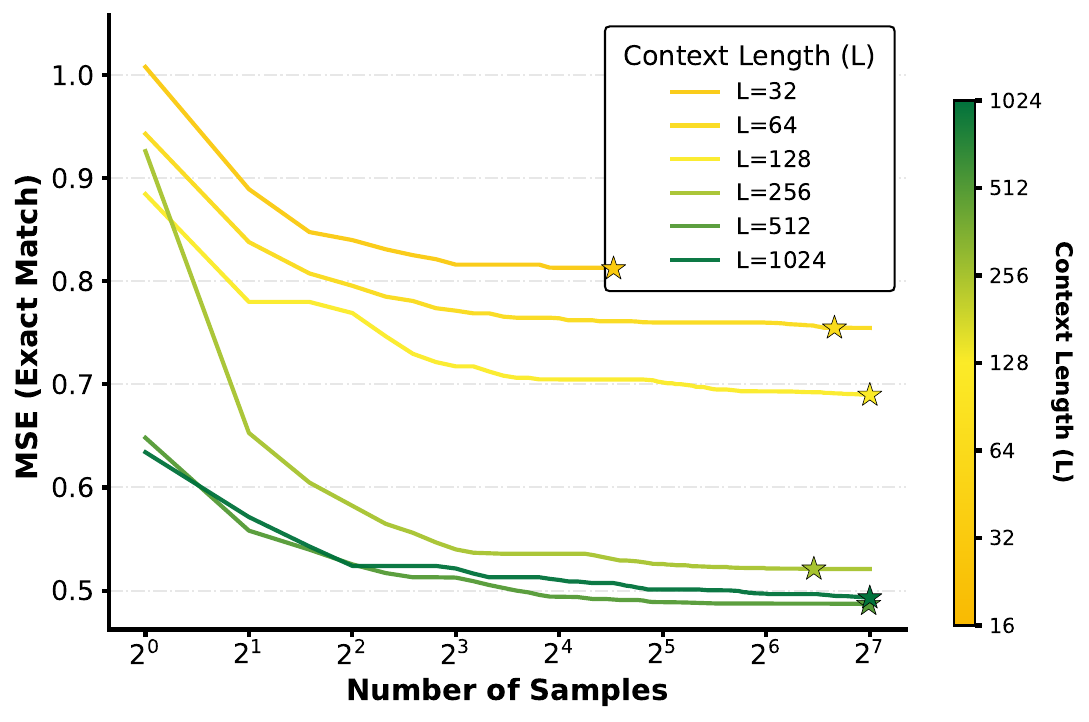}
        \caption{EM  vs. Context Length}
    \end{subfigure}
    \begin{subfigure}{0.34\textwidth}
        \includegraphics[width=\linewidth]{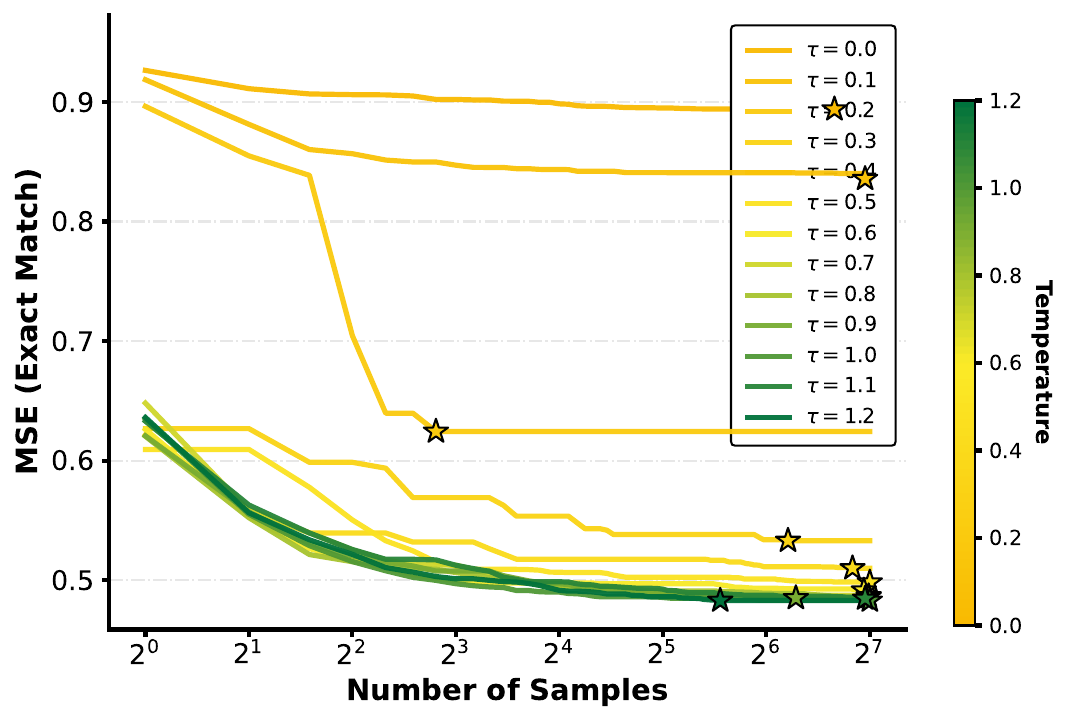}
        \caption{EM  vs. Temperature}
    \end{subfigure}
    \\
    \begin{subfigure}{0.3\textwidth}
        \includegraphics[width=\linewidth]{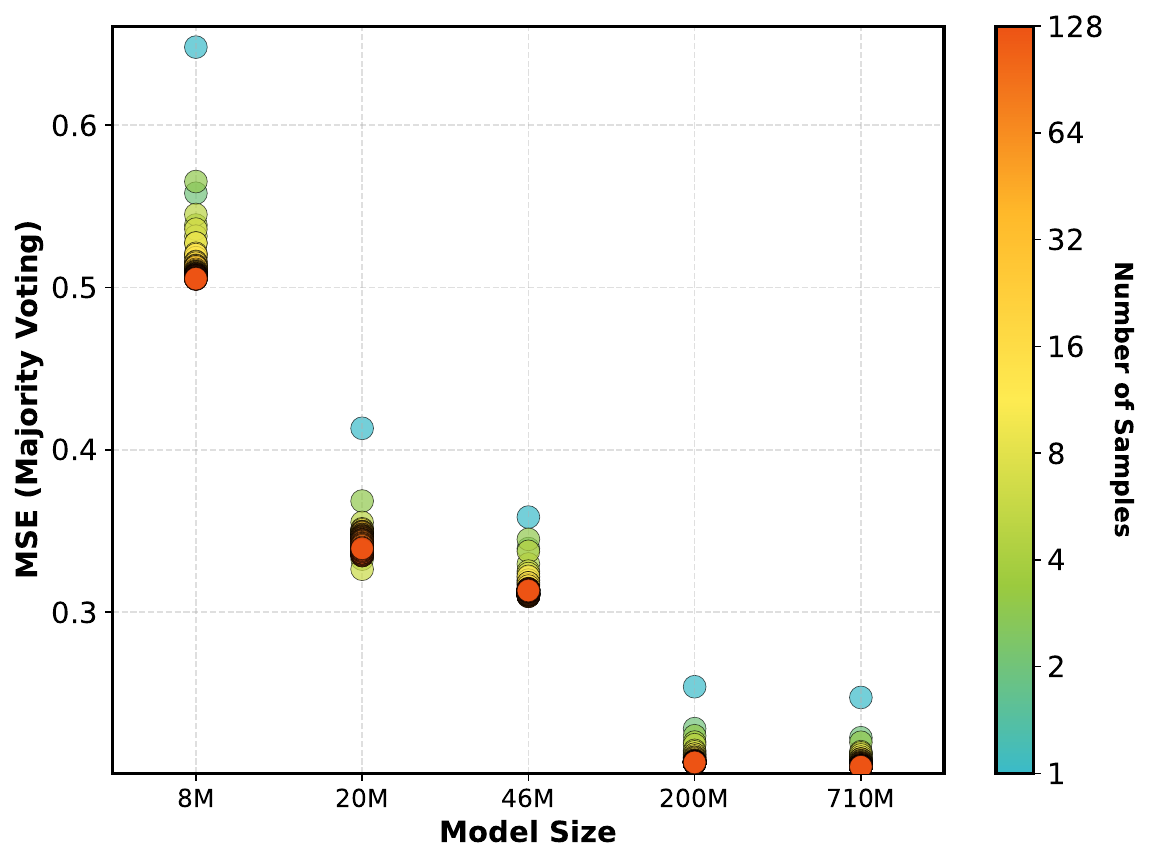}
        \caption{MV  vs. Model Size}
    \end{subfigure}
    \begin{subfigure}{0.34\textwidth}
        \includegraphics[width=\linewidth]{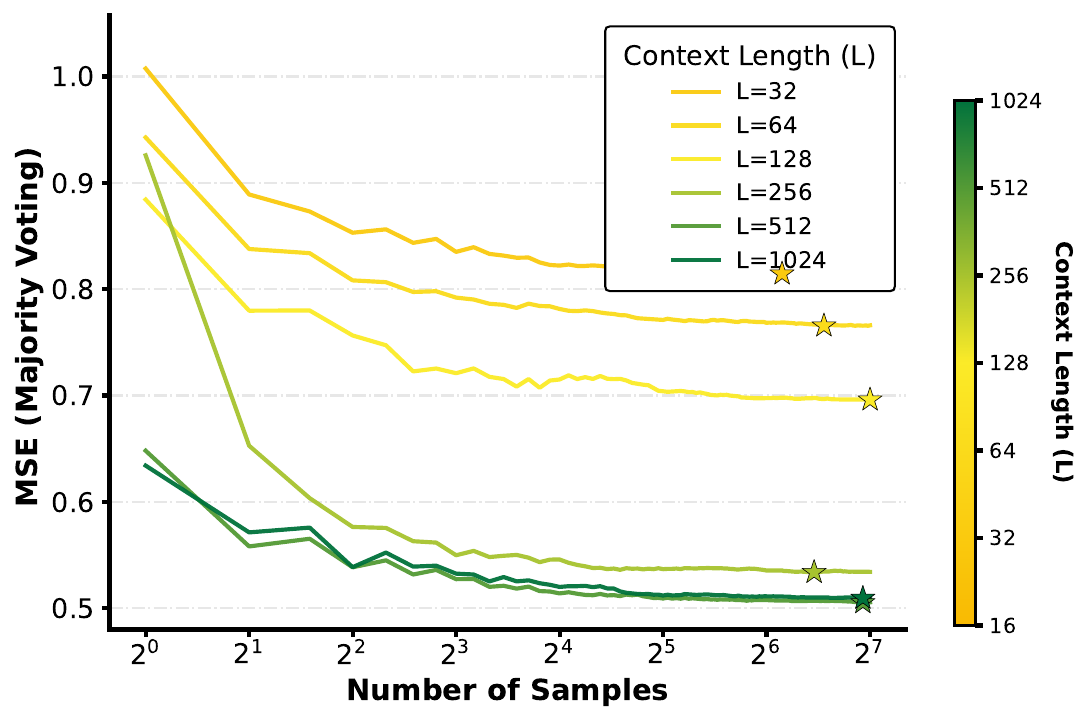}
        \caption{MV  vs. Context Length}
    \end{subfigure}
    \begin{subfigure}{0.34\textwidth}
        \includegraphics[width=\linewidth]{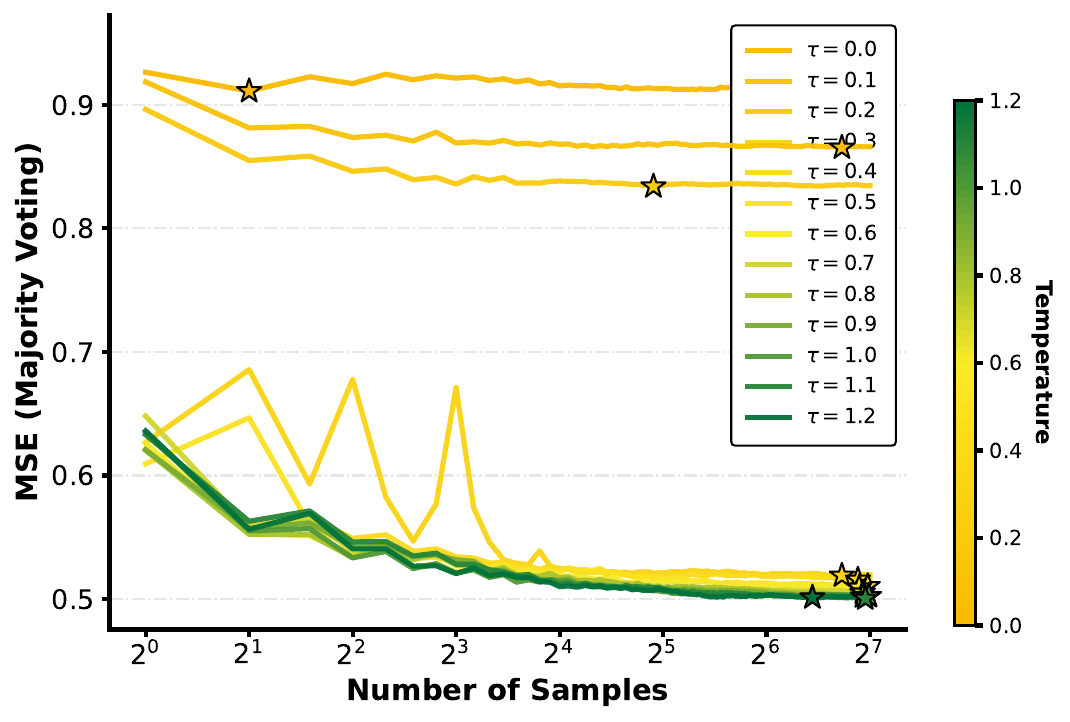}
        \caption{MV  vs. Temperature}
    \end{subfigure}
\vskip -0.1in
    \caption{Performance of \textit{Chronos} on the data \textit{Electricity} under different scaling factors and aggregation functions.
    }\label{chro1}
\end{figure*}

\begin{figure*}[h!]
    \centering
    \begin{subfigure}{0.3\textwidth}
        \includegraphics[width=\linewidth]{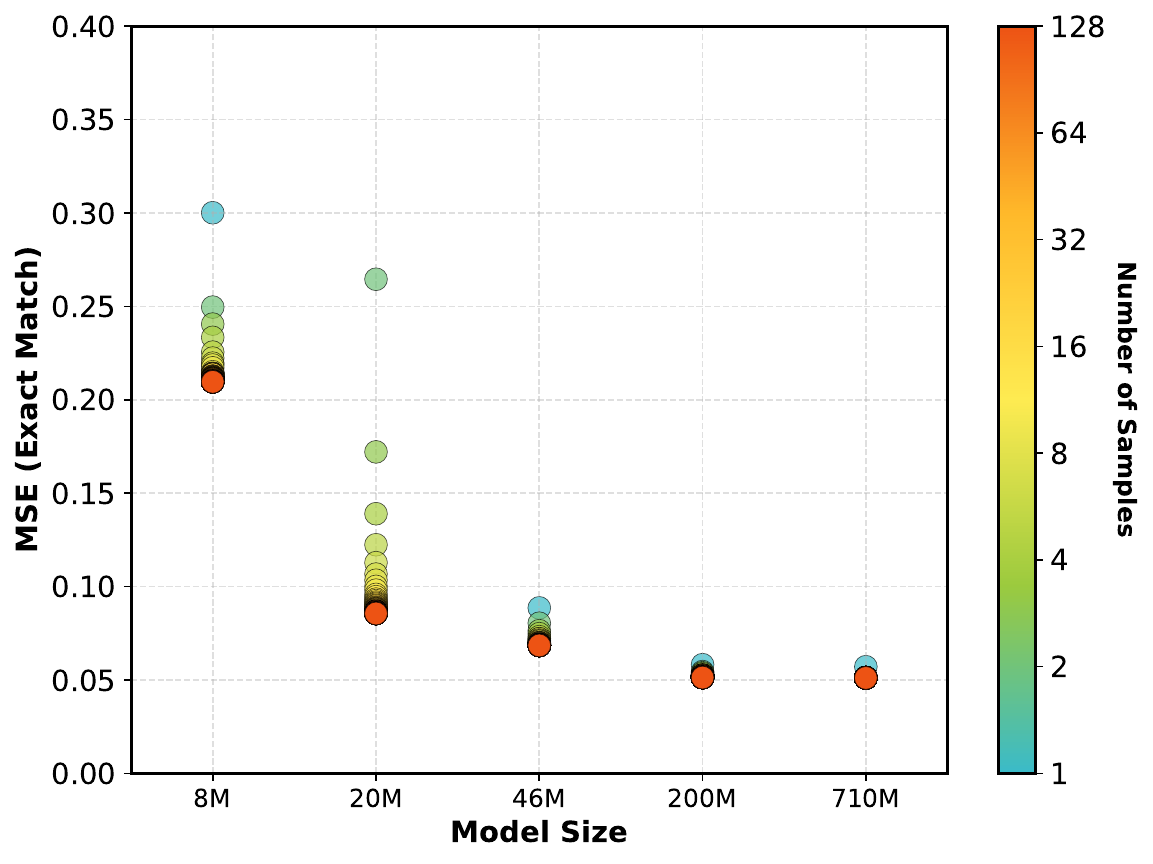}
        \caption{EM  vs. Model Size}
    \end{subfigure}
    \begin{subfigure}{0.34\textwidth}
        \includegraphics[width=\linewidth]{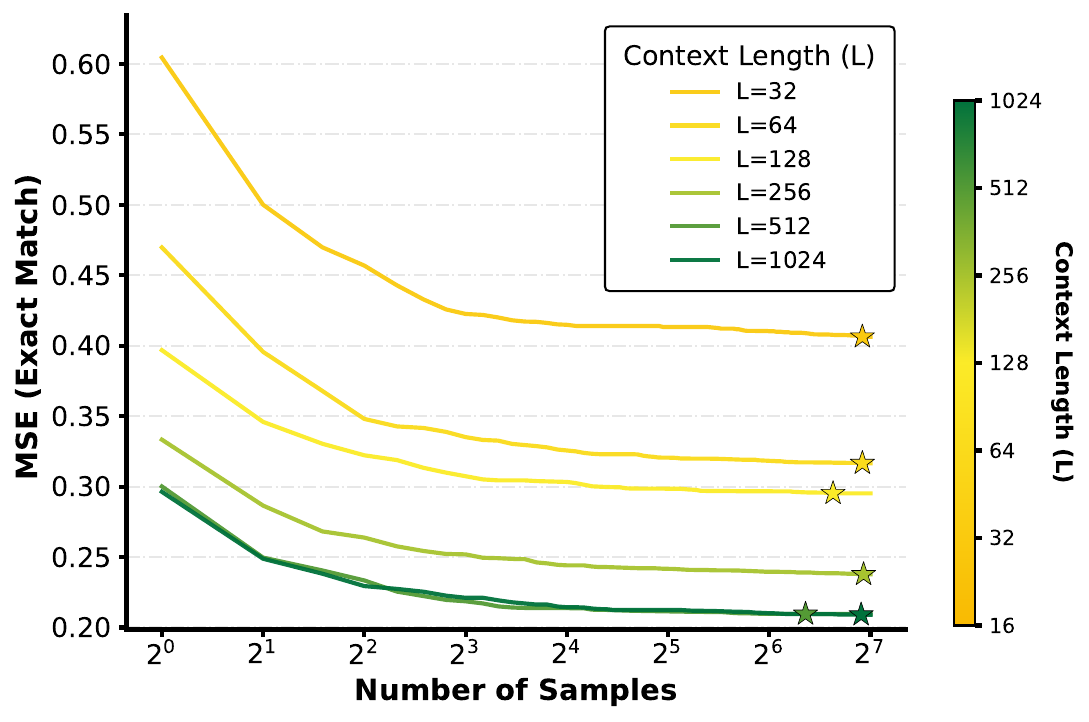}
        \caption{EM  vs. Context Length}
    \end{subfigure}
    \begin{subfigure}{0.34\textwidth}
        \includegraphics[width=\linewidth]{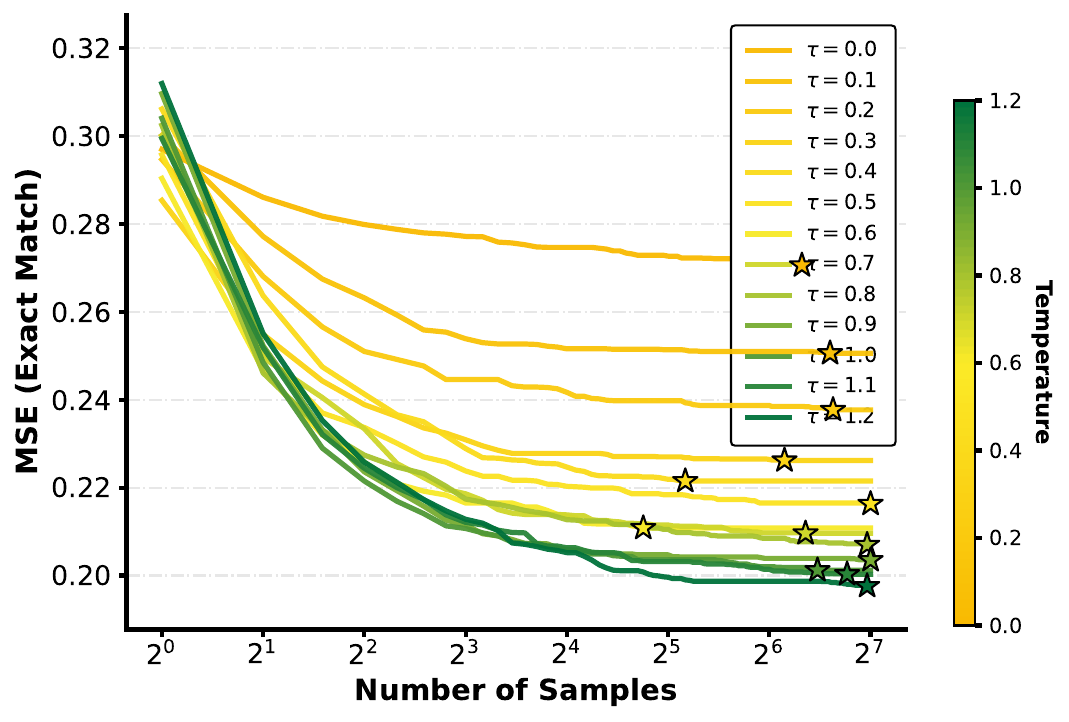}
        \caption{EM  vs. Temperature}
    \end{subfigure}
    \\
    \begin{subfigure}{0.3\textwidth}
        \includegraphics[width=\linewidth]{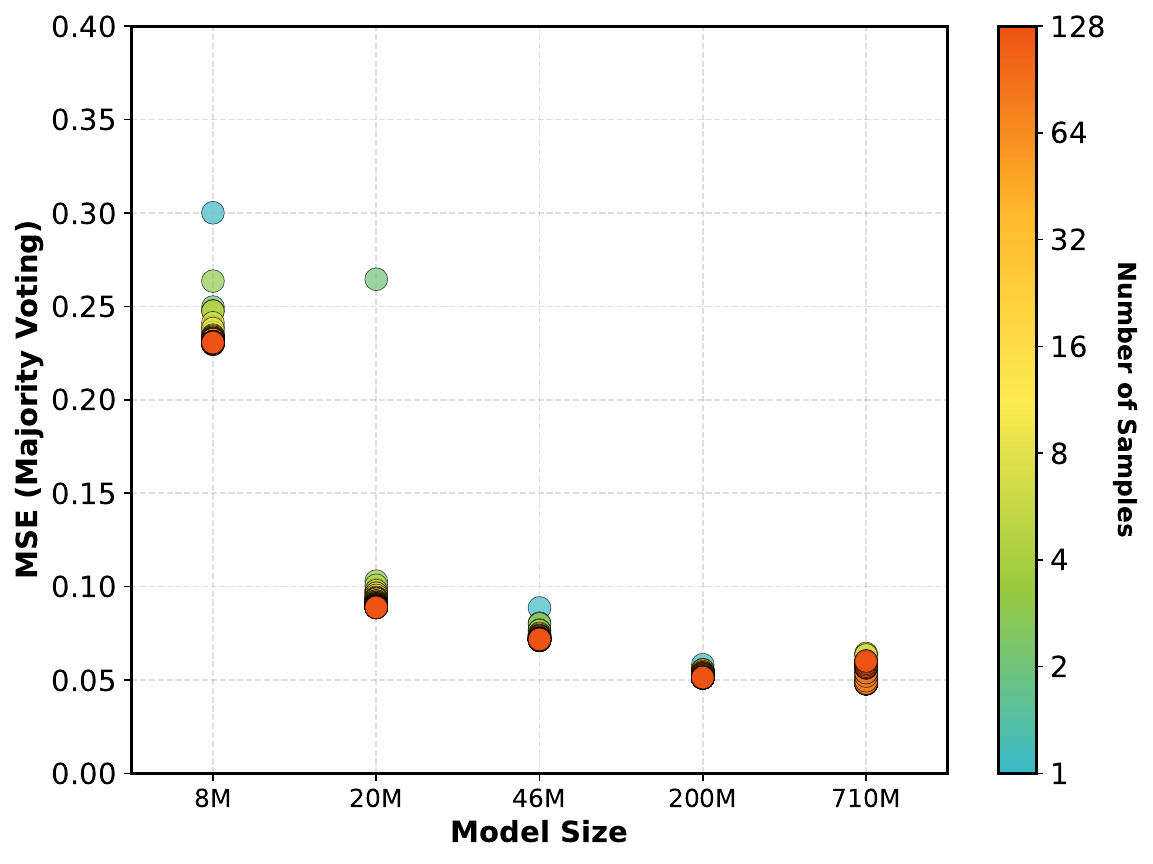}
        \caption{MV  vs. Model Size}
    \end{subfigure}
    \begin{subfigure}{0.34\textwidth}
        \includegraphics[width=\linewidth]{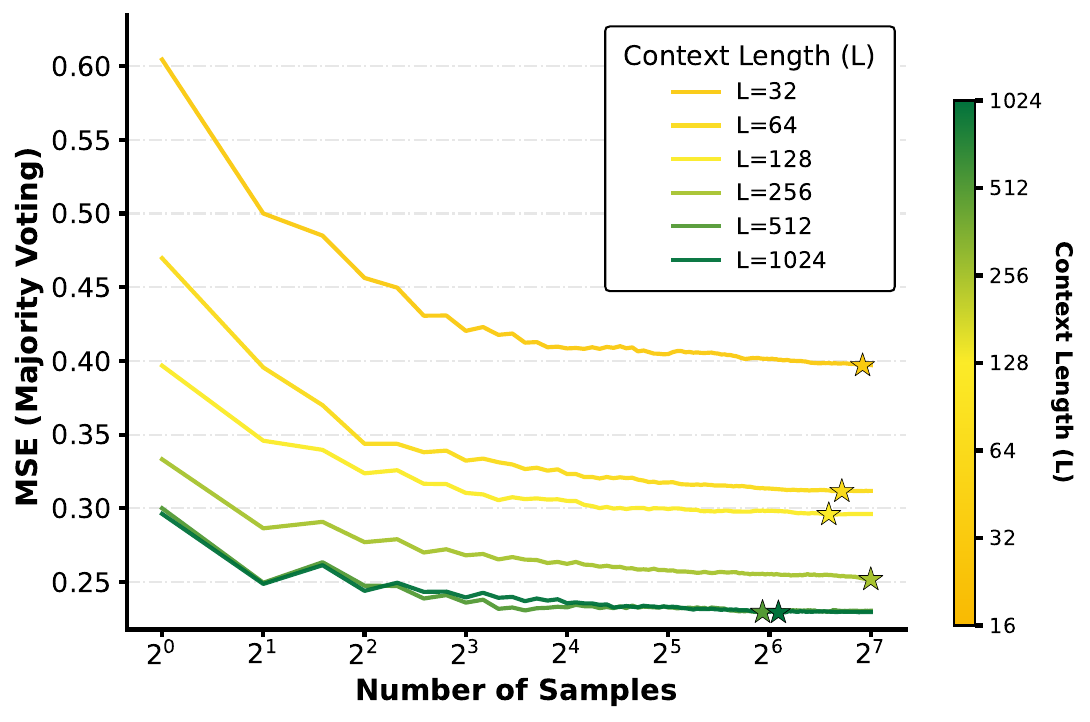}
        \caption{MV  vs. Context Length}
    \end{subfigure}
    \begin{subfigure}{0.34\textwidth}
        \includegraphics[width=\linewidth]{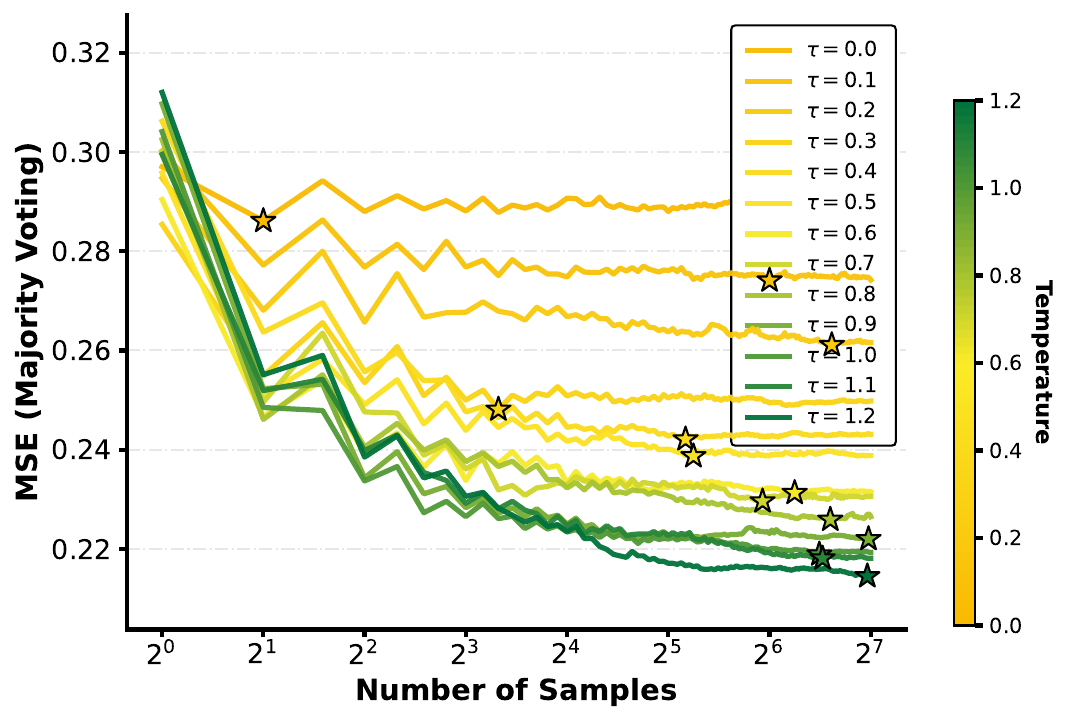}
        \caption{MV  vs. Temperature}
    \end{subfigure}
\vskip -0.1in
    \caption{Performance of \textit{Chronos} on the data \textit{Traffic} under different scaling factors and aggregation functions.
    }\label{chro2}
\end{figure*}

Chronos exhibits dataset-dependent inference scaling behavior that partially aligns with but does not strictly follow standard inference scaling expectations. The results are shown in Figures~\ref{fig:main}, \ref{chro1}, and \ref{chro2}. We observe that the optimal model size shifts systematically with dataset complexity, specifically, medium or higher-sized models achieve the lowest error on simpler datasets. The results scaling curves are irregular, indicating the absence of a universal scaling law. Increasing context length consistently improves performance across all datasets, yet the magnitude of improvement varies substantially with dataset complexity. Moreover, as sequence complexity increases, the optimal sampling temperature decreases, suggesting that Chronos benefits from reduced output diversity in more volatile settings. Overall, Chronos demonstrates structured but irregular inference scaling, where optimal configurations emerge from dataset-specific interactions rather than predictable scaling trends.

\begin{figure*}[t!]
    \centering
    \begin{subfigure}{0.3\textwidth}
        \includegraphics[width=\linewidth]{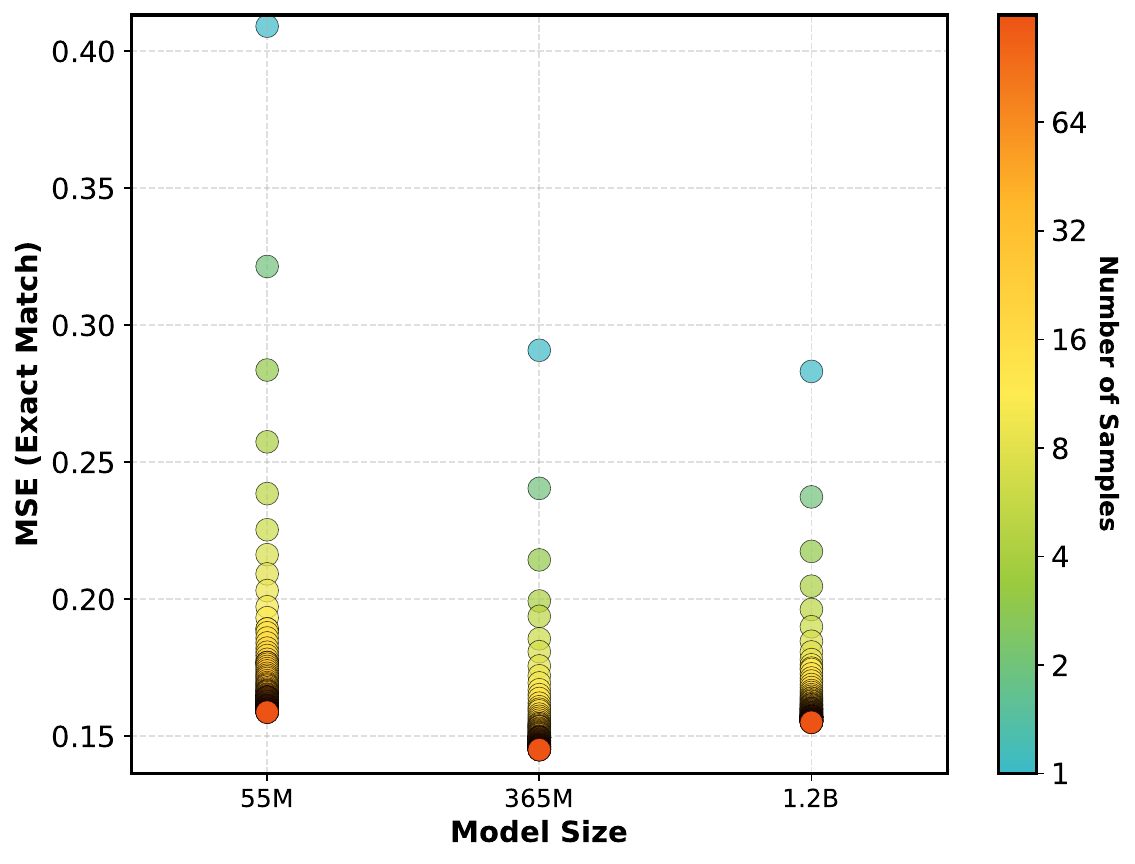}
        \caption{EM  vs. Model Size}
    \end{subfigure}
    \begin{subfigure}{0.34\textwidth}
        \includegraphics[width=\linewidth]{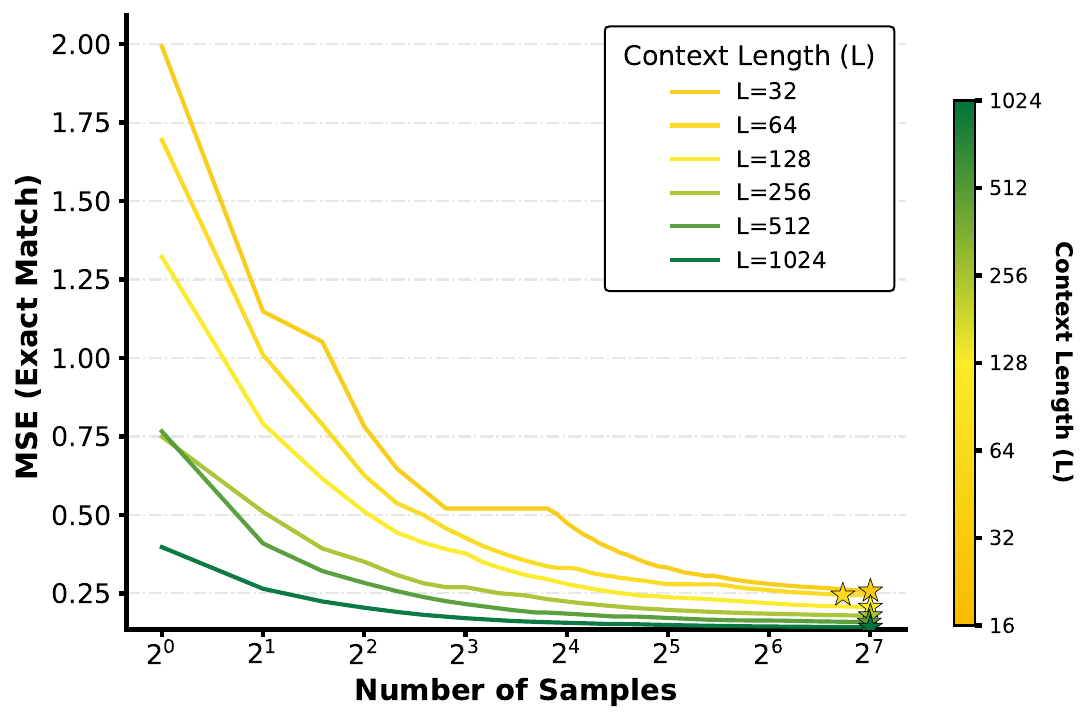}
        \caption{EM  vs. Context Length}
    \end{subfigure}
    \begin{subfigure}{0.34\textwidth}
        \includegraphics[width=\linewidth]{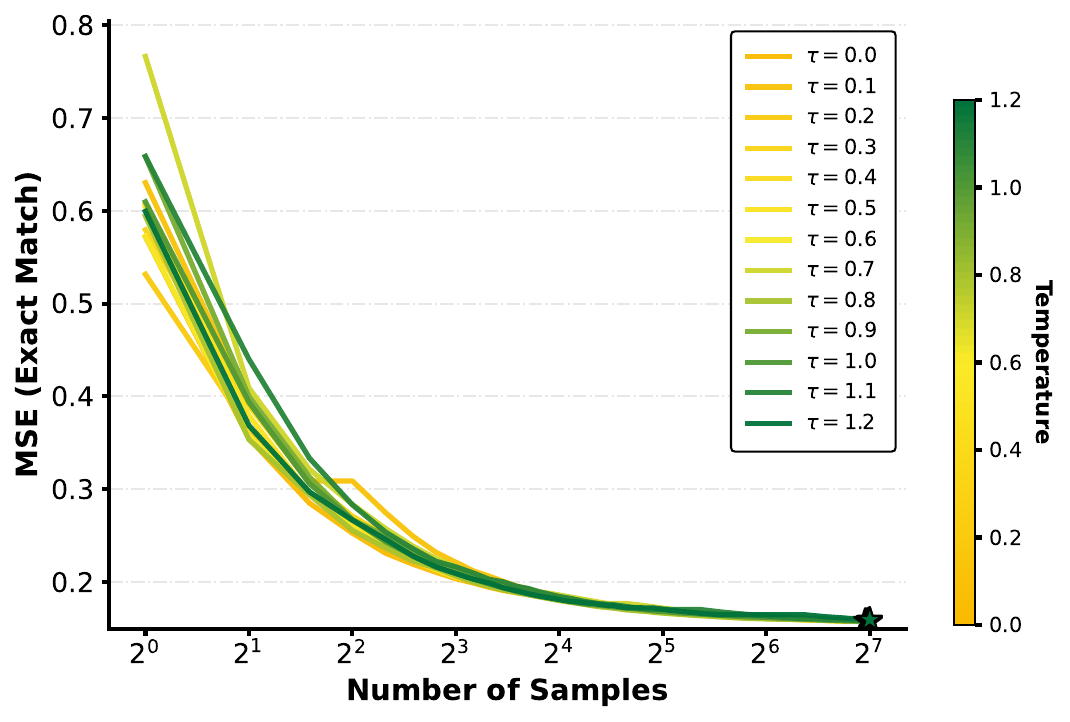}
        \caption{EM  vs. Temperature}
    \end{subfigure}
    \\
    \begin{subfigure}{0.3\textwidth}
        \includegraphics[width=\linewidth]{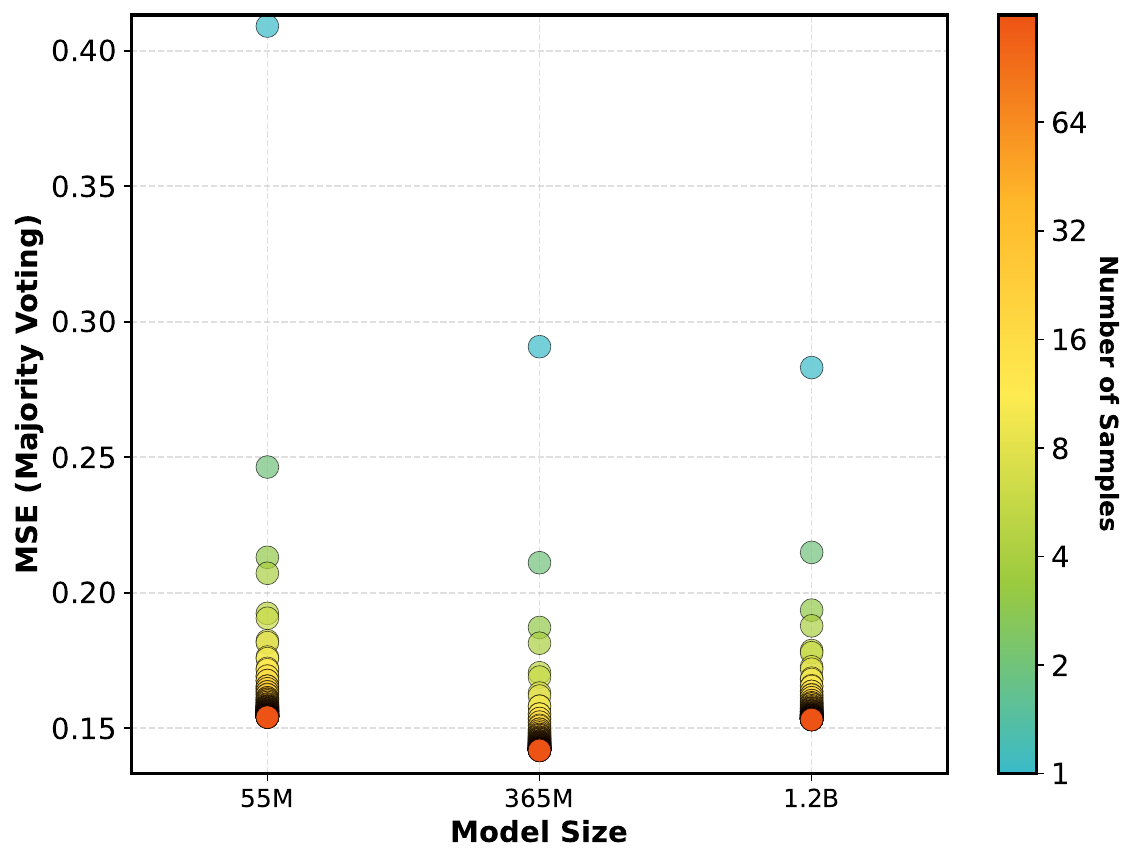}
        \caption{MV  vs. Model Size}
    \end{subfigure}
    \begin{subfigure}{0.34\textwidth}
        \includegraphics[width=\linewidth]{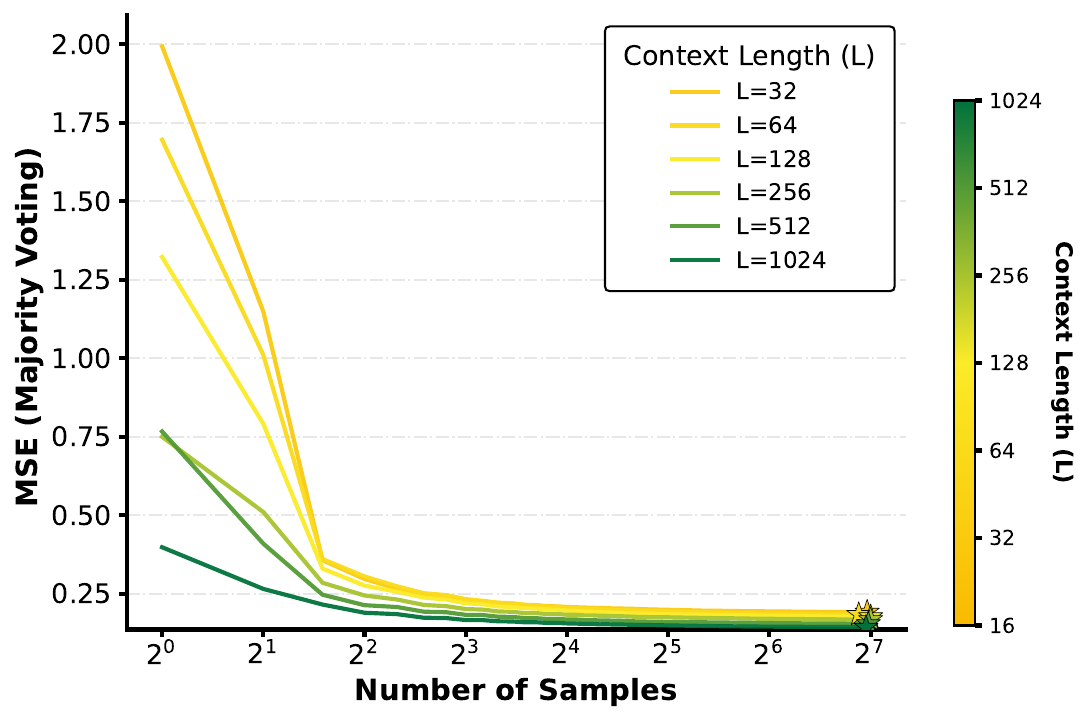}
        \caption{MV  vs. Context Length}
    \end{subfigure}
    \begin{subfigure}{0.34\textwidth}
        \includegraphics[width=\linewidth]{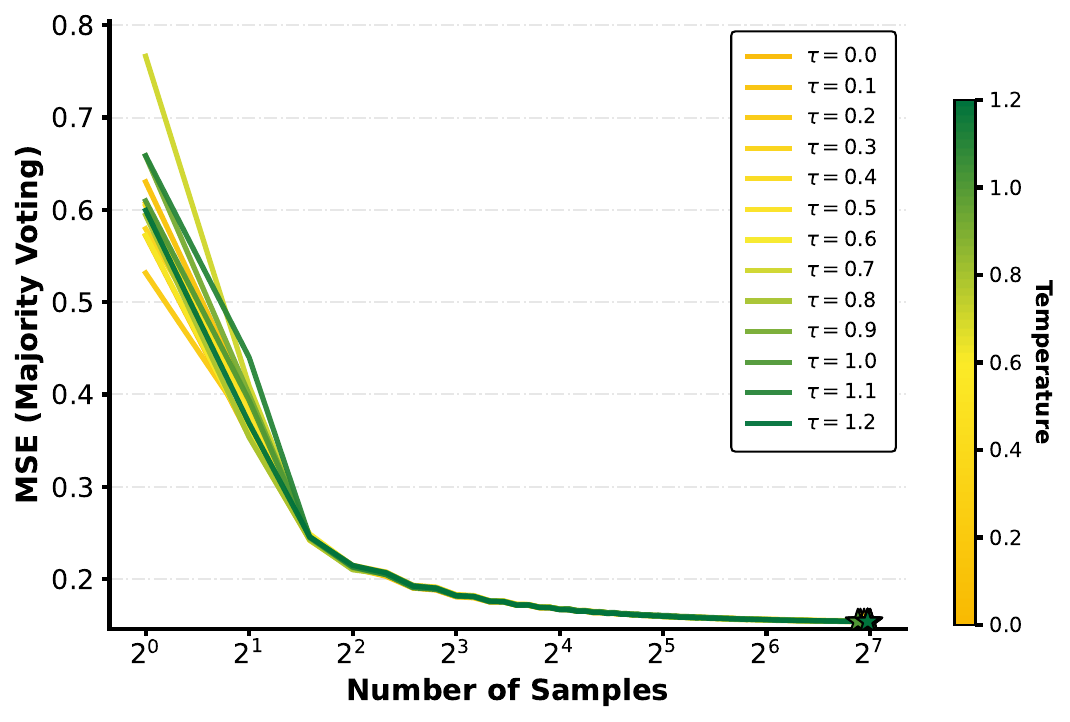}
        \caption{MV  vs. Temperature}
    \end{subfigure}
\vskip -0.1in
    \caption{Performance of \textit{Moirai} on the data \textit{ETTh1} under different scaling factors and aggregation functions.
    }
    \label{fig:moirai&etth}
\end{figure*}

\begin{figure*}[t!]
    \centering
    \begin{subfigure}{0.3\textwidth}
        \includegraphics[width=\linewidth]{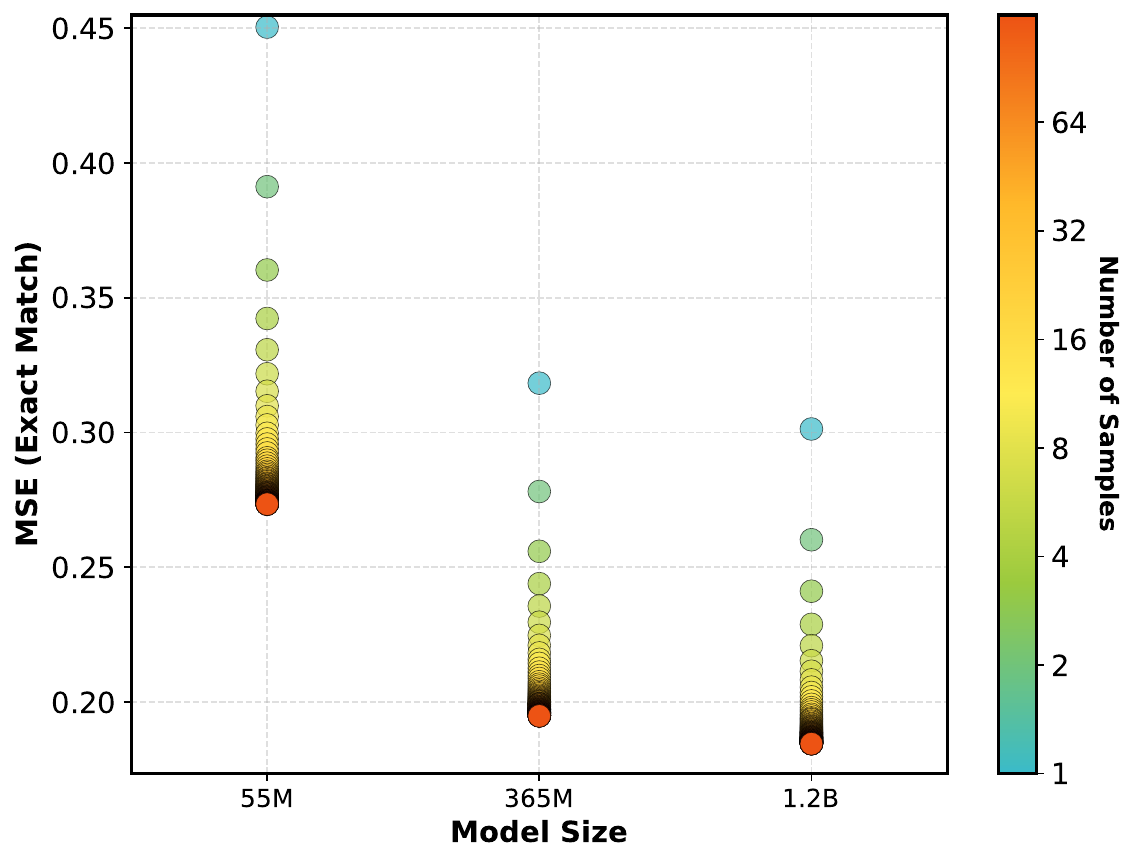}
        \caption{EM  vs. Model Size}
    \end{subfigure}
    \begin{subfigure}{0.34\textwidth}
        \includegraphics[width=\linewidth]{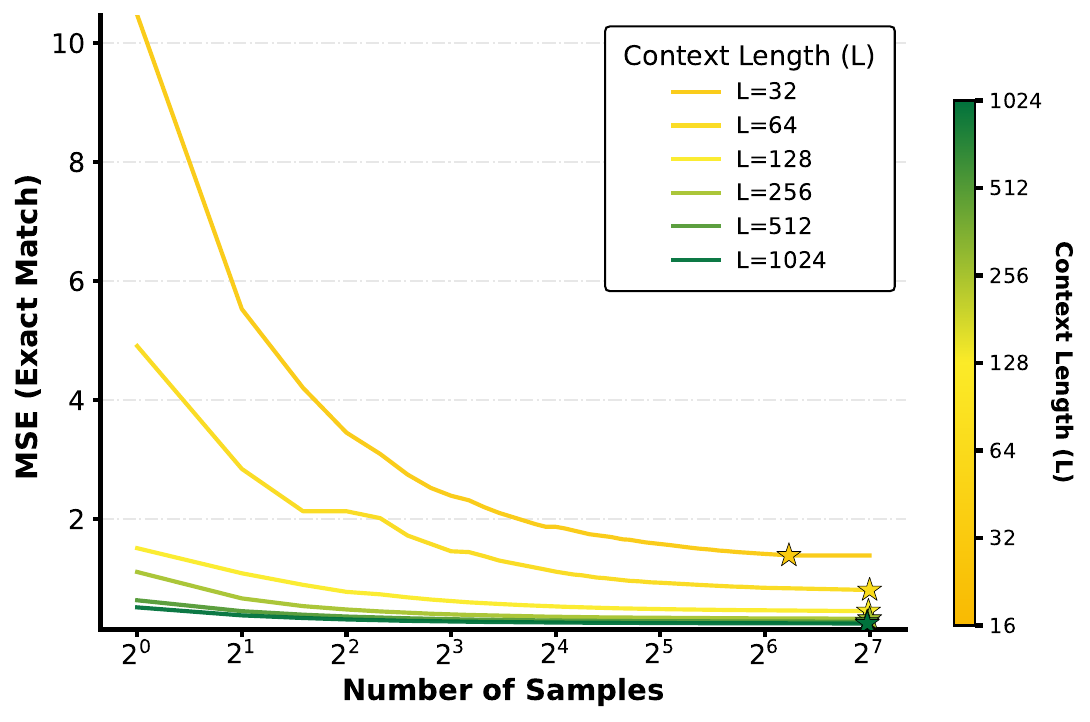}
        \caption{EM  vs. Context Length}
    \end{subfigure}
    \begin{subfigure}{0.34\textwidth}
        \includegraphics[width=\linewidth]{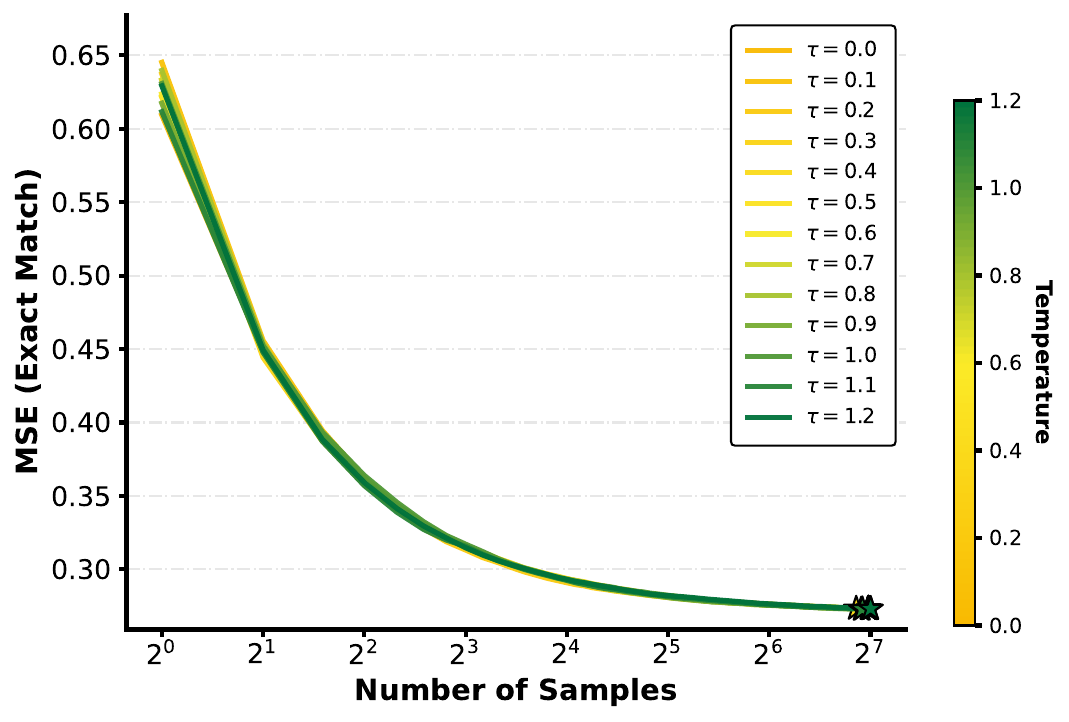}
        \caption{EM  vs. Temperature}
    \end{subfigure}
    \\
    \begin{subfigure}{0.3\textwidth}
        \includegraphics[width=\linewidth]{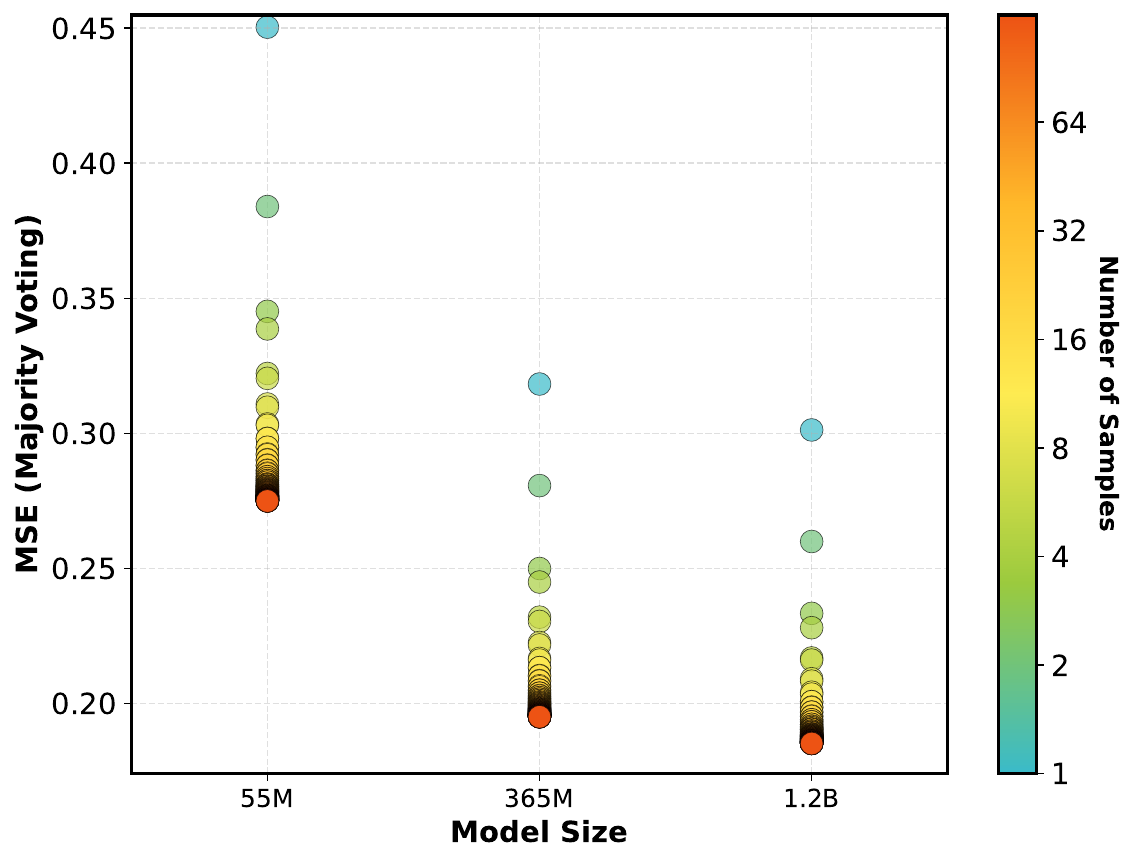}
        \caption{MV  vs. Model Size}
    \end{subfigure}
    \begin{subfigure}{0.34\textwidth}
        \includegraphics[width=\linewidth]{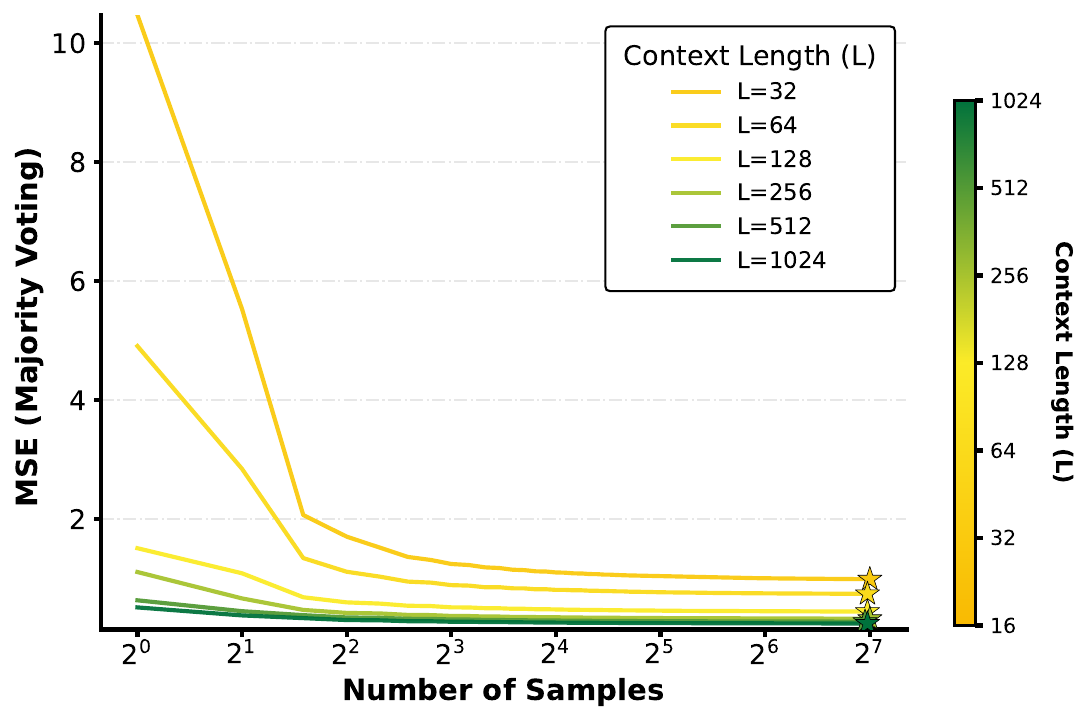}
        \caption{MV  vs. Context Length}
    \end{subfigure}
    \begin{subfigure}{0.34\textwidth}
        \includegraphics[width=\linewidth]{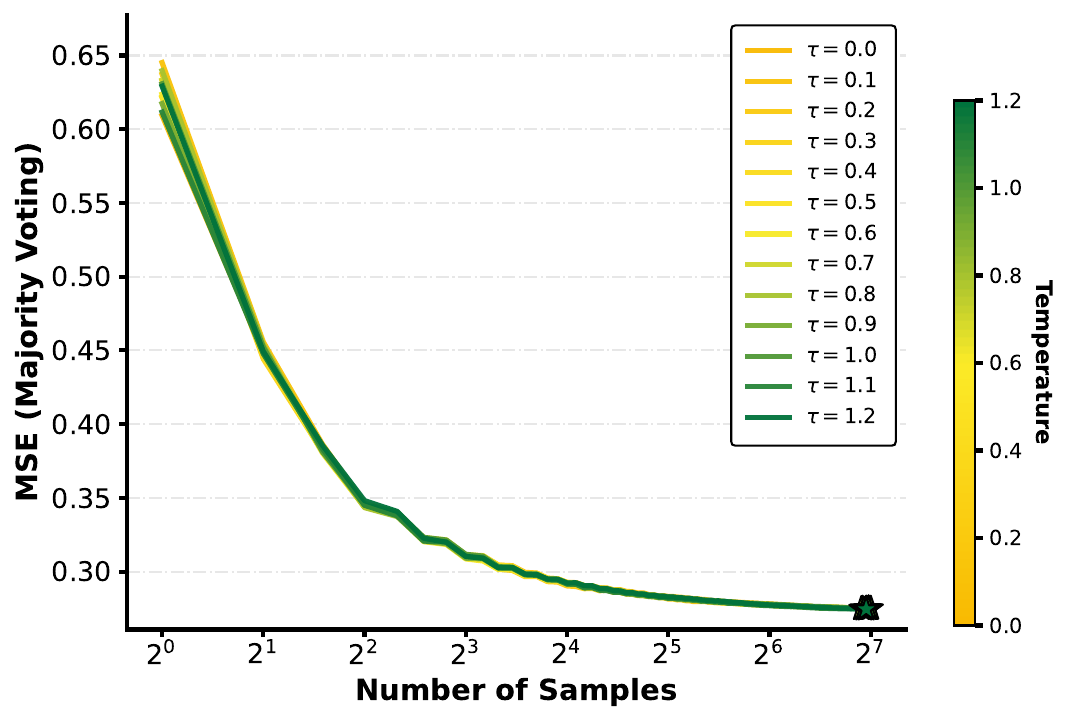}
        \caption{MV  vs. Temperature}
    \end{subfigure}
\vskip -0.1in
    \caption{Performance of \textit{Moirai} on the data \textit{Electricity} under different scaling factors and aggregation functions.
    }
    \label{fig:moirai&electricity}
\end{figure*}

\begin{figure*}[t!]
    \centering
    \begin{subfigure}{0.3\textwidth}
        \includegraphics[width=\linewidth]{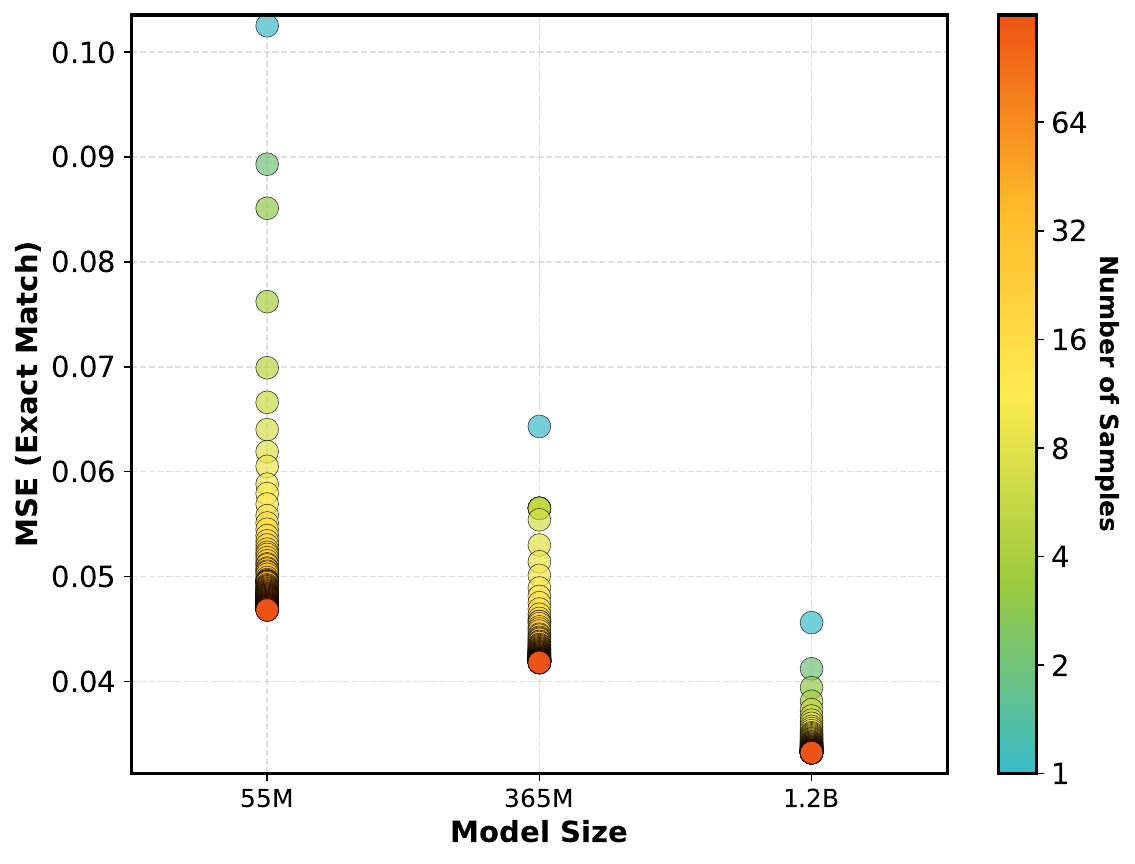}
        \caption{EM  vs. Model Size}
    \end{subfigure}
    \begin{subfigure}{0.34\textwidth}
        \includegraphics[width=\linewidth]{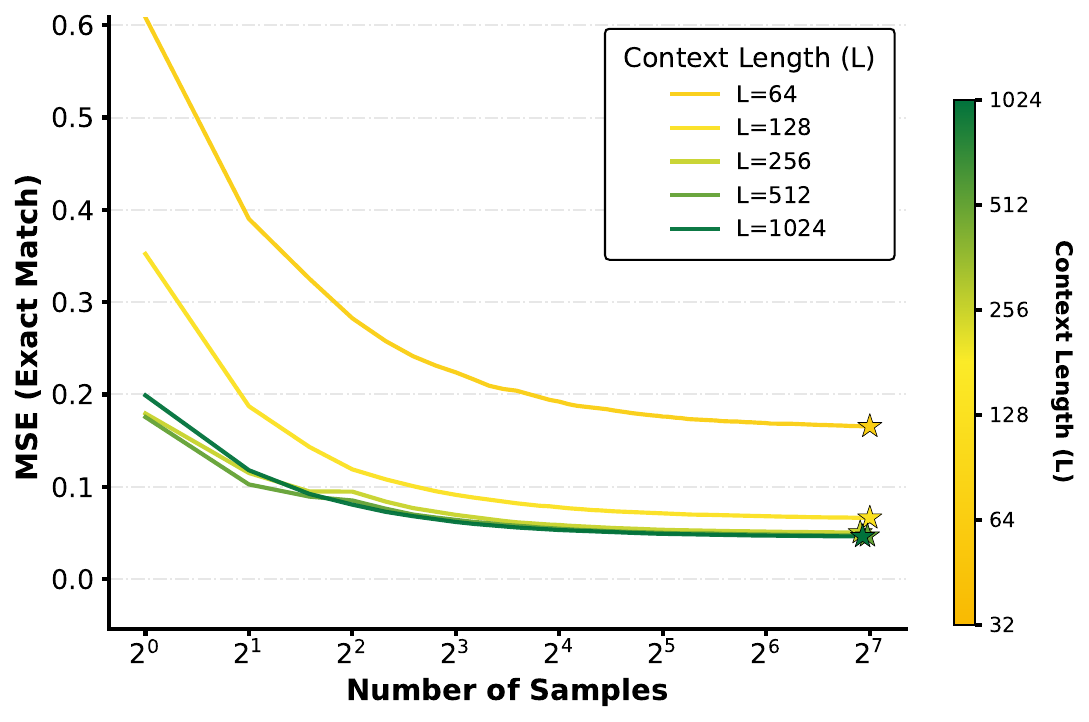}
        \caption{EM  vs. Context Length}
    \end{subfigure}
    \begin{subfigure}{0.34\textwidth}
        \includegraphics[width=\linewidth]{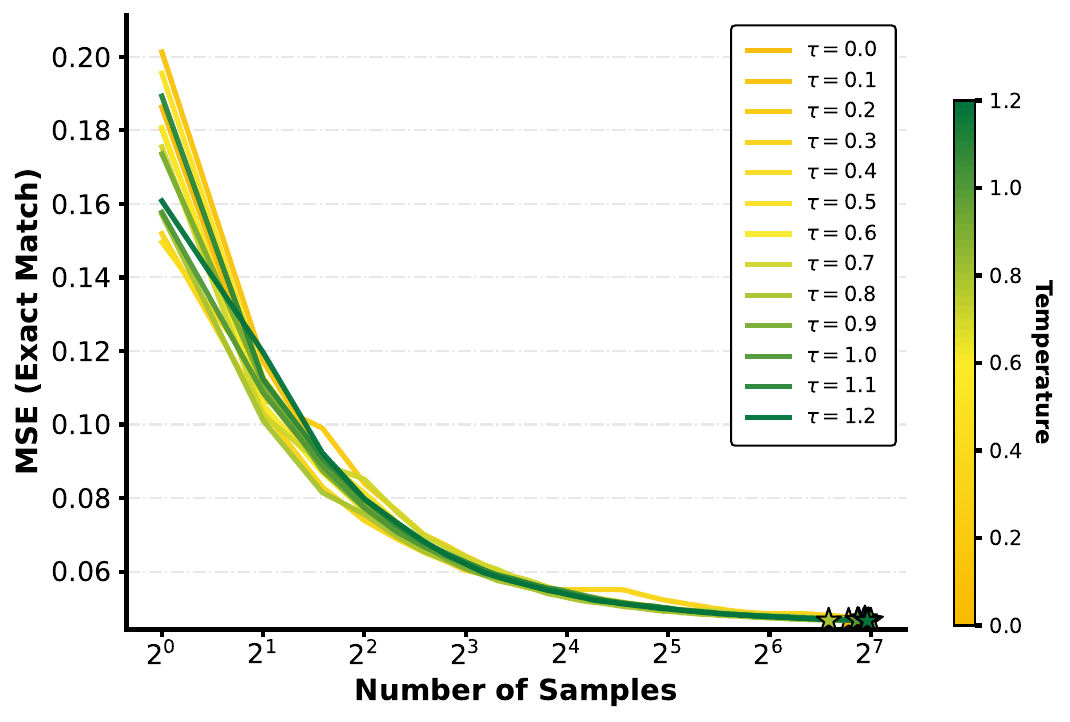}
        \caption{EM  vs. Temperature}
    \end{subfigure}
    \\
    \begin{subfigure}{0.3\textwidth}
        \includegraphics[width=\linewidth]{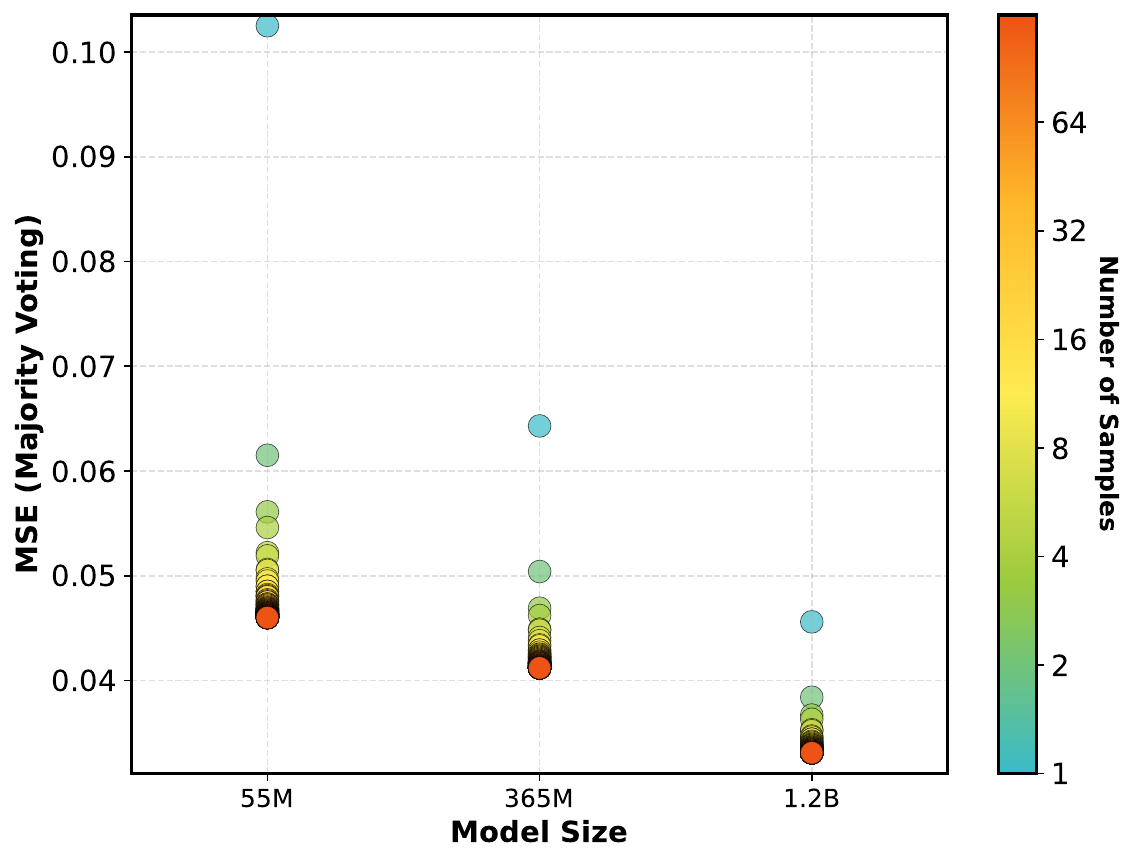}
        \caption{MV  vs. Model Size}
    \end{subfigure}
    \begin{subfigure}{0.34\textwidth}
        \includegraphics[width=\linewidth]{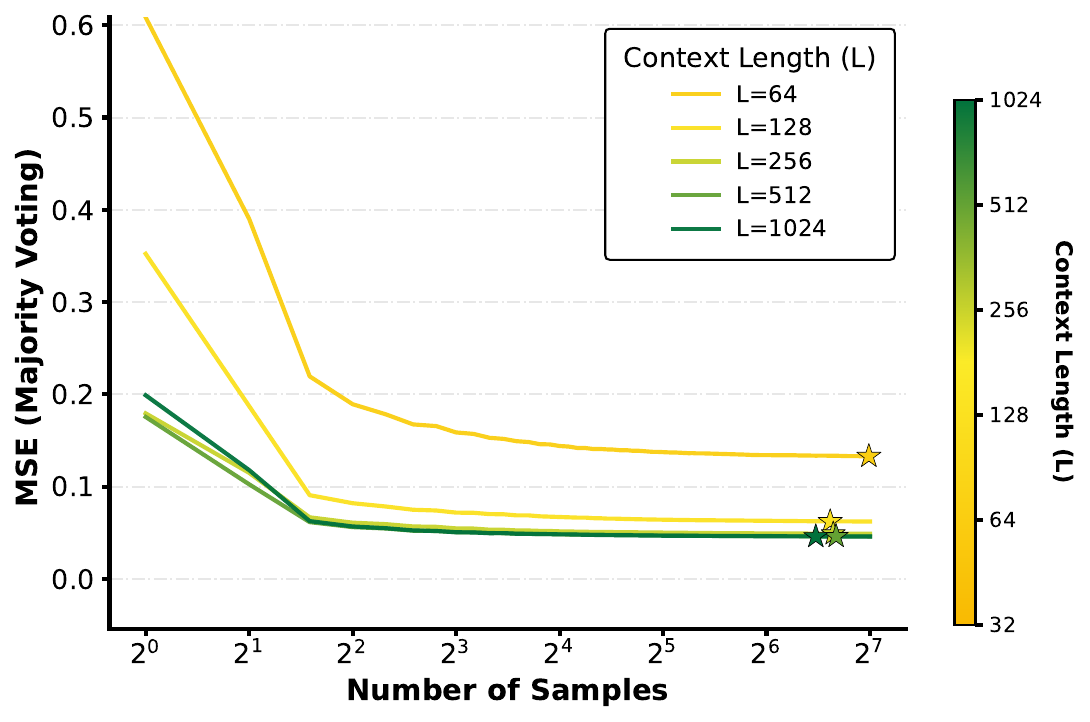}
        \caption{MV  vs. Context Length}
    \end{subfigure}
    \begin{subfigure}{0.34\textwidth}
        \includegraphics[width=\linewidth]{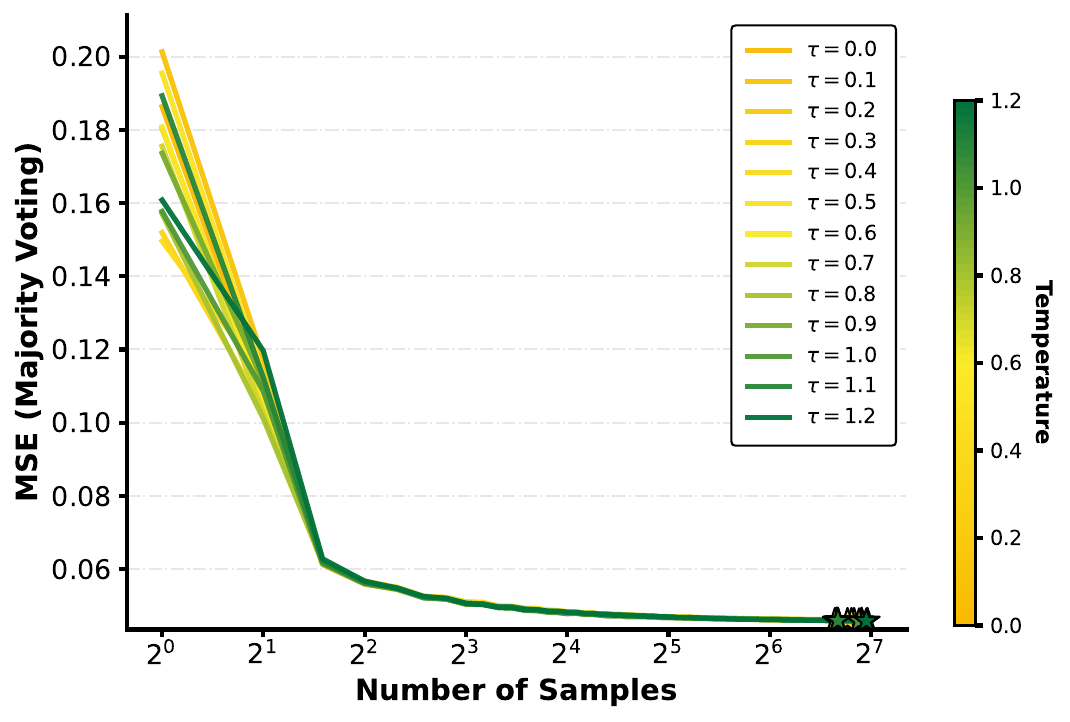}
        \caption{MV  vs. Temperature}
    \end{subfigure}
\vskip -0.1in
    \caption{Performance of \textit{Moirai} on the data \textit{Traffic} under different scaling factors and aggregation functions.
    }
    \label{fig:moirai&traffic}
\end{figure*}

\subsection{TimesFM}
Due to the lack of temperature and decoding strategies in TimesFM, scaling inference is ineffective. Thus, we only investigate the impact of context length on the model here. The results are shown in Table~\ref{tab:mse_mae_performance}, with the best results in \textbf{bold} and second-best \underline{underline}. Due to TimesFM was pre-trained on the Electricity and Traffic datasets~\cite{das2024decoder}, its predictive performance on these two datasets is generally superior, and the MSE shows a significant decreasing trend as $L$ increases, demonstrating relatively stable scaling benefits. 
On both ETTh1 and ETTm1, the forecasting error generally decreases with increasing context length, suggesting that TimesFM broadly conforms to the inference scaling law. Nevertheless, local non-monotonic variations (e.g., the optimal performance at $L=256$ on ETTh1) indicate that the benefits of longer contexts may saturate or fluctuate at specific scales.

\begin{table}[htbp]
  \centering
  \vspace{-4mm}
  \caption{Performance of \textit{TimesFM} on all datasets under the different context length using MSE and MAE metrics.}
  \label{tab:mse_mae_performance}
  \small 
    \begin{tabular}{@{}lcccccccccccc@{}}
      \toprule
      \multirow{2}{*}{Dataset} & \multicolumn{2}{c}{$L=32$} & \multicolumn{2}{c}{$L=64$} & \multicolumn{2}{c}{$L=128$} & \multicolumn{2}{c}{$L=256$} & \multicolumn{2}{c}{$L=512$} & \multicolumn{2}{c}{$L=1024$} \\
      \cmidrule(lr){2-3} \cmidrule(lr){4-5} \cmidrule(lr){6-7} \cmidrule(lr){8-9} \cmidrule(lr){10-11} \cmidrule(lr){12-13}
      & MSE & MAE & MSE & MAE & MSE & MAE & MSE & MAE & MSE & MAE & MSE & MAE \\
      \midrule
    ETTh1          & 0.1693&0.3041&0.1607&0.2936&0.1467&0.2819&\textbf{0.1376}&\textbf{0.2728}&0.1411&0.2757&\underline{0.1404}&\underline{0.2729}\\
    ETTm1          &
0.1028&0.2308&0.0973&0.2253&0.0941&0.2225&0.0853&0.2080&\underline{0.0752}&\underline{0.1934}&\textbf{0.0707}&\textbf{0.1867}\\   
      Electricity    & 0.8421&0.6881&0.6461&0.5942&0.3119&0.3944&0.2151&0.3240&\underline{0.1887}&\underline{0.2982}&\textbf{0.1857}&\textbf{0.2956} \\
      Traffic   & 0.2379&0.3373&0.1283&0.2390&0.0622&0.1485&0.0494&0.1208&\underline{0.0440}&\underline{0.1108}&\textbf{0.0400}&\textbf{0.1036}\\
      \bottomrule
    \end{tabular}
\end{table}

\subsection{Moirai}
Moirai exhibits a relatively regular inference scaling behavior, which is shown in Figures~\ref{fig:moirai&etth},\ref{fig:moirai&electricity}, and \ref{fig:moirai&traffic}. 
As the model size increases, the prediction error generally decreases. However, on the ETTh1 dataset, the model achieves optimal performance at a medium scale, and further increasing the model size leads to an increase in MSE, indicating that scaling the model size does not always yield monotonic improvements. 
In terms of context length, when the number of samples is small, the MSE decreases significantly with increasing context length. As the number of samples increases, the prediction error decreases rapidly and converges to a similar level, with an overall performance improvement of up to approximately $90\%$. Notably, with a small number of samples, the MV aggregation function shows significantly faster error reduction and convergence compared to EM. 
Regarding the temperature setting, because Moirai's output distribution is not explicitly divided by the temperature coefficient $\tau$ during inference (refer to Eq.4 in \cite{woo2024unified}), making it difficult to effectively sample the distribution under low conditions. 
This renders the temperature's regulatory effect on inference scaling largely ineffective in Moirai. Therefore, regardless of the temperature value or the number of samples, the model performance shows a consistent and significant convergence trend.

\subsection{TimeMoE}

Due to the lack of temperature and decoding strategies in TimeMoE, scaling inference is ineffective. Thus, we only investigate the impact of context length and model size on the model here. The results are shown in Table~\ref{tab:timemoe_performance_comparison}, with the best results in \textbf{bold} and second-best \underline{underline}.
We further compared the performance differences of TimeMoE at different model sizes as the context length varied. Taking the Traffic dataset as an example, \textit{TimeMoE-50M} achieved the best results at $L=512$ and the second-best at $L=1024$, while \textit{TimeMoE-200M}'s optimal performance occurred at $L=256$, followed by $L=512$. This difference indicates that different model sizes have inconsistent sensitivity to context length. Overall, regardless of model size, the performance trend of TimeMoE across different datasets as the context length increases remains largely consistent, but it does not exhibit the standard inference scaling behavior of monotonically improving with $L$, suggesting that simply increasing the context length is not sufficient to guarantee stable performance improvement.

\begin{table}[htbp]
  \centering
  \vspace{-4mm}
  \caption{Performance of \textit{TimeMoE} (50M vs 200M) across different context lengths using MSE and MAE metrics.}
  \label{tab:timemoe_performance_comparison}
  \small
  \resizebox{1\columnwidth}{!}{
    \begin{tabular}{@{}lcccccccc|cccccccc@{}}
      \toprule
      \multirow{3}{*}{{Dataset}} & \multicolumn{8}{c|}{\texttt{TimeMoE-50M}} & \multicolumn{8}{c}{\texttt{TimeMoE-200M}} \\
      \cmidrule(lr){2-9} \cmidrule(lr){10-17}
      & \multicolumn{2}{c}{$L=128$} & \multicolumn{2}{c}{$L=256$} & \multicolumn{2}{c}{$L=512$} & \multicolumn{2}{c|}{$L=1024$} & \multicolumn{2}{c}{$L=128$} & \multicolumn{2}{c}{$L=256$} & \multicolumn{2}{c}{$L=512$} & \multicolumn{2}{c}{$L=1024$} \\
      \cmidrule(lr){2-3} \cmidrule(lr){4-5} \cmidrule(lr){6-7} \cmidrule(lr){8-9} \cmidrule(lr){10-11} \cmidrule(lr){12-13} \cmidrule(lr){14-15} \cmidrule(lr){16-17}
      & MSE & MAE & MSE & MAE & MSE & MAE & MSE & MAE & MSE & MAE & MSE & MAE & MSE & MAE & MSE & MAE \\
      \midrule
      ETTh1       & 0.2429&0.3777&0.2423&0.3771&\underline{0.2412}&\underline{0.3761}&\textbf{0.2403}&\textbf{0.3755}& 0.2443&0.3787&0.2432&0.3778&\underline{0.2412}&\underline{0.3760}&\textbf{0.2403}&\textbf{0.3755}\\
       ETTm1 & 0.1192 & 0.2515 & 0.1185 & 0.2508 & \underline{0.1178} & \textbf{0.2500} & \textbf{0.1177} & \textbf{0.2500} & 0.1199 & 0.2518 & 0.1191 & 0.2511 & \underline{0.1183} & \underline{0.2502} & \textbf{0.1181} & \textbf{0.2500}\\
       Electricity & \textbf{0.8757}&\textbf{0.7023}&\underline{0.9032}&\underline{0.7151}&0.9099&0.7183&0.9144&0.7207 & \textbf{0.8781}&\textbf{0.7032}&\underline{0.9004}&\underline{0.7137}&0.9110&0.7183&0.9201&0.7222\\
      Traffic     &0.3311&0.4200&0.3274&0.4160&\textbf{0.3205}&\textbf{0.4118}&\underline{0.3222}&\underline{0.4138} & 0.3261&0.4167&\textbf{0.3255}&\textbf{0.4158}&\underline{0.3260}&\underline{0.4163}&0.3281&0.4184 \\
      \bottomrule
    \end{tabular}
  }
\end{table}

\section{Additional Diversified Scaling Inference Experiments}\label{app:how}

We provide all the results of how diversified scaling influences the inference performance under more TSFMs and varying datasets in this part, and show the characteristics of different models with diversified sampling.
\subsection{TimesFM}
\begin{figure*}[h]
    \centering
\includegraphics[width=0.92\linewidth]{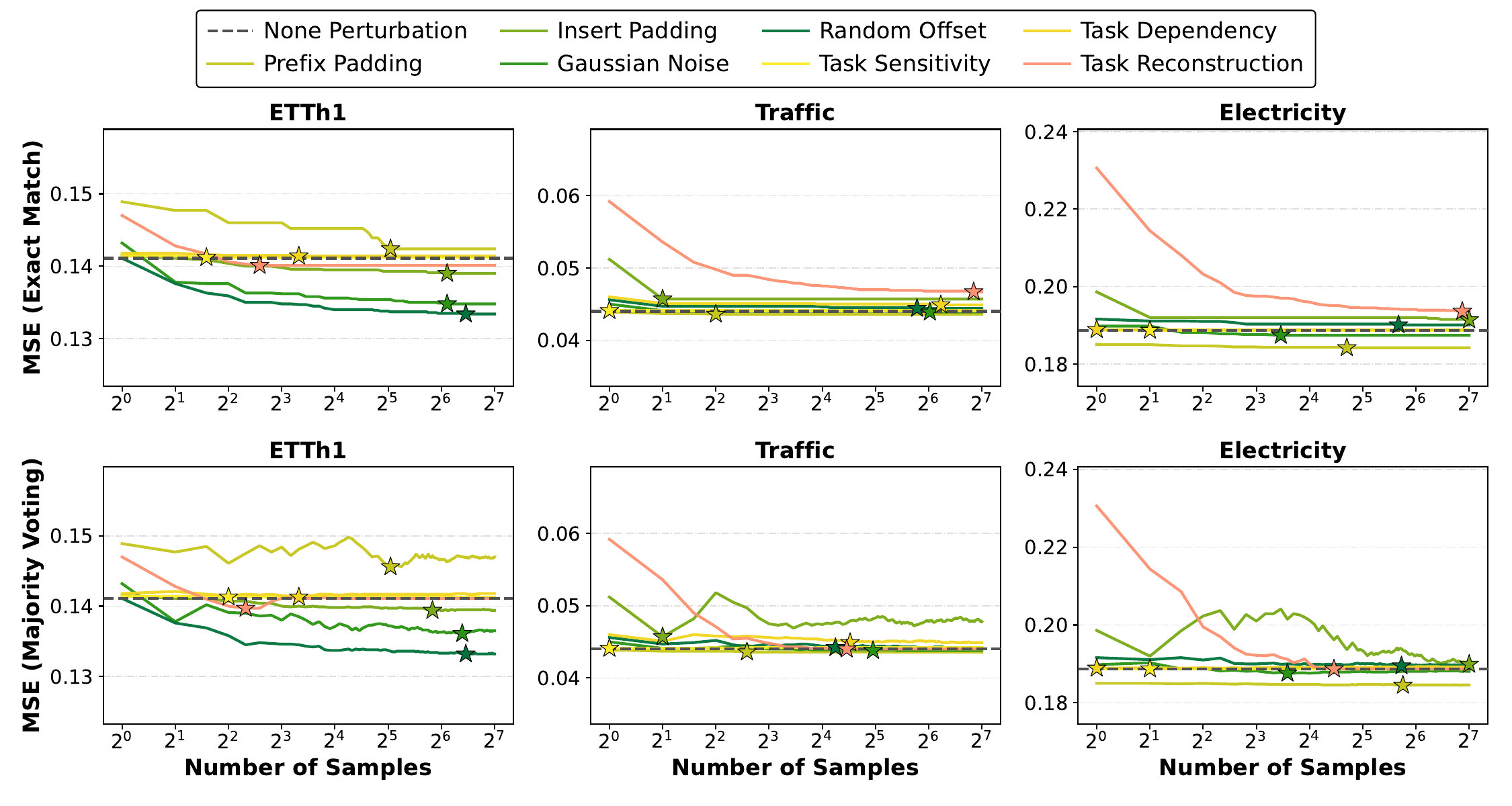}
        \vspace{-5mm}
    \caption{Performance of \textit{TimesFM} on different datasets with effective perturbations.}
    \label{fig:how timesfm}
\end{figure*}

Figure~\ref{fig:how timesfm} reports the diversified inference results of TimesFM, a TSFM without decoding strategies. 
Since TimesFM was pre-trained on the Electricity and Traffic datasets, we primarily focus on its performance on the dataset ETTh1. 
Experimental results show that on ETTh1, all effective input perturbation methods, except for Prefix Padding, significantly reduce the MSE as the number of samples increases, eventually converging to a level comparable to or even better than standard sampling. 
In contrast, on the Traffic and Electricity datasets, the inference performance after introducing input perturbations did not improve to the same extent with increasing sampling budget. Although the MSE still showed a decreasing and gradually converging trend, its performance ceiling usually could not surpass or even approach that of standard sampling. This may be related to their use in pre-training.
Further comparison of the two aggregation functions reveals that under the EM aggregation method, the MSE changes more smoothly with the number of samples, and in most settings, it achieves optimal performance superior to MV.

\begin{figure*}[!t]
    \centering
    \includegraphics[width=0.92\linewidth]{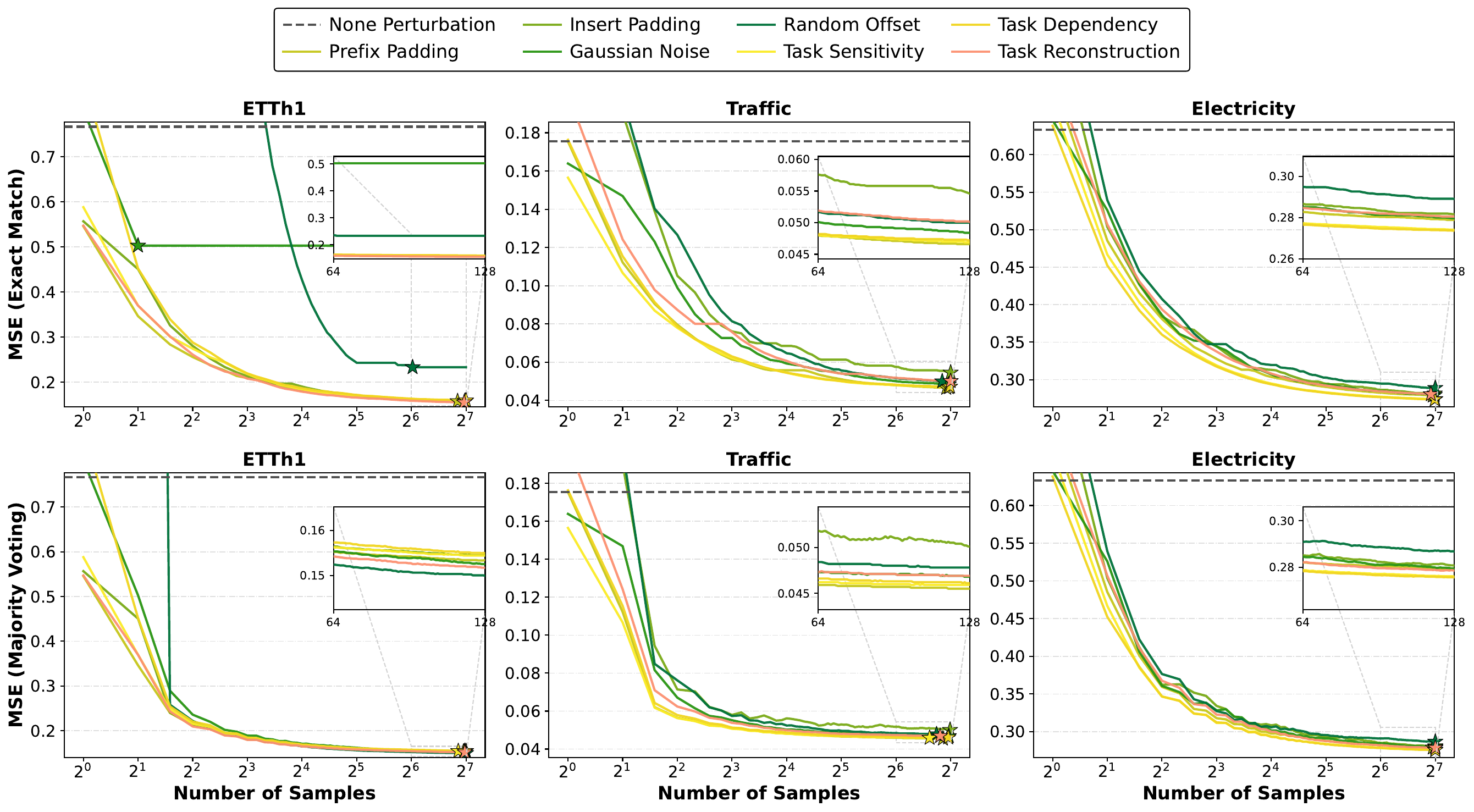}
        \vspace{-5mm}
    \caption{Performance of \textit{Moirai} on different datasets with effective perturbations.}
    \label{fig:how moirai}
\end{figure*}

\begin{figure*}[!t]
    \centering
    \includegraphics[width=0.92\linewidth]{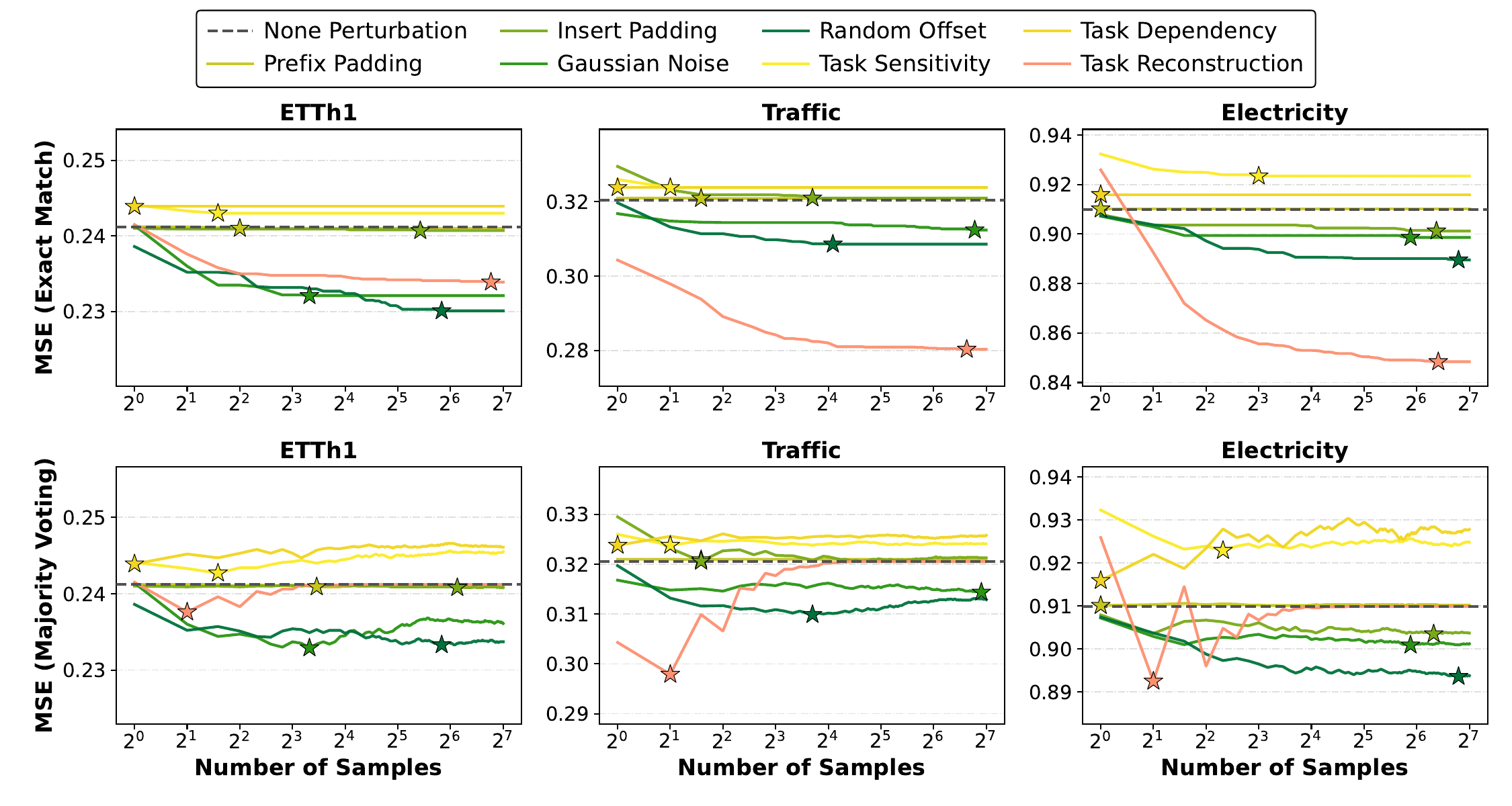}
        \vspace{-5mm}
    \caption{Performance of \textit{TimeMoE} on different datasets with effective perturbations.}
    \label{fig:how timemoe}
\end{figure*}

\subsection{Moirai}
Figure~\ref{fig:how moirai} shows that Moirai benefits clearly from increased inference budgets. On all three datasets, MSE consistently decreases as the number of samples grows, with a sharp drop at small budgets followed by gradual convergence, and the performance can be partly attributed to Moirai, which inherently adopts a multi-sample averaging strategy during inference, making additional samples directly contributive to performance improvement.
The reduction is observed under both EM and MV aggregation, although EM typically yields smoother curves and lower final errors. Notably, similar to the Table~\ref{tab:model_dataset_perturbation}, \textit{Gaussian Noise} and \textit{Random Offset} perturbations are largely ineffective for the Moirai model. In comparative analyses across perturbation types, these two methods consistently result in higher MSEs than other perturbations and even display an atypical scaling behavior in which performance improves initially but fails to converge toward the curves of more effective perturbations. In contrast, \textit{Task Dependency} and \textit{Task Sensitivity}, perturbations remain consistently achieving the lowest errors across sampling budgets on Moirai. Overall, these results suggest that while increasing the sampling budget universally benefits inference, the effectiveness of diversified sampling critically depends on the alignment between perturbation design and model-specific inductive biases.

\subsection{TimeMoE}

Figure~\ref{fig:how timemoe} indicates that TimeMoE without decoding strategies exhibits weaker sensitivity to inference scaling. In this setting, \textit{Noise Perturbations} and \textit{Task Reconstruction perturbations} consistently achieve the best performance, benefiting from increased sampling budgets. In contrast, \textit{Structural Perturbations} display little variation as the number of samples grows: their MSE remains consistently higher than that of standard sampling and in some cases even degrades with additional samples. Moreover, due to the differences in aggregation methods, the optimal MSE achieved under MV aggregation in TimeMoE is often inferior to that obtained with EM aggregation. Overall, these results indicate that the effect of inference scaling in TimeMoE is highly dependent on the characteristics of the induced perturbations.

\section{Combination of Perturbation and Scaling Inference}\label{app:ablation}
We conduct a series of ablation studies on model sizes, context length, and temperature to better understand how inference scaling interacts with stochasticity and input capacity. Input diversity in these experiments is mainly introduced via Gaussian Noise perturbations. The results reveal how these factors jointly influence the diversity and quality of sampled candidates under aggregation functions.

\subsection{Ablation of Model Sizes}
We evaluate diversified inference scaling across \textit{Chronos-T5} models of varying sizes on ETTh1, as reported in Table~\ref{tab:ablationsss}. While larger models achieve lower absolute error, the relative gains from diversified sampling decrease with model size. In particular, strong models already attain low error without perturbations, leaving limited space for additional gains. Diversified sampling therefore benefits smaller and medium-sized models most, and should be applied more selectively to larger models where inference scaling returns saturate.

\begin{table}[htbp]
  \centering
  \vspace{-5mm}
  \caption{Comparison of \textit{Chronos-T5} across different size and perturbations under 128 samples.}
  \label{tab:chronos-comparison}
  \resizebox{\textwidth}{!}{%
    \begin{tabular}{lcccccccccc}
    \toprule
    \multirow{2}{*}{Perturbation} & \multicolumn{2}{c}{\texttt{Chronos-T5-Tiny}} & \multicolumn{2}{c}{\texttt{Chronos-T5-Mini}} & \multicolumn{2}{c}{\texttt{Chronos-T5-Small}} & \multicolumn{2}{c}{\texttt{Chronos-T5-Base}} & \multicolumn{2}{c}{\texttt{Chronos-T5-Large}} \\
    \cmidrule(lr){2-3} \cmidrule(lr){4-5} \cmidrule(lr){6-7} \cmidrule(lr){8-9} \cmidrule(lr){10-11}
          & EM    & MV    & EM    & MV    & EM    & MV    & EM    & MV    & EM    & MV \\
    \midrule   \rowcolor{gray!20}
None & 0.1543 & 0.1564 & 0.1516 & 0.1523 & \textbf{0.1418} & \textbf{0.1445} & \underline{0.1425} & \underline{0.1453 }& 0.1444 & 0.1481 \\
Prefix & 0.1548 (-0.3\%) & 0.1563 (+0.1\%) & 0.1532 (-1.1\%) & 0.1550 (-1.8\%) & \textbf{0.1421} (-0.2\%) & \textbf{0.1456} (-0.8\%) & \underline{0.1432} (-0.5\%) & \underline{0.1463} (-0.7\%) & 0.1439 (+0.3\%) & 0.1486 (-0.3\%) \\
Gaussian & 0.1402 (+9.1\%) & 0.1406 (+10.1\%) & 0.1527 (-0.7\%) & 0.1548 (-1.6\%) & \underline{0.1414} (+0.3\%) & 0.1441 (+0.3\%) & \underline{0.1414} (+0.8\%) & \underline{0.1431} (+1.5\%) & \textbf{0.1359} (+5.9\%) & \textbf{0.1377} (+7.0\%) \\
Dependency & 0.1539 (+0.3\%) & 0.1539 (+1.6\%) & 0.1592 (-5.0\%) & 0.1617 (-6.2\%) & \textbf{0.1402} (+1.1\%) & \textbf{0.1432} (+0.9\%) & \underline{0.1434} (-0.6\%) & \underline{0.1467 }(-1.0\%) & 0.1471 (-1.9\%) & 0.1502 (-1.4\%) \\
    \bottomrule
    \end{tabular}\label{tab:ablationsss}
  }\vspace{-5mm}
\end{table}

\subsection{Ablation of Context Length}
\begin{figure*}[h!]
    \centering
    \begin{subfigure}{0.33\textwidth}
        \includegraphics[width=\linewidth]{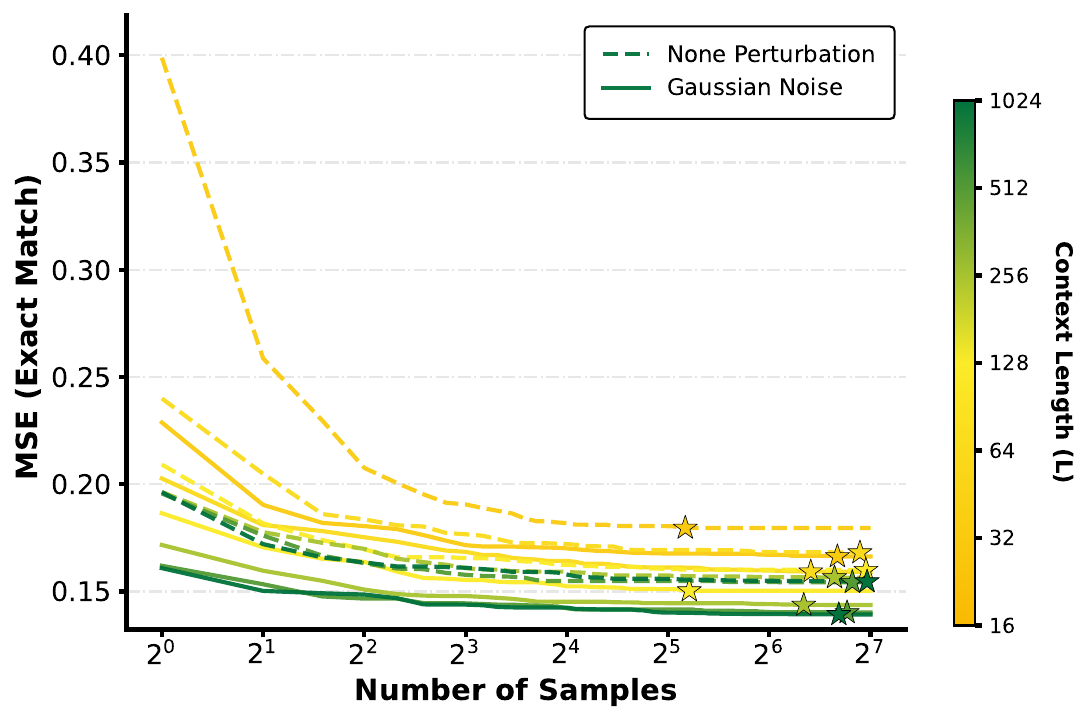}
        \caption{EM on \textit{ETTh1}}
    \end{subfigure}
    \begin{subfigure}{0.33\textwidth}
        \includegraphics[width=\linewidth]{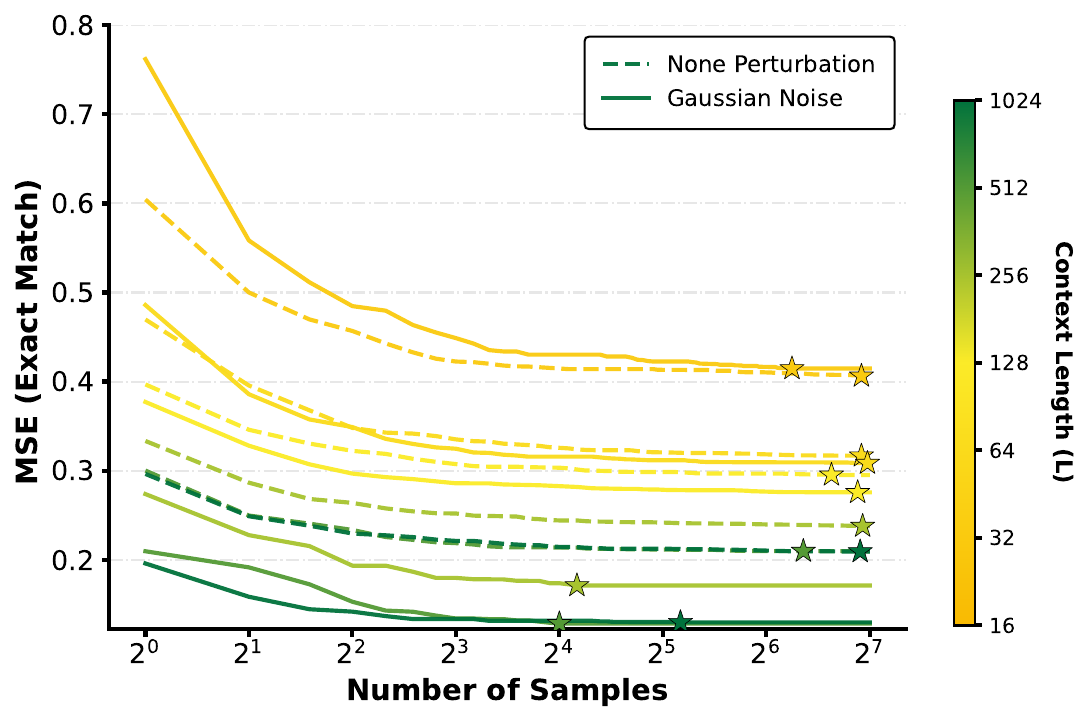}
        \caption{EM on \textit{Traffic}}
    \end{subfigure}
    \begin{subfigure}{0.33\textwidth}
        \includegraphics[width=\linewidth]{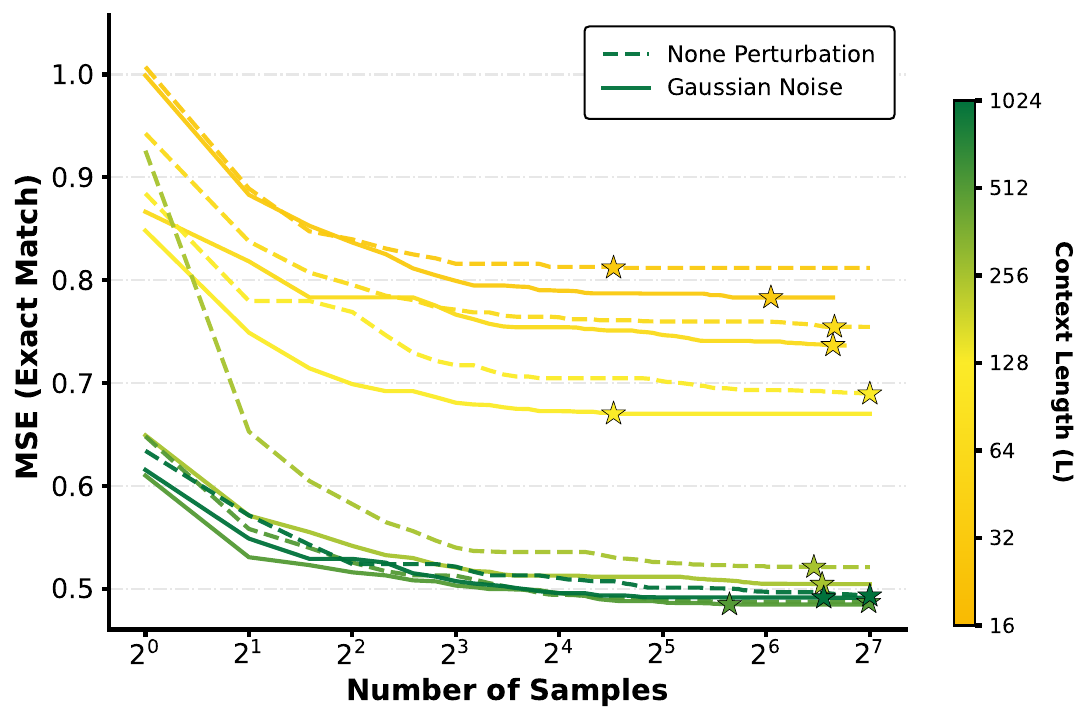}
        \caption{EM on \textit{Electricity}}
    \end{subfigure}
    \\
    \begin{subfigure}{0.33\textwidth}
        \includegraphics[width=\linewidth]{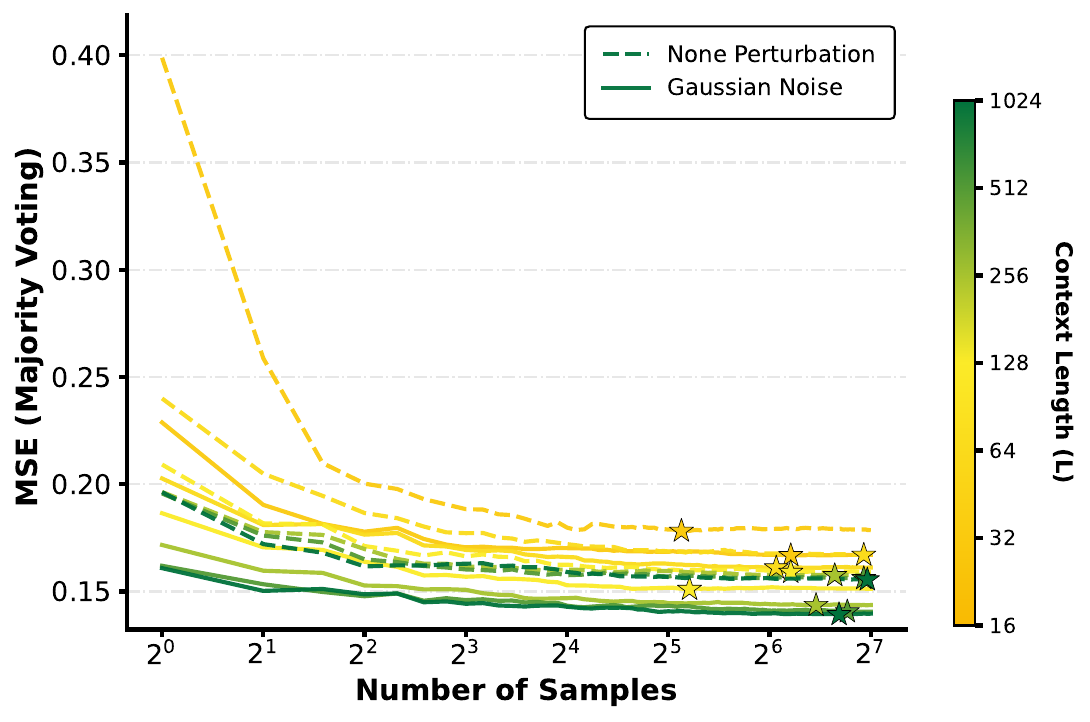}
        \caption{MV on \textit{ETTh1}}
    \end{subfigure}
    \begin{subfigure}{0.33\textwidth}
        \includegraphics[width=\linewidth]{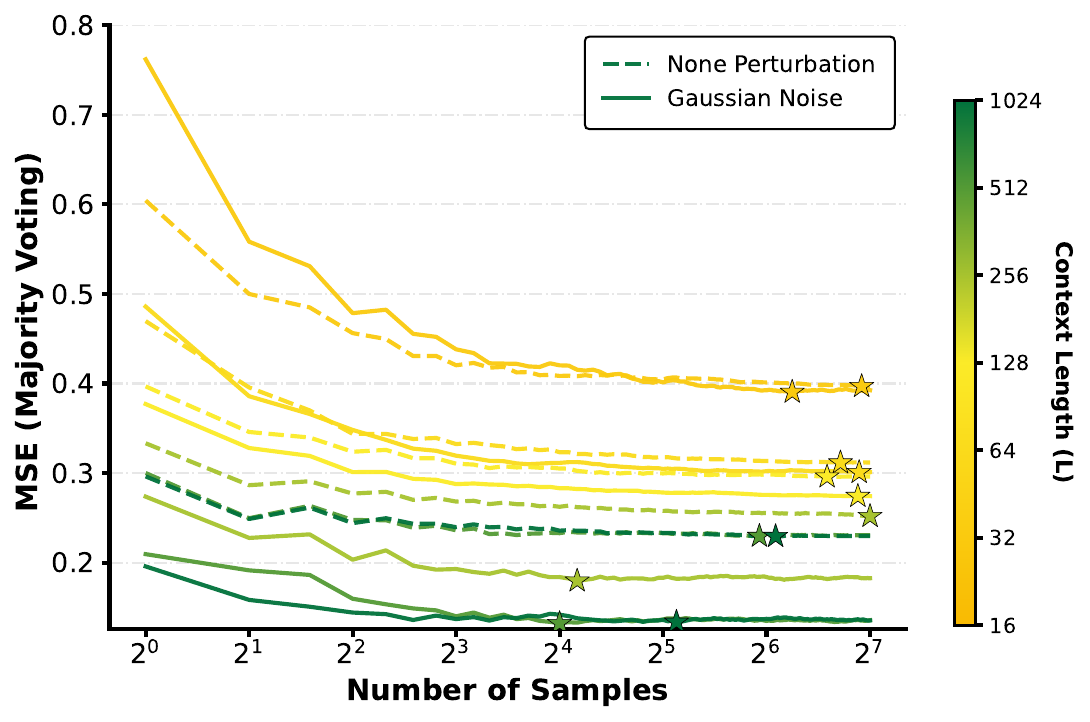}
        \caption{MV on \textit{Traffic}}
    \end{subfigure}
    \begin{subfigure}{0.33\textwidth}
        \includegraphics[width=\linewidth]{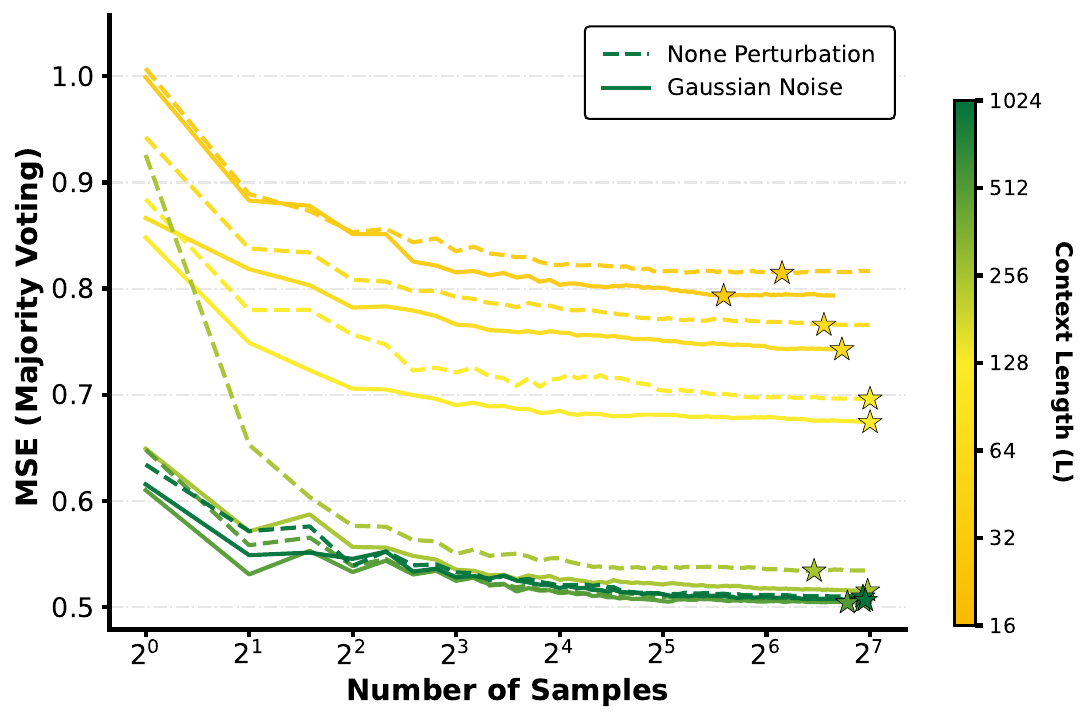}
        \caption{MV on \textit{Electricity}}
    \end{subfigure}
\vskip -0.1in
    \caption{Relationship between context length and inference scaling performance on \textit{Chronos-T5}.}
    \label{fig:ablation of context length}
\end{figure*}

We vary the input context length while increasing the number of sampled predictions on \textit{Chronos-T5}, as shown in Figure~\ref{fig:ablation of context length}. Extending the context length consistently improves inference performance across datasets and aggregation functions by lowering the achievable error floor. Without input perturbations, these gains quickly saturate. When input perturbations are introduced, longer contexts lead to more stable and sustained scaling as $N$ increases. This effect is more pronounced under EM aggregation, while MV exhibits smoother but weaker improvements, indicating that longer contexts primarily enhance the quality of the best sampled candidates rather than the average prediction.

\subsection{Ablation of Temperature}
\begin{figure*}[t!]
    \centering
    \begin{subfigure}{0.33\textwidth}
        \includegraphics[width=\linewidth]{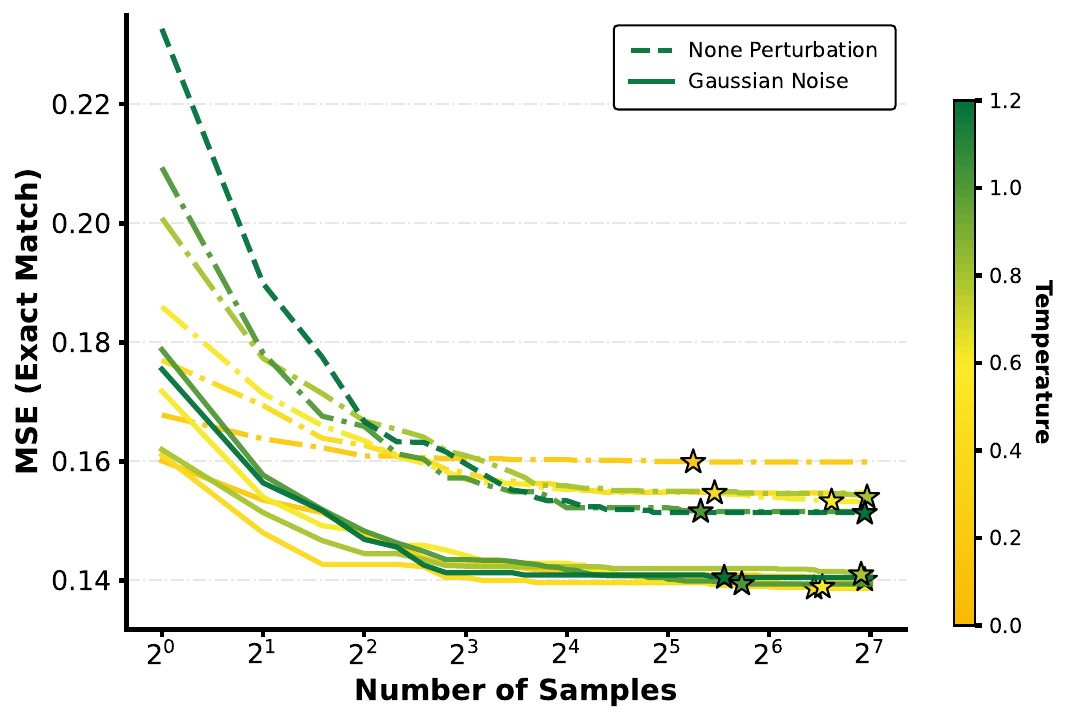}
        \caption{EM on \textit{ETTh1}}
    \end{subfigure}
    \begin{subfigure}{0.33\textwidth}
        \includegraphics[width=\linewidth]{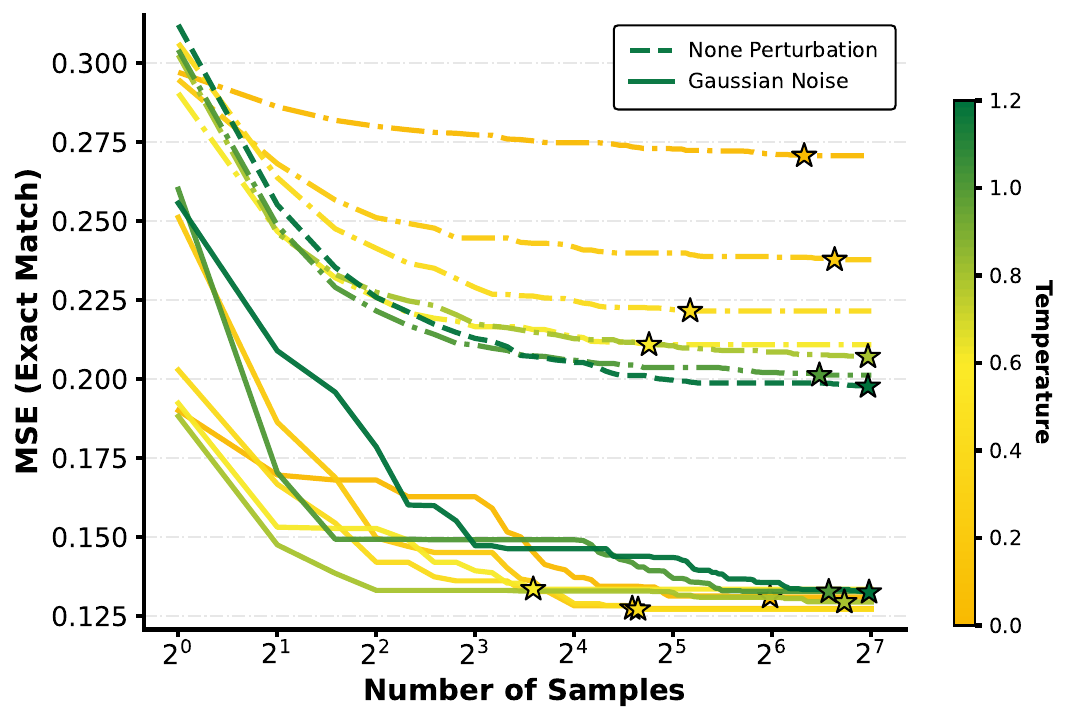}
        \caption{EM on \textit{Traffic}}
    \end{subfigure}
    \begin{subfigure}{0.33\textwidth}
        \includegraphics[width=\linewidth]{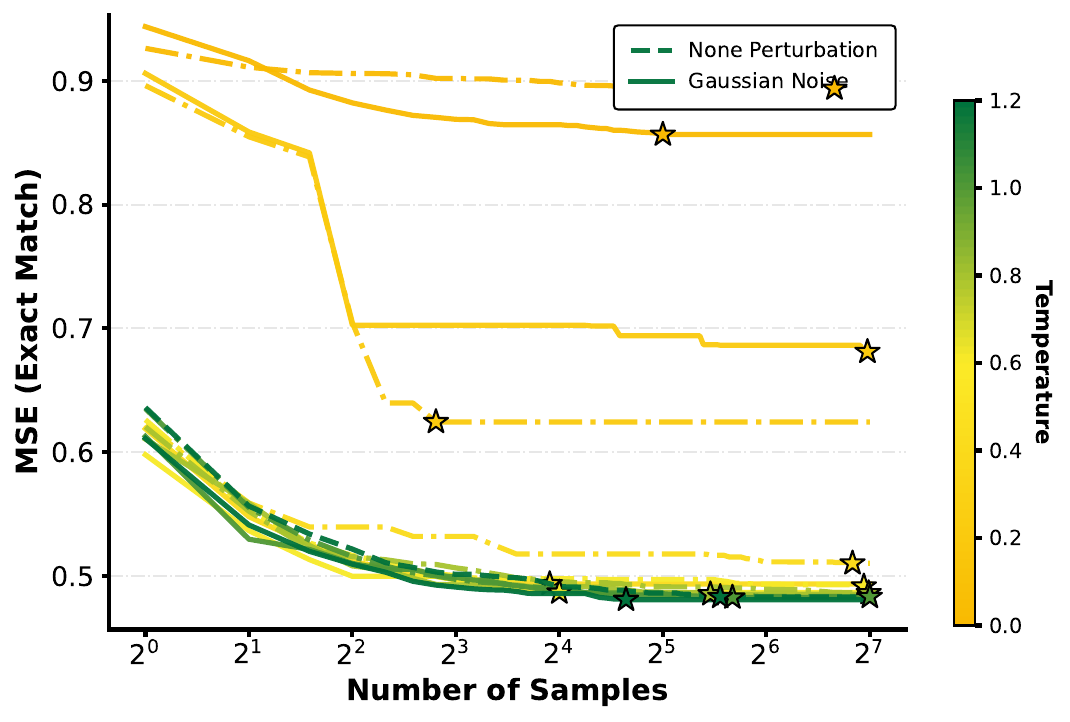}
        \caption{EM on \textit{Electricity}}
    \end{subfigure}
    \\
    \begin{subfigure}{0.33\textwidth}
        \includegraphics[width=\linewidth]{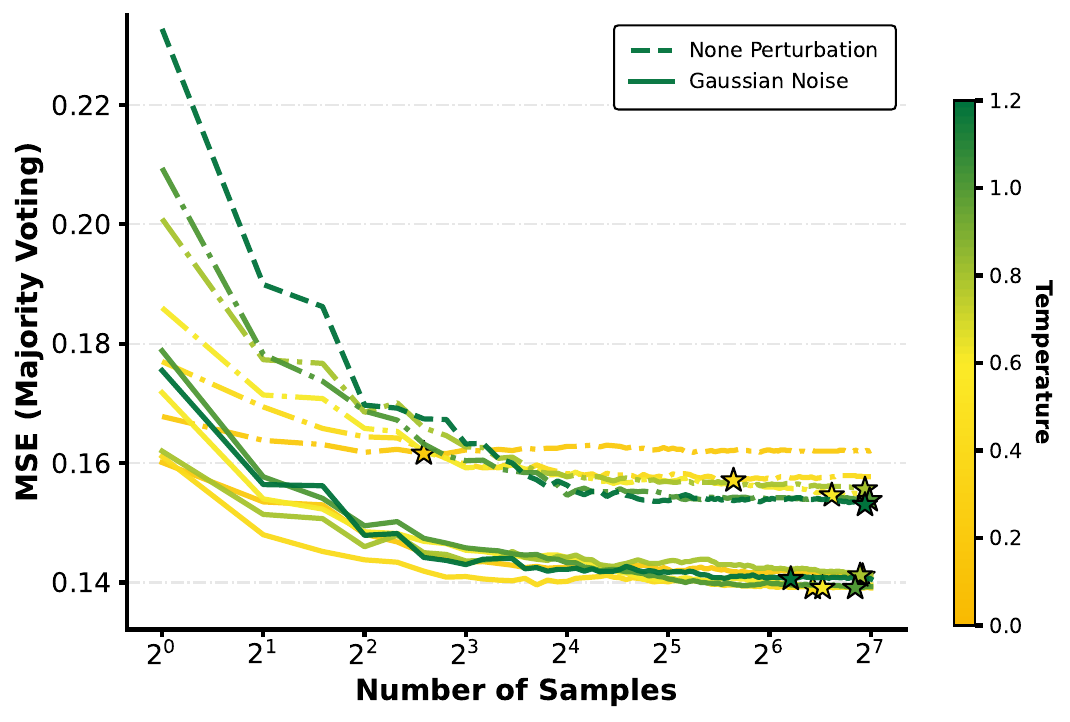}
        \caption{MV on \textit{ETTh1}}
    \end{subfigure}
    \begin{subfigure}{0.33\textwidth}
        \includegraphics[width=\linewidth]{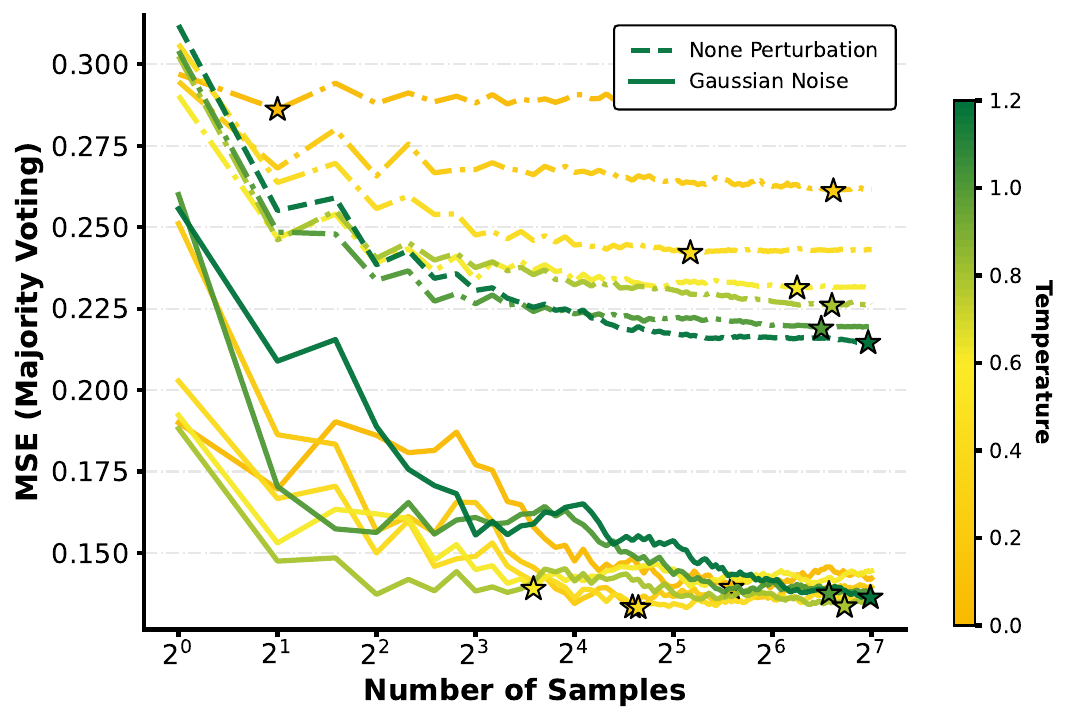}
        \caption{MV on \textit{Traffic}}
    \end{subfigure}
    \begin{subfigure}{0.33\textwidth}
        \includegraphics[width=\linewidth]{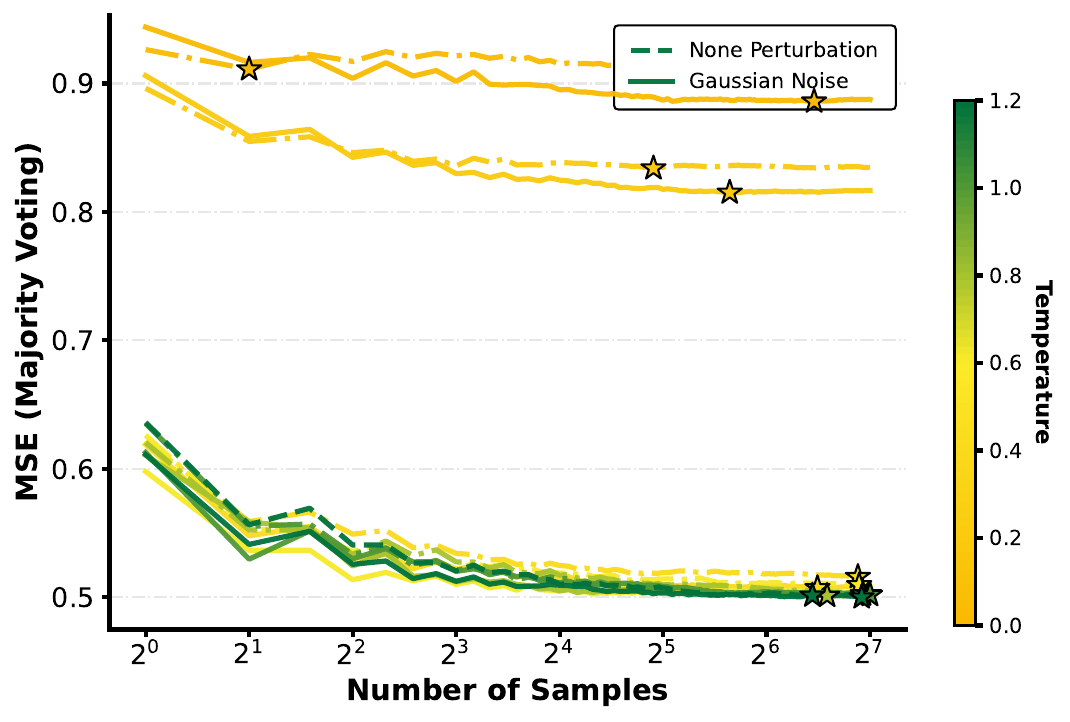}
        \caption{MV on \textit{Electricity}}
    \end{subfigure}
\vskip -0.1in
    \caption{Relationship between temperature and inference scaling performance on \textit{Chronos-T5}.}
    \label{fig:ablation of tem}
\end{figure*}

As shown in Figure~\ref{fig:ablation of tem}, we vary the sampling temperature on \textit{Chronos-T5} while fixing the model and context length. Excessively high temperatures already improve performance as $N$ increases but quickly saturate, while lower temperatures lead to unstable or degraded scaling. An intermediate temperature range consistently yields stronger and more stable inference gains under aggregations. Overall, the results indicate that while temperature tuning alone can improve early scaling, sustained inference gains require controlled diversity beyond purely stochastic decoding.

\end{document}